\documentclass{article} 
\usepackage{iclr2026_conference,times}

\makeatletter
\newif\if@submission
\@submissionfalse  
\makeatother

\makeatletter
\def\input@path{{./}{../}}
\makeatother

\usepackage{url}

\usepackage{amsmath,amssymb,mathtools,amsthm}
\usepackage{mathtools}
\usepackage{physics}
\usepackage{empheq}
\usepackage{enumerate}
\usepackage{enumitem}
\usepackage{pifont}
\usepackage[hidelinks]{hyperref} 
\usepackage[capitalize]{cleveref}
\usepackage{pgfplots}
\usepackage{pgfplotstable}
\usepgfplotslibrary{fillbetween}
\pgfplotsset{compat=1.17}
\usepackage{multirow}
\usepackage{xcolor}
\usepackage{colortbl}
\usepackage{adjustbox}
\usepackage{xspace}

\newtheorem{assumption}{Assumption}

\newtheorem*{remark}{Remark}

\usepackage[acronym]{glossaries}
\newacronym{acr:kkt}{KKT}{Karush-Kuhn-Tucker}
\newacronym{pde}{PDE}{partial differential equations}
\newacronym{cnn}{CNN}{convolutional neural network}
\newacronym{mpc}{MPC}{model predictive control}
\newacronym{piv}{PIV}{particle image velocimetry}
\newacronym{mlp}{MLP}{multi-layer perceptron}
\newacronym{dc3}{DC3}{Deep Constraint Completion and Correction}
\newacronym{jvp}{JVP}{Jacobian-vector product}
\newacronym{vjp}{VJP}{vector-Jacobian product}
\newacronym{qp}{QP}{quadratic program}
\newacronym{socp}{SOCP}{second order cone programming}
\newacronym{mse}{MSE}{mean squared error}
\newacronym{aid}{AID}{approximate implicit differentiation}
\newacronym{nn}{NN}{neural network}
\newacronym{hcnn}{HCNN}{hard-constrained neural network}

\usepackage{algorithm}
\usepackage{algpseudocode}
\usepackage[framemethod=tikz]{mdframed}

\usepackage{pgfplots}
\usepackage{tikzpeople}
\usepackage{amsmath}
\usepackage{pgfplots}
\pgfplotsset{compat=1.18} 

\usepackage{tcolorbox}
\crefname{figure}{Figure}{Figures}
\crefname{myalgorithm}{Algorithm}{Algorithms}

\usepackage{xifthen}

\newcommand{\ouralg}{$\Pi$\texttt{net}\xspace}

\newcommand{\st}{\,|\,}

\DeclareMathOperator{\argmin}{argmin}

\newcommand{\reals}{\mathbb R}

\newcommand{\mc}[1]{\mathcal{#1}}

\newcommand{\projection}[2]{%
\mathbb{P}_{#1}%
\ifthenelse{\isempty{#2}}{}{\left[#2\right]}
}

\newcommand{\convexSet}[2]{%
\mathcal{C}%
\ifthenelse{\isempty{#2}}{}{_{#2}}%
\ifthenelse{\isempty{#1}}{}{\left(#1\right)}
}

\newcommand{\loss}[2]{
\mathcal{L}_{#2}%
\ifthenelse{\isempty{#1}}{}{\left(#1\right)}
}
\newcommand{\constraintset}[2]{%
\mc{C}%
\ifthenelse{\isempty{#2}}{}{_{#2}}%
\ifthenelse{\isempty{#1}}{}{\left(#1\right)}
}
\newcommand{\objfun}{\varphi}

\newcounter{myalgorithm}
\newenvironment{myalgorithm}[1][]
{	\refstepcounter{myalgorithm}
	\begin{minipage}{\linewidth}
		\medskip
		\hrule
		\smallskip
		\textsc{Algorithm \themyalgorithm}. #1
		\smallskip
		\hrule 
		\smallskip
	\end{minipage}
}
{
	\smallskip
	\hrule width\linewidth\relax
	\smallskip
}

\newcommand{\reviewerOne}[1]{#1}
\newcommand{\reviewerTwo}[1]{#1}
\newcommand{\reviewerThree}[1]{#1}
\newcommand{\reviewerFour}[1]{#1}
\graphicspath{{../}}

\title{
$\mathbf{\Pi}$net: Optimizing hard-constrained neural networks with orthogonal projection layers
}

\author{%
  Panagiotis D.\ Grontas\thanks{Equal contribution.}\\
  ETH Z\"urich\\
  \texttt{pgrontas@ethz.ch} \\
  \And
  Antonio Terpin$^\ast$\\
  ETH Z\"urich\\
  \texttt{aterpin@ethz.ch}
  \And
  Efe C.\ Balta\\
  inspire AG \& ETH Z\"urich\\
  \texttt{efe.balta@inspire.ch}
  \AND
  Raffaello D'Andrea\\
  ETH Z\"urich\\
  \texttt{rdandrea@ethz.ch}
  \And
  John Lygeros\\
  ETH Z\"urich\\
  \texttt{jlygeros@ethz.ch}
}

\iclrfinalcopy 
\begin{document}

\maketitle

\begin{abstract}
We introduce an output layer for neural networks that ensures satisfaction of convex constraints. 
Our approach, \ouralg, leverages operator splitting for rapid and reliable projections in the forward pass, and the implicit function theorem for backpropagation. 
We deploy \ouralg as a \textit{feasible-by-design} optimization proxy for 
parametric constrained optimization problems and obtain modest-accuracy
solutions faster than traditional solvers when solving a single problem, and significantly faster for a batch of problems. 
We surpass state-of-the-art learning approaches by orders of magnitude in terms of training time, solution quality, and robustness to hyperparameter tuning, while maintaining similar inference times. 
Finally, we tackle multi-vehicle motion planning with non-convex trajectory preferences and provide \ouralg as a GPU-ready package implemented in \texttt{JAX}.
\if@submission%
\else%
\vspace{.25cm}
\makebox[\linewidth][c]{%
    \raisebox{-0.2\height}{\includegraphics[height=1em]{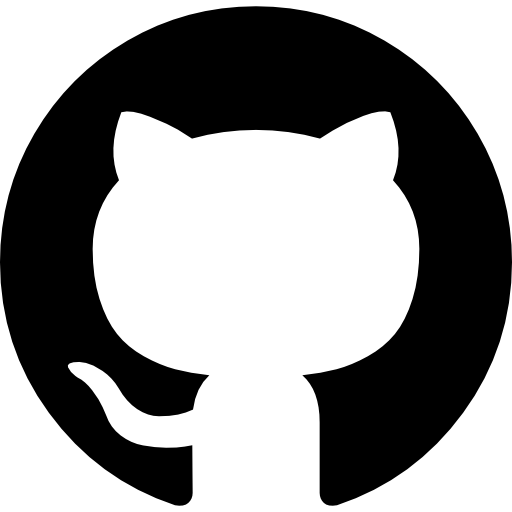}}%
    \hspace{1em}\url{https://github.com/antonioterpin/pinet}%
    \hspace{2em}\raisebox{-0.2\height}{\includegraphics[height=1em]{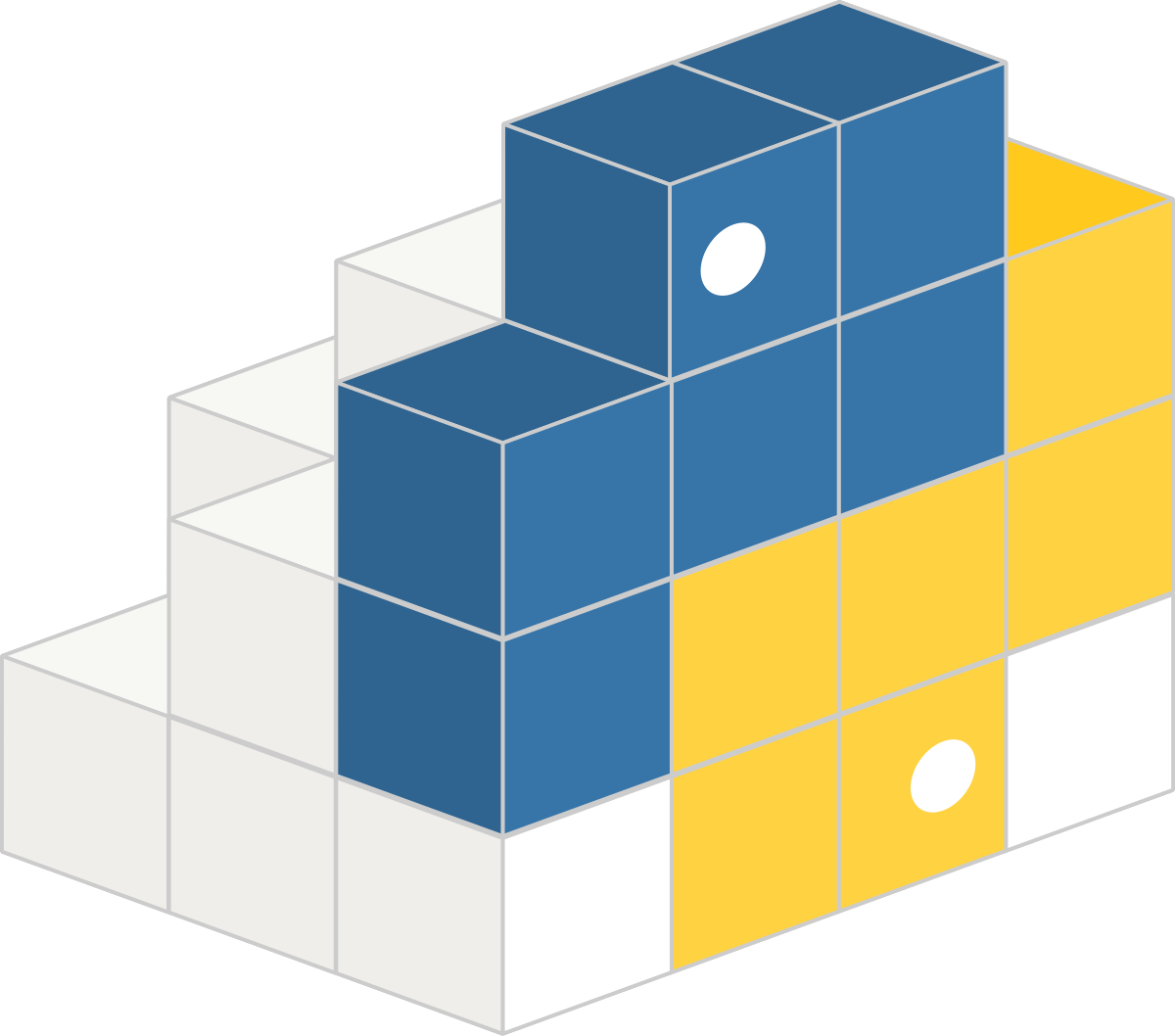}}%
    \hspace{1em}\texttt{pip install pinet-hcnn}%
}
\fi
\end{abstract}
\begin{figure}[h]
    \centering
    \includegraphics[width=0.8\linewidth]{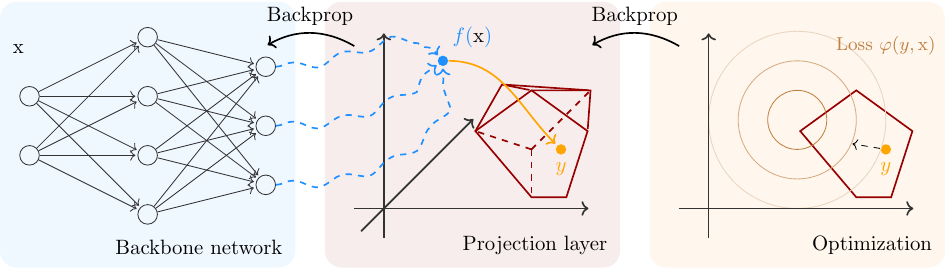}
    \caption{Illustration of the \ouralg architecture. 
    The infeasible output of the \emph{backbone} network is projected onto the feasible set
    through an operator splitting scheme. 
    To train the backbone network,
    we use the implicit function theorem to backpropagate the loss through the projection layer.
    }
    \label{fig:cover}
\end{figure}
\section{Introduction}
In this work, we deploy \glspl*{nn} to generate \textit{feasible-by-design} candidate solutions (see \cref{fig:cover})
for the parametric constrained optimization problem:
\begin{equation}
\label{eq:parametric_constrained_optimization}
    \underset{ y }{\mathrm{minimize}}
    \quad \objfun(y, \mathrm{x})
    \quad
\mathrm{subject~to} \quad
y \in \mathcal{C}(\mathrm{x}),
\tag*{\text{$\mathcal{P}(\mathrm{x})$}} %
\end{equation}%
where $y \in \reals^d$ is the decision variable,
$\mathrm{x} \in \reals^p $ is the context (or parameter) of the problem instance,
$\objfun : \reals^d \times \reals^p \to \reals $ is the objective function, and $ \mathcal{C}(\mathrm{x}) \subseteq \reals^d $ is a non-empty, closed, convex set for all $\mathrm{x}$. We provide a pedagogical example to explain this formulation in \cref{sec:experiments:toy-example}.

Constrained optimization has universal applicability, from safety-critical applications such as the optimal power flow in electrical grids
\citep{nellikkath2022physics}, to logistics and scheduling \citep{bengio2021machine}, 
and even biology, where enforcing priors on the solution can enhance its interpretability \citep{balcerak2025energy,terpin2024learning}. 
In many applications, optimization programs are solved repeatedly, given different contexts: in logistics, the demands and forecasts vary \citep{baptiste2001constraint}; in model predictive control (MPC) \reviewerOne{\citep{chen2018approximating,tabas2022safe}}, the initial conditions;
in motion planning, the position of obstacles \citep{marcucci2023motion}; in trust-region policy optimization, the advantage function and the trust-region center \citep{terpin2022trust}.
This task often becomes computationally challenging when, for example,
$y$ is high-dimensional, $\varphi$ is non-convex, or new solutions are required at a high frequency.
Rather than solving each problem instance from scratch, the mapping from contexts to solutions can be learned with \glspl*{nn}.
Despite their successes, \glspl*{nn} typically lack inherent mechanisms to ensure satisfaction of explicit constraints, a limitation that motivates a large body of existing work:

\paragraph{Soft-constrained NNs.}
Soft penalty terms for constraint violations in the objective function are one approach to incorporate constraints in \glspl*{nn} \citep{marquez2017imposing}, and have been adopted to solve parametric constrained optimization problems \citep{tuor2021neuromancer},
and partial differential equations through physics-informed NNs \citep{erichson2019physics,raissi2019physics}.
Despite their ability to handle general constraints,
these approaches offer no guarantees at inference time.
Beside requiring manual tuning of the penalty parameters, which is challenging yet critical for good performance, the use of soft constraints is discouraged for the following reasons. First, the structure of the constraints set can be exploited to design more efficient algorithms; see, e.g., the simplex algorithm \citep{dantzig2002linear}. Second, treating constraints softly may significantly alter the problem solution regardless of tuning \citep{grontas2024operatorsplittingconvexconstrained}. Third, certain constrained optimization problems (e.g., linear programs) may not admit a solution at all when constraints are treated softly.

\paragraph{Hard-constrained NNs.}
To circumvent the shortcomings of soft constraints, \glspl*{hcnn} aim to enforce constraints
on the \gls*{nn} output by design.
\citet{frerix2020homogeneous} address linear homogeneous inequality constraints 
by parameterizing the feasible set.
Similarly, RAYEN \citep{tordesillas2023rayen} enforces various convex constraints by scaling the line segment between infeasible points and a fixed point in the feasible set's interior.
While these methods enjoy rapid inference, they require expensive offline preprocessing
and are not directly applicable to constraints that depend on the NN input, i.e.,
they consider feasible sets $\mathcal{C}$ and not $\mathcal{C}(\mathrm{x})$ in \ref{eq:parametric_constrained_optimization}.
Differently, \cite{min2024hard} propose a closed-form expression to recover feasibility
for polyhedral constraints and employs \texttt{cvxpylayers} \citep{agrawal2019differentiable}
for more general convex sets.
\citet{chen2018approximating,cristian2023end}
orthogonally project the \gls*{nn} output or intermediate layers using Dykstra's algorithm \citep{boyle1986method},
but rely on loop unrolling for backpropagation, which can be prohibitive in
terms of memory and computation.
Departing from convex sets, \texttt{DC3} \citep{donti2021dc3} introduces an equality completion
and inequality correction procedure akin to soft-constrained approaches, 
but applied during inference.
\citet{lastrucci2025enforce} impose non-linear equality constraints
by recursively linearizing them.
Lagrangian and augmented Lagrangian approaches are considered by \citet{fioretto2021lagrangian,park2023self} for general
non-convex constraints, with drawbacks similar to soft constraints\reviewerOne{, whereas \citet{kratsios2021universal} consider a (probabilistic) sampling approach}. Recently, LinSATNet \citep{wang2023linsatnet} has been proposed to impose non-negative linear constraints, which is a restrictive constraint class that renders none of the problems of interest for this work amenable to LinSATNet. 
This limitation is partially relaxed by GLinSAT \citep{zeng2024glinsatgenerallinearsatisfiability},
which however requires bounded constraints, an assumption not satisfied by, e.g., epigraph reformulations \citep[Appendix A.5-A.7]{Stellato_2020}.

\paragraph{Implicit layers.}
\textit{Implicit layers} embed optimization problems such as \glspl*{qp} \citep{amos2017optnet, butler2023efficient}, conic programs \citep{agrawal2019differentiable}, non-linear least-squares \citep{pineda2022theseus},
or fixed-point equations \citep{bai2019deep, winston2020monotone} as \gls*{nn} layers, and apply the implicit function theorem \citep{dontchev2009implicit} to various measures of optimality. Our work is an instance of implicit layers; see \cref{sec:pinet-implicit-layer}.

\paragraph{ML for optimization.}
Using constrained \gls*{nn} architectures to learn solution mappings is among many successful efforts to exploit ML techniques to accelerate \citep{king2024metric, sambharya2024learning}
or replace optimization solvers \citep{bertsimas2022online, zamzam2020learning}.
These approaches are referred to as \textit{amortized optimization}, \textit{learning to optimize} or \textit{optimization learning}, and
the surveys \citep{amos2023tutorial,van2025optimization} cover multiple aspects of the topic.

\begin{mdframed}[hidealllines=true,backgroundcolor=blue!5]
\paragraph{Contributions.}
We propose an \gls*{nn} architecture, \ouralg, that generates feasible-by-design solutions for \ref{eq:parametric_constrained_optimization}.
Given a context $\mathrm{x}$, we deploy a \emph{backbone} \gls*{nn} to produce a
raw output $y_\text{raw}$ that we
orthogonally project onto $\mathcal{C}(\mathrm{x})$,
$y =\argmin_{z \in \mathcal{C}(\mathrm{x})} \norm{z - y_{\text{raw}}}^2.$
In particular:
\begin{itemize}[leftmargin=1em]
    \item We use an operator splitting scheme to compute the projection in the forward pass,
    and backpropagate through it via the implicit function theorem.
    Our work is an instance of implicit layers (see \cref{sec:pinet-implicit-layer}), but it specializes in projection problems whose structure can be significantly exploited,
    achieving rapid training and improved inference speed. Our simple idea yields a learning architecture that surpasses state-of-the-art learning approaches by orders of magnitude in terms of training time, solution quality, and robustness to hyperparameter tuning, while maintaining similar inference times. 
    \item We implement a hyperparameter tuning and matrix equilibration strategy that boosts \ouralg's performance
    and robustifies it against data scaling.
    As a result, \ouralg solves challenging benchmarks on which existing methods struggle,
    and is substantially less sensitive to hyperparameter tuning. 
    In addition, we deploy \ouralg on a real-world application in multi-vehicle motion planning with
    non-convex trajectory costs. %
    \item 
    We provide an efficient and GPU-ready implementation of \ouralg in \texttt{JAX}.
    We make our code available
    \if@submission%
    in the supplementary material.%
    \else
    at \url{https://github.com/antonioterpin/pinet}.%
    \fi
\end{itemize}

\end{mdframed}

\paragraph{Notation.}
The indicator function of a set $\mathcal{K}$ is $\mathcal{I}_{\mathcal{K}}(y) = 0$ if \( y \in \mathcal{K} \),
and $+\infty$ otherwise.
The proximal operator of a proper, closed and convex function $f$ with parameter $\sigma > 0$ is $\mathrm{prox}_{\sigma f}(x) = \argmin_y \{f(y) + \frac{1}{2 \sigma} \norm{y-x}^2 \}$.
The projection onto a non-empty, closed, and convex set $\mathcal{C}$ is $\Pi_{\mathcal{C}}(x) = \mathrm{prox}_{\mathcal{I}_{\mathcal{C}}}(x) $.
For a differentiable mapping $F : \reals^n \to \reals^m$, its Jacobian evaluated at $\hat{x}\in \reals^n$ is denoted by 
$\frac{\partial F(x)}{\partial x} \big|_{x=\hat{x}} \in \reals^{m \times n}$;
we use $\mathrm{d}$ instead of $\partial $ when referring to total derivatives.
Given a vector $v \in \reals^m$, the \gls*{vjp} of $F$ at $\hat{x}$ with $v$ is
$v^{\top} \frac{\partial F(x)}{\partial x} \big|_{x=\hat{x}} \in \reals^n$.
\section{Training Hard-Constrained Neural Networks}
We develop an \gls*{nn} layer that projects the output of
any backbone \gls*{nn} onto $\mathcal{C}(\mathrm{x})$, and discuss how 
\ouralg
can be trained to generate solutions of \ref{eq:parametric_constrained_optimization},
for any given $\mathrm{x}$.
\subsection{Projection layer}
\label{sec:projection-layer}
Given a context $\mathrm{x}$, the backbone network produces the raw output 
\(y_{\text{raw}} = f(\mathrm{x}; \theta)\), where $\theta$ are the network weights.
We enforce feasibility by projecting $y_\text{raw}$ onto \(\mathcal{C}(\mathrm{x})\):
\begin{equation}
\label{eq:orthogonal-projection}
    y = \Pi_{\mathcal{C}(\mathrm{x})}\bigl(y_{\text{raw}}\bigr) = {\argmin}_{z \in \mathcal{C}(\mathrm{x})} \, \bigl\| z - y_\text{raw} \bigr\|^2.
\end{equation}

We highlight two benefits of this approach:
\begin{itemize}[leftmargin=*]
    \item[\ding{51}] \textbf{Constraint satisfaction.}
    By design, the output \(y\) of the projection layer always lies in \(\mathcal{C}(x)\).
    
    \item[\ding{51}] \textbf{Decomposition of specifications.} 
    The hard constraints prescribe the \emph{required} behavior of the output,
    while the objective prescribes the \emph{desired} behavior. 
    Contrary to soft-constrained \glspl*{nn}, \emph{no tradeoff between the two behaviors is introduced in our proposed framework}.

\end{itemize}

We consider constraints that can be expressed as 
$\mathcal{C} = \Pi_d(\mathcal{A} \cap \mathcal{K})$,
where $\mathcal{A}, \mathcal{K} \subseteq \reals^n,\, n \geq d, $ are closed, convex sets that we design,
and $\Pi_d$ is the projection onto the first $d$ coordinates.
We omit the dependence on $\mathrm{x}$ for brevity, but stress that our method readily handles
context-dependent constraints.
Notice that we work in the possibly higher-dimensional $\reals^n$, by introducing
an auxiliary variable $y_\text{aux} \in \reals^{n-d}$.
This provides us the flexibility to choose $\mathcal{A}$ and $\mathcal{K}$ such that their respective projections
$\Pi_{\mathcal{A}}$ and $\Pi_{\mathcal{K}}$
admit a closed-form expression or are numerically-efficient. 
In particular, $\mathcal{A}$ will be a hyperplane defined by the coefficient matrix $A$ and the offset vector $b$, and
$\mathcal{K}$ a Cartesian product of the form
$\mathcal{K} = \mathcal{K}_1 \times \mathcal{K}_2 \subseteq \reals^d \times \reals^{n - d}$.
This representation can describe many constraints of practical interest
and we instantiate it with an example next (see \cref{appendix:extra-sets} for more).
Consider polytopic sets that are often employed in robotics \citep{chen2018approximating}, numerical solutions to \gls*{pde} \citep{raissi2019physics}, and non-convex relaxations for trajectory planning \citep{malyuta2022convex}, among others. They are expressed as
$
\{ y \in \reals^d \st  E y= q, l \leq Cy \leq u \},
$
for some $E, q, l, C, u$ of appropriate dimensions. 
We introduce the auxiliary variable $y_{\text{aux}} = Cy \in \reals^{n_\text{ineq}}$
with dimension $n-d = n_\text{ineq}$.
Then, we define $\mathcal{A}, \, \mathcal{K} \subseteq \reals^{n}$ as the following hyperplane and box
\begin{align*}
\mathcal{A} & = \bigg\{
\begin{bmatrix}
    y\\
    y_{\text{aux}}
\end{bmatrix}
\, \bigg| \,
\underbrace{
\begin{bmatrix}
    E & 0\\C & -I
\end{bmatrix}
}_{=A}
\begin{bmatrix}
    y \\ y_\text{aux}
\end{bmatrix} 
=
\underbrace{
\begin{bmatrix}
    q \\ 0
\end{bmatrix}
}_{=b}
\bigg\},
\quad
\mathcal{K} = \bigg\{ \begin{bmatrix}
    y\\
    y_{\text{aux}}
\end{bmatrix} \, \bigg| \,\,
y \in \reals^d, ~ l \leq y_\text{aux} \leq u
\bigg\}.
\end{align*}
Importantly, $\mathcal{C} = \Pi_d(\mathcal{A} \cap \mathcal{K})$ and both $\Pi_{\mathcal{A}}$ and $\Pi_{\mathcal{K}}$ can be evaluated in closed form. 
\reviewerOne{More generally, the proposed decomposition, with $\Pi_{\mathcal{A}}$ and $\Pi_{\mathcal{K}}$ admitting closed-form expressions or being numerically efficient,
is applicable to any constraint set of the form:
\begin{equation}
\begin{aligned}
&A_\text{eq} y = b_\text{eq} \quad &(\text{Equality}) \\
&\ell_\text{box} \leq y \leq u_\text{box} \quad &(\text{Box}) \\
&\ell_\text{ineq} \leq A_\text{ineq} y \leq u_\text{ineq} \quad &(\text{Inequality}) \\
&\lVert A_{\text{soc}, i} y + a_{\text{soc,} i} \rVert_2 \leq f_{\text{soc}, i}^{\top} y + b_{\text{soc}, i}, \quad i = 1, \ldots, N_\text{soc} \quad &(\text{SOC}) \\
&\lVert A_{\ell_1, i} y + a_{\ell_1, i} \rVert_1 \leq f_{\ell_1, i}^{\top} y + b_{\ell_1, i}, \quad i = 1, \ldots, N_{\ell_1} \quad &(\ell_1-\text{ball}) \\
&\lVert A_{\ell_\infty, i} y + a_{\ell_\infty, i} \rVert_\infty \leq f_{\ell_\infty, i}^{\top} y + b_{\ell_\infty, i}, \quad i = 1, \ldots, N_{\ell_\infty} \quad &(\ell_\infty-\text{ball}) \\
&y_{\mathcal{S}_j} \geq 0, \quad \lVert y_{\mathcal{S}_j} \rVert_1 = 1, \quad \text{for index sets }\mathcal{S}_j. \quad &(\text{Simplex})
\end{aligned}
\label{eq:constraints-examples}
\end{equation}}

\reviewerOne{
\begin{remark}
    We stress that the decomposition $\mathcal{C} = \Pi_d(\mathcal{A} \cap \mathcal{K})$ 
    is not an assumption.
    One can always decompose a convex set in this way, e.g., by considering the trivial decomposition 
    $\mathcal{A} = \mathcal{C}$ and $\mathcal{K} = \mathbb{R}^d$.
    Instead, determining $\mathcal{A}$ and $\mathcal{K}$ is a design choice which we leverage to make the
    projections $\Pi_\mathcal{A}$ and $\Pi_\mathcal{K}$ computationally efficient.
    We note two important points regarding this design choice:
    \begin{itemize}[leftmargin=*]
    \item The only assumption is that $\Pi_\mathcal{A}$ and $\Pi_\mathcal{K}$ and their \gls*{vjp} are computable.
    Being computationally efficient is an added benefit of our decomposition, but is not necessary. In this work, our focus is on the sets listed as those are the ones that are most interesting in practice.
    \item We show in \cref{appendix:extra-sets} that many practically-relevant constraints $\mathcal{C}$, such as the ones listed in \eqref{eq:constraints-examples}, as well as their intersections and Cartesian products
    all admit efficient decompositions.
    In fact, this list is not exhaustive; see, e.g., the work of \citet{condat2016fast} and \citet{boyd2004convex}.
    \end{itemize}
\end{remark}}

\subsubsection{Forward pass}
\label{sec:forward-pass}
To compute the projection $y = \Pi_{\mathcal{C}(\mathrm{x})}(y_{\text{raw}})$ we employ the Douglas-Rachford algorithm \citep[Sec.\ 28.3]{bauschke_convex_2017},
which solves optimization problems of the form 
$\min_z \, g(z) + h(z) $,
where $g$ and $h$ are proper, closed, convex functions.
To rewrite \eqref{eq:orthogonal-projection} in this composite form, we use the auxiliary variable $y_{\text{aux}} \in \reals^{n-d}$ and
the indicator function of $\mathcal{A}$ and $\mathcal{K}$ to obtain: 
\begin{equation} \label{eq:projection_in_lifted_form}
    (\Pi_{\mathcal{C}}(y_{\text{raw}}), y_{\text{aux}}^{\star}) = \underset{ y, y_{\text{aux}} }{\mathrm{argmin}}
    \bigg\{\norm{y - y_{\text{raw}}}^2
        + \mathcal{I}_{\mathcal{A}}\left(\begin{bmatrix}
    y\\y_{\text{aux}}
\end{bmatrix}\right)
+
\mathcal{I}_{\mathcal{K}}\left(
\begin{bmatrix}
    y\\y_{\text{aux}}
\end{bmatrix}
\right)
\bigg\}.
\end{equation}%
Then, we split the objective function as follows
\[
g\left(\begin{bmatrix}
    y\\y_{\text{aux}}
\end{bmatrix}\right)
=
\mathcal{I}_{\mathcal{A}}\left(\begin{bmatrix}
    y\\y_{\text{aux}}
\end{bmatrix}\right)
\quad\text{and}\quad
h\left(\begin{bmatrix}
    y\\y_{\text{aux}}
\end{bmatrix}\right) = \norm{y - y_{\text{raw}}}^2
    + \mathcal{I}_{\mathcal{K}}\left(\begin{bmatrix}
    y\\y_{\text{aux}}
\end{bmatrix}\right).
\]

By applying the Douglas-Rachford algorithm we obtain the fixed-point iteration:

\vspace{-.4cm}
\begin{minipage}{0.45\linewidth}
\begin{subequations} \label{eq:dra_for_projection}
\begin{align}
z_{k + 1} &= \mathrm{prox}_{\sigma g}(s_k)
\label{eq:equality_constraint}
\\
t_{k + 1} &= \mathrm{prox}_{\sigma h}(2z_{k + 1} - s_k)
\label{eq:inequality_projection}
\\
s_{k + 1} &=  s_k + \omega (t_{k + 1} - z_{k + 1})
\label{eq:dra_governing_update}
\end{align}
\end{subequations}
where $\sigma > 0$ is a scaling and $\omega\in(0,2)$ a relaxation parameter.
The proximal operators in
\eqref{eq:equality_constraint} and \eqref{eq:inequality_projection}
can be evaluated explicitly,
see \cref{appendix:extra-calculations:inequality_projection},
allowing us to implement \eqref{eq:dra_for_projection} as in \cref{alg:os-project}.
Note that we write $z_k = \begin{bmatrix}
    z_{k,1} & z_{k,2}
\end{bmatrix}^{\top}$
where $z_{k,1} \in \reals^d$ and $ z_{k,2} \in \reals^{n-d}$
correspond to $y$ and $y_\text{aux}$.
\end{minipage}
\hfill%
\begin{minipage}{0.5\linewidth}
    \begin{myalgorithm}[Operator splitting for projection] \label{alg:os-project}%
        \textbf{Inputs:} $ \mathrm{x}, y_\text{raw}, K \in \mathbb{N}, \sigma , \omega $. \\
        \textbf{Initialization:} $s_0 \in \reals^{n}$ \\
        \textbf{For $k = 0$ to $K - 1$:} \\
        $
        \left\lfloor
        \begin{array}{l}
               z_{k + 1} \leftarrow \Pi_{\mathcal{A}(\mathrm{x})}(s_k) \\
               t_{k + 1} \leftarrow 
               \begin{bmatrix} 
               \Pi_{\mathcal{K}_1 (\mathrm{x})}(\frac{2z_{k + 1,1} - s_{k,1} + 2\sigma y_{\text{raw}}}{1 + 2\sigma}) \\
                \Pi_{\mathcal{K}_2 (\mathrm{x})}(2z_{k + 1,2} - s_{k,2}) \\
            \end{bmatrix} \\
             s_{k + 1} \leftarrow s_k + \omega (t_{k + 1} - z_{k + 1})
        \end{array}
        \right.
        $ \\[.3em]
        \textbf{Output:} $z_{K,1}, s_K$
    \end{myalgorithm}%
\end{minipage}

Under mild conditions, namely strict feasibility of \eqref{eq:projection_in_lifted_form}, we show in \cref{appendix:extra-calculations:convergence_of_dra} that the iterates
$z_k$ and $t_k$ converge to a solution of \eqref{eq:projection_in_lifted_form}.
We denote the limits $z_{\infty}(y_\text{raw}) = \lim_{k\to\infty} z_k$ and $s_{\infty}(y_\text{raw}) = \lim_{k \to \infty} s_k$,
and highlight their dependence on the point-to-be-projected $y_\text{raw}$.
In particular, we have $ z_{\infty,1}(y_\text{raw}) = \Pi_{\mathcal{C}}(y_{\text{raw}}) $.
In practice, we run a finite number of iterations $K \in \mathbb{N}$ of \eqref{eq:dra_for_projection}, which we set
to $K = \texttt{n\_iter\_fwd}$ during training and  $K = \texttt{n\_iter\_test}$ during testing,
and take $y = z_{K,1}$ as the output of the projection layer.
We detail our hyperparameters in \cref{subsec:the_sharp_bits_short_version}.
We note that, although $z_{K}$ will not necessarily lie on $\mathcal{A}\cap\mathcal{K}$, because $K \in \mathbb{N}$ is finite, 
\cref{alg:os-project} guarantees that $z_{K} \in \mathcal{A}$.
By our choice of $\mathcal{A}$, this implies that $z_{K,1}$ satisfies any equality constraints in the problem.
The feasibility-by-design comes from the convergence rates, which we derive and empirically demonstrate in \cref{appendix:convergence_of_dra:rates}: for a sufficiently high number of iterations, the output of our projection layer is arbitrarily close to the true projection.

\subsubsection{Backward pass}
\label{sec:backward-pass}
To train the backbone network using backpropagation, we need to efficiently differentiate the loss $\mathcal{L}$ 
(which in general depends on the projected output of the network and on the input;
see \cref{subsec:loss}) with respect to the backbone network parameters $\theta$.
The chain rule gives us:
\begin{equation} \label{eq:chain_rule_backpropagation}
\frac{\mathrm{d} \mathcal{L}(\Pi_{\mathcal{C}}(f(\mathrm{x};\theta)), \mathrm{x})}{\mathrm{d}\theta}
=
\frac{\partial \mathcal{L}(y, \mathrm{x})}{\partial y}\bigg|_{y = \Pi_{\mathcal{C}}(f(\mathrm{x};\theta))}
\frac{\partial\Pi_{\mathcal{C}}(y_{\text{raw}})}{\partial y_{\text{raw}}}\bigg|_{y_{\text{raw}} = f(\mathrm{x};\theta)}\frac{\partial f(\mathrm{x}; \vartheta)}{\partial \vartheta} \bigg|_{\vartheta = \theta}.
\end{equation}
Since the first and last terms are standard and typically computed with automatic differentiation, we only need to provide an efficient computational routine for the \gls*{vjp}
\begin{equation} \label{eq:projection_vjp}
v\mapsto v^\top\left(
\frac{\partial\Pi_{\mathcal{C}}(y_{\text{raw}})}{\partial y_{\text{raw}}}\bigg|_{y_{\text{raw}} = f(\mathrm{x};\theta)}\right).
\end{equation}

Rather than differentiating through all the iterations of \cref{alg:os-project} by loop unrolling, 
we exploit the implicit function theorem \citep{dontchev2009implicit} to efficiently evaluate \eqref{eq:projection_vjp}. 
To do so, we first recall that 
$\Pi_{\mathcal{C}}(y_{\text{raw}}) = \begin{bmatrix} I_d & \mathbf{0}_{d \times (n - d)} \end{bmatrix} \Pi_{\mathcal{A}}(s_{\infty}(y_{\text{raw}}))$ which implies that:
\begin{equation} \label{eq:projection_vjp_expanded}
v^\top\left(
\frac{\partial\Pi_{\mathcal{C}}(y_{\text{raw}})}{\partial y_{\text{raw}}}\bigg|_{y_{\text{raw}} = f(\mathrm{x};\theta)}\right) = 
\left( \begin{bmatrix} v^\top & \mathbf{0} \end{bmatrix} \frac{\partial \Pi_{\mathcal{A}}(s) }{\partial s} \bigg|_{s = s_\infty(y_{\text{raw}})} \right)
\frac{\partial s_\infty(y_\text{raw})}{\partial y_\text{raw}} \bigg|_{y_{\text{raw}} = f(\mathrm{x}; \theta)},
\end{equation}
and note that the first \gls*{vjp} is straightforward to compute since $\Pi_{\mathcal{A}} $ is an affine mapping.
Therefore, to evaluate \eqref{eq:projection_vjp_expanded} we need to backpropagate through
$s_{\infty}(y_\text{raw})$.
Since $s_{\infty}(y_\text{raw})$ is the fixed point of iteration \eqref{eq:dra_for_projection},
it satisfies the equation
$
s_{\infty}(y_{\text{raw}}) = \Phi(s_{\infty}(y_{\text{raw}}), y_{\text{raw}}),
$
where $\Phi$ represents one iteration of \eqref{eq:equality_constraint}-\eqref{eq:dra_governing_update}, namely, $s_{k+1} = \Phi(s_{k}, y_{\text{raw}})$.
The implicit function theorem applied to $s_{\infty}(y_{\text{raw}}) = \Phi(s_{\infty}(y_{\text{raw}}), y_{\text{raw}})$ yields the \gls*{vjp}, see \cref{appendix:extra-calculations:derivation_of_backward},
\begin{equation}
    \label{eq:vjp}
v \mapsto  \xi(y_{\text{raw}}, v)^\top \frac{\partial \Phi(s_\infty(y_\text{raw}), y_{\text{raw}})}{\partial y_{\text{raw}}}\bigg|_{y_\text{raw} = f(\mathrm{x};\theta)} ,
\end{equation}
where $\xi(y_{\text{raw}}, v)$ is a solution of the linear system
\begin{equation}
\label{eq:linear-system}
\left(I - \frac{\partial \Phi(s, y_{\text{raw}})}{\partial s}\bigg|_{ s = s_\infty(y_{\text{raw}})}\right)^\top \xi(y_{\text{raw}}, v) 
=
v.
\end{equation}
The matrix in \eqref{eq:linear-system} may not be invertible \citep{agrawal2019differentiable}; 
even if it is, constructing it and computing its inverse may be prohibitively expensive in high dimensions.
This difficulty can be circumvented by computing a heuristic quantity.
In this vein, we deploy the \texttt{JAX} \citep{jax2018github} implementation of the bi-conjugate gradient stable iteration (\texttt{bicgstab}, \citet{bicgstab}),
an indirect linear system solver that requires only matrix-vector products.
Therefore, we implement \eqref{eq:vjp} and \eqref{eq:linear-system} using \glspl*{vjp} involving
$\partial \Phi(s_\infty(y_\text{raw}), y_{\text{raw}})/\partial y_{\text{raw}}$ and $\partial \Phi(s, y_{\text{raw}})/\partial s$, respectively.
We efficiently do this using \texttt{JAX} \gls*{vjp} routines and note that each step of the solver for \eqref{eq:linear-system}
has essentially the same computational cost as one step of \eqref{eq:dra_for_projection}.
The maximum number of \texttt{bicgstab} steps for \eqref{eq:linear-system}, \texttt{n\_iter\_bwd}, is a hyperparameter.
In both \eqref{eq:vjp} and \eqref{eq:linear-system}, we use the output of the forward pass, i.e., 
$s_{K}$ in place of the intractable $s_\infty$. 
\reviewerThree{We discuss almost-everywhere differentiability of our projection layer and the applicability of the implicit function theorem in \cref{appendix:theory}.}

\subsection{Loss} \label{subsec:loss}

The training loss $\mathcal{L}\big( \Pi_{\mathcal{C}(\mathrm{x})}(f(\mathrm{x}; \theta)), \mathrm{x} \big) $,
which is a function of the context and the constrained \gls*{nn} output, can be crafted according to the specific requirements of the problem. 
In our experiments, we directly minimize the objective of \ref{eq:parametric_constrained_optimization} using the network’s output by
setting $\mathcal{L}(y, \mathrm{x}) = \objfun(y, \mathrm{x})$.
In our framework we can, loosely speaking, interpret training as performing 
projected gradient descent on the raw \gls*{nn} output space, akin to the work of \citet{cristian2023end}. 

    \subsection{Network architecture}
    \ouralg is flexible: the projection layer can be appended to \emph{any} \gls*{nn}; 
    see \cref{fig:cover}. 
    We summarize the forward and backward pass of \ouralg in \autoref{alg:pinet},
    while its training and testing is outlined in \autoref{alg:training-testing}.
    
\begin{minipage}[t]{0.47 \linewidth}
    \begin{myalgorithm}[\ouralg \vphantom{Training/Testing}] \label{alg:pinet}%
        \texttt{Forward:} \\[.3em]
        $
        \left|
            \begin{array}{l}
                \textbf{Inputs: } \mathrm{x}, \theta, K, \sigma , \omega \\
                y_\text{raw} \leftarrow f(\mathrm{x}; \theta) \\
                y, s \leftarrow \text{Call \cref{alg:os-project}}(\mathrm{x}, y_\text{raw}, K, \sigma, \omega) \\
                \textbf{Output: } y, s
            \end{array}
        \right.
        $ \\[.3em]
        \texttt{Backward:} \\[.3em]
        $
        \left|
            \begin{array}{l}
                \textbf{Inputs: } s_\infty, v \in \reals^d \\
                v_1 \leftarrow \begin{bmatrix} v^\top & \mathbf{0} \end{bmatrix} (\partial \Pi_{\mathcal{A}}(s) / \partial s)  \big|_{s = s_\infty } \\
                \xi(y_\text{raw}, v_1) \leftarrow \text{Solve \eqref{eq:linear-system} with $v_1$ as RHS} \\
                v_2 \leftarrow 
                \xi(y_{\text{raw}}, v_1)^\top (\partial \Phi(s, y_{\text{raw}}) / \partial y_{\text{raw}}) \big|_{s = s_\infty} \\
                v_3 \leftarrow 
                v_2^{\top} (\partial f(\mathrm{x}; \vartheta) / \partial \vartheta) \big|_{\vartheta = \theta} \\
                \textbf{Output: } v_3
            \end{array}
        \right.
        $
    \end{myalgorithm}  
\end{minipage}
\hspace{1em}
\vrule width 0.4pt  
\hspace{1em}
\begin{minipage}[t]{0.47 \linewidth}
  \begin{myalgorithm}[Training/Testing of \ouralg] \label{alg:training-testing}%
        \textbf{Inputs}:
            \text{Chosen/Tuned hyperparameters} \\
            \hphantom{\textbf{Inputs}:}
            (see \cref{subsec:the_sharp_bits_short_version}) \\
            \texttt{Training}: \\
            $
            \left| \\
            \begin{array}{l}
                \theta_1 \leftarrow \text{Initialize weights} \\
                \textbf{For $\ell = 1$ to \texttt{n\_epochs}:} \\
                \left\lfloor
                \begin{array}{l}
                    \mathrm{x} \leftarrow \text{Sample from training set} \\
                    y, s \leftarrow \texttt{Forward}(\mathrm{x}, \theta, \texttt{n\_iter\_fwd}, \sigma, \omega) \\
                    \mathcal{L}(y, \mathrm{x}), 
                    \partial \mathcal{L}(y, \mathrm{x}) / \partial y \leftarrow \text{Compute loss} \\
                    \partial \mathcal{L} / \partial \theta \leftarrow 
                    \texttt{Backward}(s, \partial \mathcal{L}(y, \mathrm{x}) / \partial y) \\
                    \theta_{\ell+1} \leftarrow \text{Optimizer update}
                \end{array}
                \right. \\[.5em]
                \textbf{Output}: \theta_{\texttt{n\_epochs}}
            \end{array} \\[.5em]
            \right. 
            $
            \texttt{Testing}: \\
            $
            \left| \\
        \begin{array}{l}
            \mathrm{x} \leftarrow \text{Sample from test set} \\
            y, s \leftarrow \texttt{Forward}(\mathrm{x}, \theta, \texttt{n\_iter\_test}, \sigma, \omega) \\
            \textbf{Output}: y
        \end{array}
        \right.
        $
  \end{myalgorithm}
\end{minipage}

\subsection{The sharp bits (short version)} \label{subsec:the_sharp_bits_short_version}
To push the performance of \ouralg, we adopt two important numerical techniques. 
First, we improve the conditioning of the matrix $A(\mathrm{x})$, that defines the set $\mathcal{A}(\mathrm{x})$, by implementing the 
Ruiz equilibration algorithm \citep{wathen2015preconditioning}; see \cref{sec:equilibration}. 
Second, we exploit the fact that, compared to existing methods, \ouralg relies only on a few hyperparameters. Specifically, \texttt{n\_iter\_fwd} (number of iterations for the forward pass during training), \texttt{n\_iter\_test} (number of iterations for the forward pass during inference), \texttt{omega}, \texttt{sigma} (standard Douglas-Rachford parameters), and \texttt{n\_iter\_bwd} (number of iterations in the \texttt{bicgstab} procedure).
We describe an auto-tuning procedure that recommends hyperparameters by evaluating the projection
on a subset of the validation set in \cref{sec:autotuning}, and assess its effectiveness in \cref{sec:the-sharp-bits:ablation}.
Finally, we highlight our choice of enforcing constraints during training, as opposed to training an unconstrained network and introducing the projection layer only afterwards. We do so because the latter approach may result in training instabilities or suboptimal performance; see \cref{sec:constraints-training}.

\section{Numerical experiments}
\label{sec:experiments}
The code is made available %
\if@submission%
in the supplementary material.
\else
at \url{https://github.com/antonioterpin/pinet}.
\fi 
The empirical data was collected on an Ubuntu 22.04 machine equipped with an AMD Ryzen Threadripper PRO 5995WX processor and an Nvidia RTX 4090 GPU. %
For experiments with second-order cone constraints, see \cref{sec:soc}.

\subsection{Benchmarks and comparisons with state-of-the-art}
\label{sec:experiments:benchmarks}
\begin{figure}
    \centering
    \begin{minipage}{\linewidth}
    \centering
        \includegraphics[width=.90\linewidth]{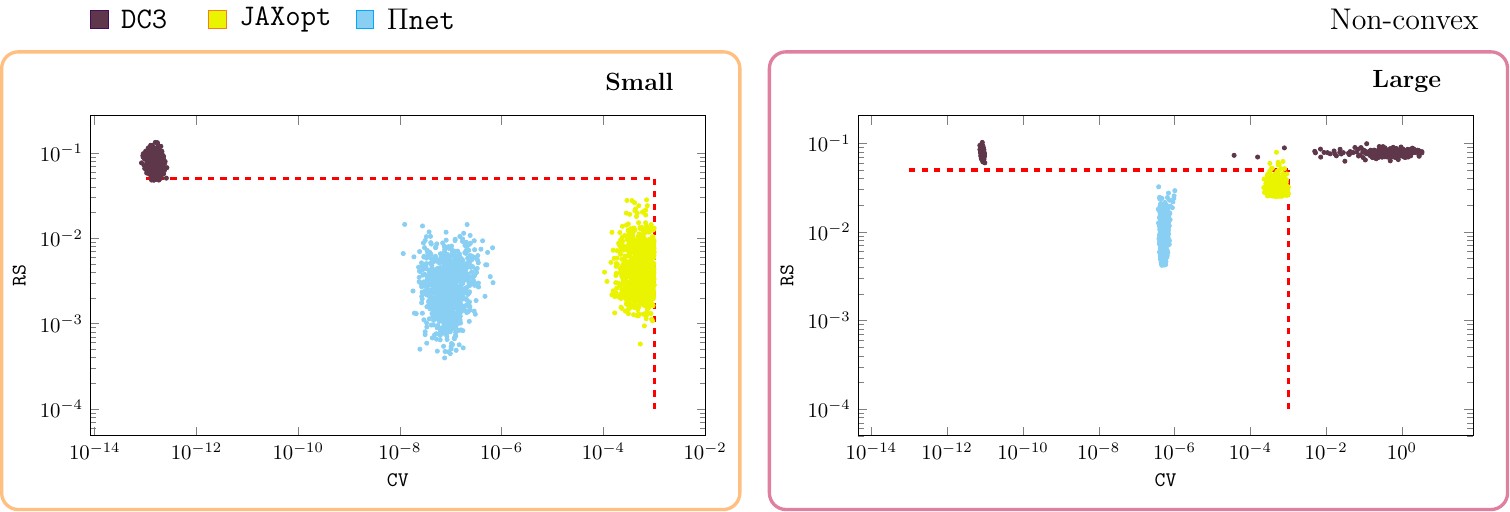}%
    \end{minipage}
    \caption{
    Scatter plots of \texttt{RS} and \texttt{CV} on the small and large non-convex problems on the test set. 
    The red dashed lines show the thresholds to consider a candidate solution optimal. 
    }
    \label{fig:benchmark:non-convex}
\end{figure}
We consider a set of standard convex and non-convex problems
classically used to compare \glspl{hcnn}.
\paragraph{Experimental setup.} We consider the 
benchmark problems introduced by \cite{donti2021dc3}:
\begin{equation*}\label{eq:dc3_benchmark}
    \underset{ y \in \reals^{d}}{\mathclap{\mathrm{minimize}}} 
    \qquad J(y)
    \qquad
\overset{\hphantom{ y \in \reals^{d}}}{\mathclap{\mathrm{subject~to}}} \qquad
A y = \mathrm{x}, \quad C y \leq u,
\end{equation*}%
where $J(\cdot)$ is defined as $J(y) = y^\top Q y + q^\top y$ for the convex problem setup, and as $J(y) = y^\top Q y + q^\top \sin(y)$ for the non-convex,
and \( Q \in \reals^{d \times d} \succ 0,\, q \in \reals^d,\, A \in \reals^{n_{\mathrm{eq}} \times d},\, \mathrm{x} \in \reals^{n_\mathrm{eq}},\, C \in \reals^{n_{\mathrm{ineq}} \times d},\, u \in \reals^{n_\mathrm{ineq}} \).
In particular, in the work of \citet{donti2021dc3}, \( Q \) is diagonal with positive entries, \( A, C, u \) are fixed matrices/vector and the 
contexts $\mathrm{x}$ are generated so that all problem instances are guaranteed to be feasible. 
\citet{donti2021dc3} consider problems with \( d = 100 \).
Here, we include larger problem dimensions, $(d, n_{\mathrm{eq}}, n_{\mathrm{ineq}})  \in \{(100, 50, 50), (1000, 500, 500)\}$,
which we generated with the same scheme. 
We refer to these datasets as small and large.
For each dataset, we generated $10000$ contexts split as $7952/1024/1024$ among training/validation/test sets.

\paragraph{Baselines.} 
We compare \ouralg to \texttt{DC3} \citep{donti2021dc3} and a traditional \texttt{Solver}. For
the convex objective the \texttt{Solver} is the \gls{qp} solver \texttt{OSQP} \citep{Stellato_2020},
while for the non-convex objective is \texttt{IPOPT} \citep{wachter2006implementation}.
Further, we compare to an implicit layer approach, where instead of computing the projection
\eqref{eq:orthogonal-projection} with \cref{alg:os-project},
we use the \texttt{JAXopt} \citep{blondel2022efficient} GPU-friendly implementation of \texttt{OSQP}
that employs implicit differentiation.
Both \ouralg, \texttt{DC3}, and the \texttt{JAXopt} approach use a self-supervised loss (i.e., $\mathcal{L} = J$) and as backbone 
a \gls{mlp} with 2 hidden layers of 200 neurons each and \texttt{ReLU} activations.
Additionally, \texttt{DC3} includes batch normalization and drop out, as well as soft penalty terms in the loss.
We use the default parameters of \texttt{DC3} unless otherwise stated. In particular, the \texttt{DC3} algorithm with default parameters diverged during training on the large datasets, an effect observed by \citet{tordesillas2023rayen}.
To rectify this, we tuned the learning rate of \texttt{DC3}'s correction process for the large dataset, and found that $10^{-8}$ enables the network to learn. 
In \cref{appendix:subsec:additional_benchmark_results}, we investigate if \texttt{DC3}'s performance can be improved by adapting hyperparameters.
For \ouralg we use only $50$ training epochs, while for \texttt{DC3} we use the default $1000$.
For \texttt{JAXopt} we use a tolerance of $10^{-3}$ and $12$ epochs, in the interest of training time.
On both convex and non-convex benchmarks, we use JAXopt as a replacement for our custom projection layer after the backbone NN.
The training times reported are, thus, the ones of the backbone network. We omit comparisons with \texttt{cvxpylayers} \citep{agrawal2019differentiable} since \texttt{JAXopt} is a more recent and stronger baseline: it implements similar functionalities, but it is executable on the GPU; see also \cref{sec:additional-results}.

\paragraph{Metrics.} We compare methods in terms of the following metrics on the test set:
\begin{itemize}[leftmargin=*]
    \item Relative suboptimality (\texttt{RS}): 
    The suboptimality of a candidate solution $\hat{y}$ compared to the optimal objective $J(y^{\star})$,
    computed by the \texttt{Solver}.
    Since methods may violate constraints and obtain a better solution we clip this value,
    $
        \texttt{RS} \coloneqq \max\left(0, (J(\hat{y}) - J(y^\star))/{J(y^\star)}\right).
    $
    \item Constraint violation (\texttt{CV}): 
    We define $\texttt{CV} = \max( \norm{A \hat{y} - \mathrm{x}}_\infty, \norm{\max(C \hat{y} - u, 0)}_\infty )$.
    \item Learning curves: Progress on \texttt{RS} and \texttt{CV} over wall-clock time on the validation set.
    \item Single inference time: The time required to solve one instance at test time.
    \item Batch inference time: The time required to solve a batch of $1024$ instances at test time.
\end{itemize}

Next, we report and discuss the results on the non-convex datasets.
In the interest of space, the results on the convex problems are given in \cref{appendix:subsec:additional_benchmark_results}.
\begin{figure}
    \centering
    \includegraphics[width=.90\linewidth]{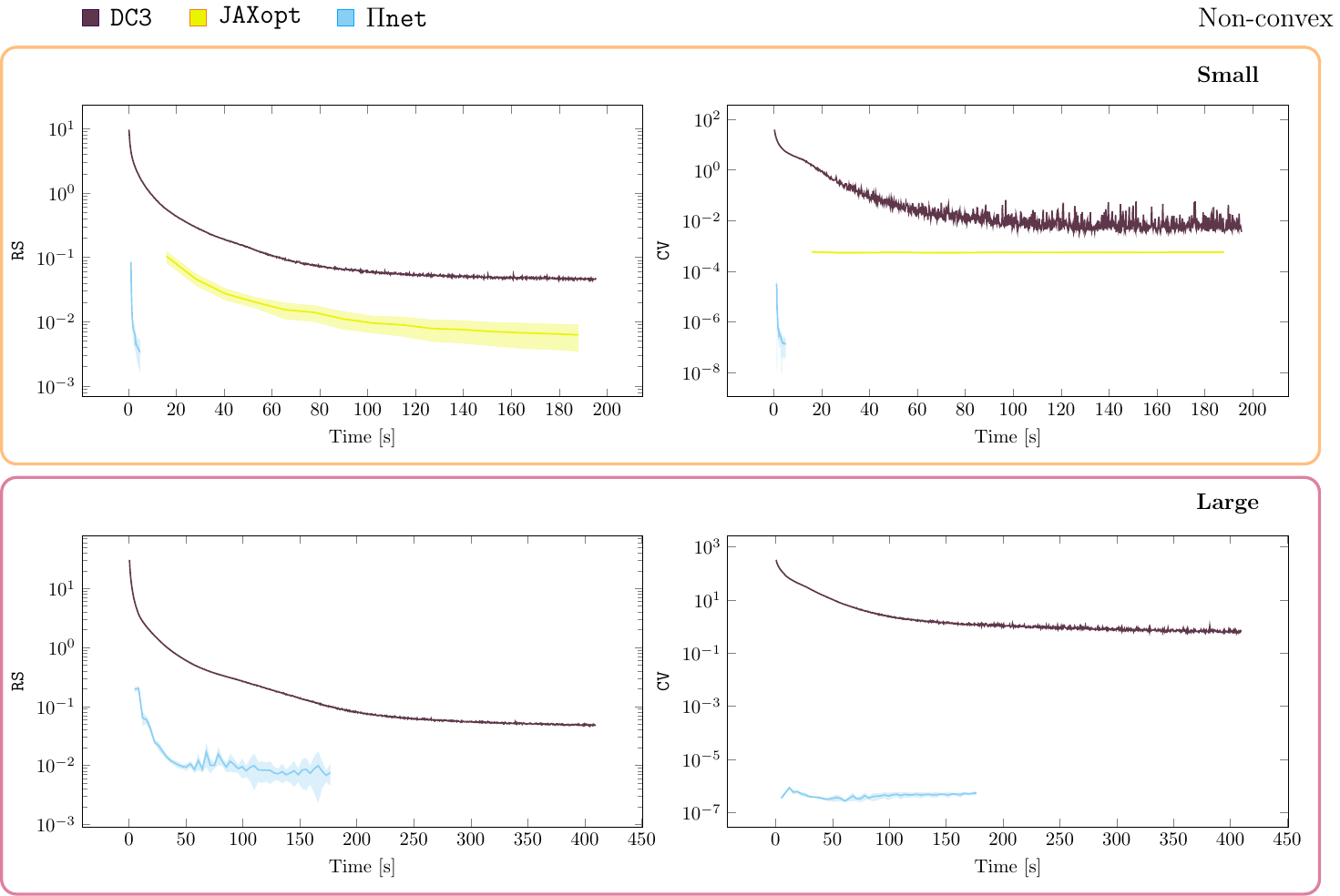}
    \caption{
    Comparison of the learning curves in terms of average \texttt{RS} and \texttt{CV} on the validation set,
    on the small and large non-convex problems. 
    The solid lines denote the mean and the shaded area the standard deviation across 5 seeds. The learning curves for \texttt{JAXopt}
    on the large dataset are reported only in \cref{sec:additional-results} because of the orders of magnitude longer training times.
    }
    \label{fig:benchmark:times:non-convex:small_medium}
\end{figure}
\paragraph{Results.} 
The \texttt{RS} and \texttt{CV} for each problem instance in the test set are reported in \cref{fig:benchmark:non-convex}.
We consider a candidate solution to be optimal if the condition $\texttt{CV} \leq 10^{-3}$ and $\texttt{RS} \leq 5\%$
is satisfied.
These prerequisites for low accuracy solutions are similar to, though somewhat looser from, those employed by numerical solvers \citep{o2016conic,Stellato_2020}. In fact, \ouralg clears these thresholds by a margin.  In practice, any solver achieving a \texttt{CV} below $10^{-5}$ is considered high-accuracy \citep{Stellato_2020} and there is little benefit to go below that. Instead, when methods have sufficiently low \texttt{CV}, having a low \texttt{RS} is better.

Compared to \texttt{DC3}, we correctly solve the vast majority of test problems. 
Importantly, the very low and consistent constraint violation across all problem instances significantly facilitates the tuning of the number of iterations.
By contrast, on the large problems, \texttt{DC3} exhibits \texttt{CV} and \texttt{RS} that are unacceptably large for any meaningful application.
We conjecture that \ouralg significantly outperforms \texttt{DC3} in \texttt{RS} due to the 
absence of soft penalties in our training loss and the orthogonality of the projection.
The \texttt{JAXopt} approach performs better than \texttt{DC3} but significantly worse than \ouralg.

The learning curves are shown in \cref{fig:benchmark:times:non-convex:small_medium}.
\ouralg achieves better performance at a fraction of the training time.
Crucially, our scheme attains satisfactory \texttt{CV} \emph{throughout} training,
implying that \ouralg can reliably compute feasible solutions even with a tiny training budget.
Note that our training curves include the setup time for \ouralg 
(the matrix equilibration, the calculation of the pseudo-inverse for the projection onto the affine subspace, and just-in-time compilation).
We omit the \texttt{JAXopt} results on the large dataset as it requires roughly 14 hours to complete training. The substantial difference in training times between \ouralg and \texttt{JAXopt} underscores the importance of the specialized splitting we employed exploiting the structure of the projection problem,
and our specialized implementation.

We report inference times in \cref{tab:nonconvex_small_medium_times} in \cref{sec:additional-results}.
All approaches significantly outperform the \texttt{Solver}.
Although \texttt{DC3} is slightly faster than \ouralg, it generates solutions of much lower quality (in \texttt{RS} and \texttt{CV}),
hence demonstrating that the advantages of \ouralg come with only a minor runtime trade-off compared to \texttt{DC3}.

To summarize our findings, we showed that existing \gls*{hcnn} methods suffer either from long training/inference times or low quality solutions.
\ouralg address both shortcomings, while retaining modularity and simplicity.

\subsection{\texorpdfstring{\ouralg}{Pinet} applied: Multi-vehicle motion planning}
\label{sec:multi-robot}
\begin{figure}
    \centering
    \includegraphics[width=.95\linewidth]{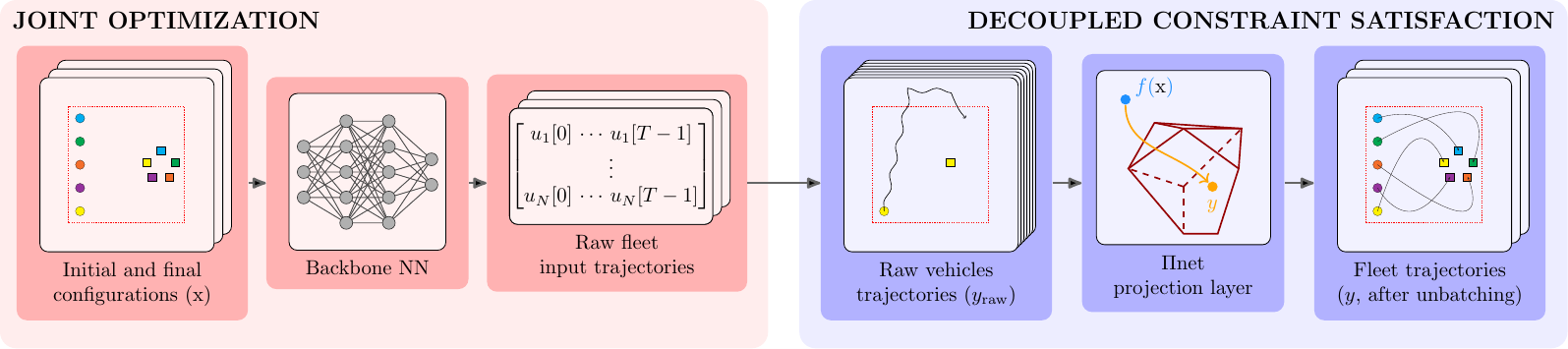}
    \vspace{.1mm}\\
    \hfill
    \begin{minipage}{.3\linewidth}
    \centering\includegraphics[width=.89\linewidth]{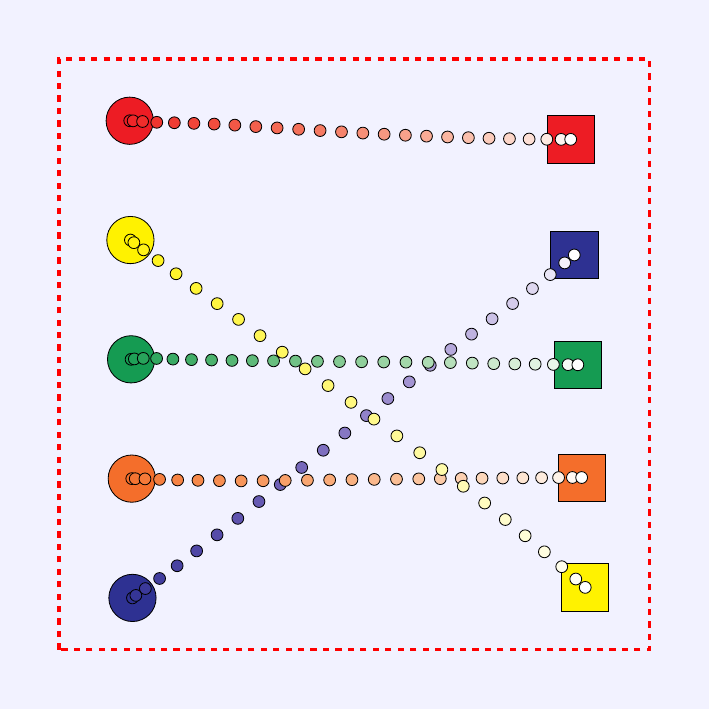}
    \end{minipage}
    \begin{minipage}{.3\linewidth}
\centering\includegraphics[width=.89\linewidth]{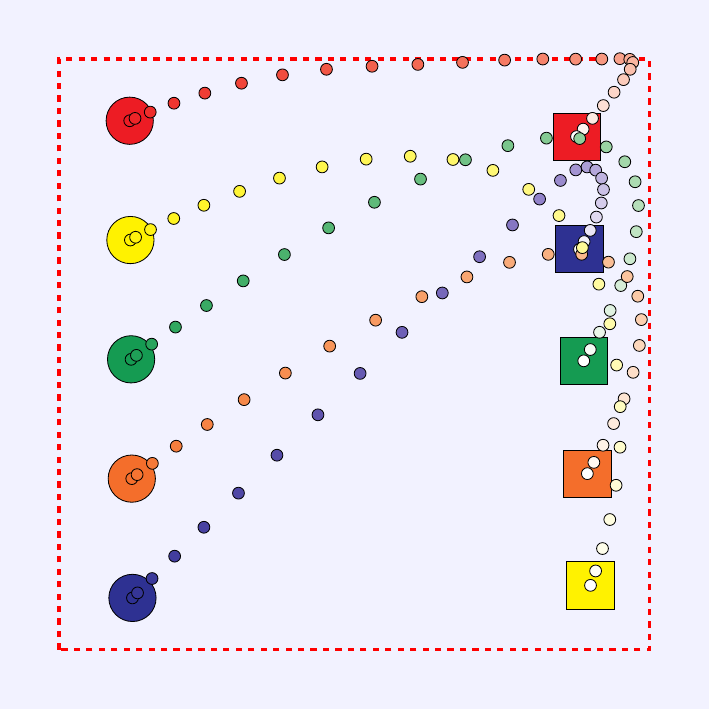}
    \end{minipage}
    \begin{minipage}{.3\linewidth}
    \centering\includegraphics[width=.89\linewidth]{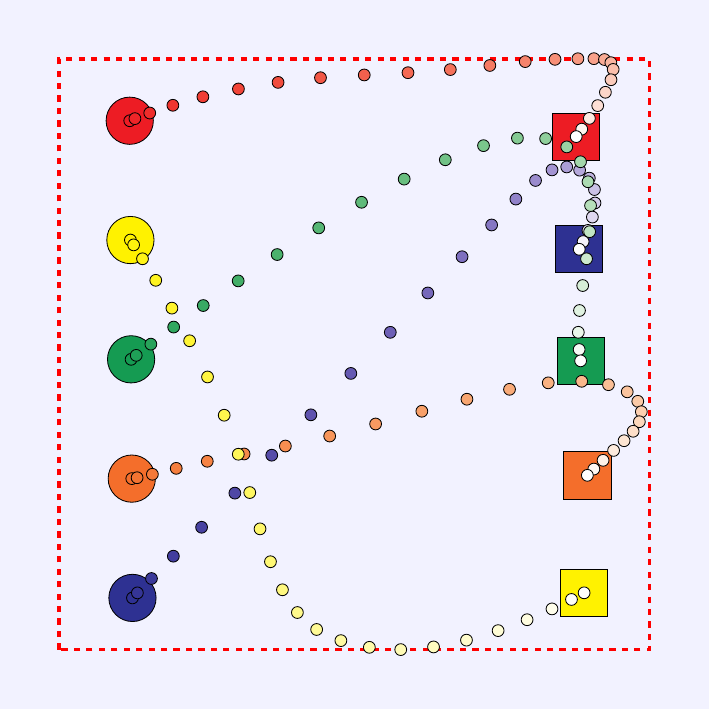}
    \end{minipage}
    \hfill
    \caption{(Top) The \ouralg approach to constrained multi-vehicle motion planning with arbitrary differentiable objective functions $\varphi$. 
    (Bottom) From left to right, we show examples of the synthesized trajectories for 3 different objectives:
    $\varphi_\text{left} = \texttt{effort},\, \varphi_\text{mid} = \varphi_\text{left} + \texttt{preference}, 
    \varphi_\text{right} = \varphi_\text{mid} + \texttt{coverage}$.
    We refer the reader to \cref{appendix:multi-robot} for formal definitions and additional plots.
    }
    \label{fig:multi-robot}
\end{figure}

We present an approach to synthesize transition trajectories between multi-vehicle configurations that optimize some non-linear, 
\emph{fleet-level} objective subject to dynamics, state and input constraints. 
We feed an \gls*{nn} with the initial and terminal fleet configurations (the context $\mathrm{x}$), 
obtain the raw input trajectories and use the vehicle dynamics to infer the full state-input trajectories 
that serves as the raw output $y_{\text{raw}}$, which are then projected for ensured constraint satisfaction; see \cref{fig:multi-robot}.

\paragraph{Experimental setup.} We follow the formulation in the work of \citet{augugliaro2012generation}. 
Specifically, we denote with $p_i[t] \in \reals^m$ the generalized coordinates of vehicle $i$ at the discrete times $t \in \{1, \ldots, T\}$, 
and with $v_i[t]$ and $a_i[t]$ its generalized velocity and acceleration. 
Its simple discretized dynamics read $v_i[t + 1] = v_i[t] + h a_i[t]$, $p_i[t + 1] = p_i[t] + h v_i[t] + \frac{h^2}{2}a_i[t]$.
We formulate the motion planning task for the fleet as a parametric program in the form of \ref{eq:parametric_constrained_optimization},
where $\mathcal{C}(\mathrm{x})$ includes box constraints on positions (workspace constraints), velocities, and accelerations (physical limits), 
affine inequality constraints for jerk limits, and equality constraints for the dynamics and initial/final configuration \emph{for each vehicle}. 
The objective function $\varphi$ encapsulates a \emph{fleet-level} objective;
here we consider a weighted sum of workspace coverage, input effort, and trajectory preference given by a potential function; see \cref{appendix:multi-robot} for a rigorous definition.

\paragraph{Qualitative results.}
We display some of the resulting trajectories for different weights in the objective in \cref{fig:multi-robot} and report more visualizations and analysis for larger fleets and longer horizons in \cref{appendix:multi-robot}. Crucially, both convex and non-convex objectives are handled effectively by \ouralg, resulting in trajectories that adhere to the goals prescribed by the different objective functions.

\paragraph{Practical Relevance.}
Multi-vehicle motion planning has received substantial attention for its practical applications 
\citep{terpin2022distributed,terpin2024dynamic}, 
and we highlight three benefits of our approach:
\begin{itemize}[leftmargin=*]
\item[\ding{51}] \textbf{Constraint satisfaction.} We ensure dynamics, state, and input constraint satisfaction, 
similar to optimization-based trajectory generation methods \citep{augugliaro2012generation}.

\item[\ding{51}] \textbf{Parallelizability.} Our approach is parallelizable in two ways. First, it enables multiple problem instances (different initial and final configurations) to be solved in batches. Second,
since the constraints we consider are decoupled between the vehicles, we can \emph{jointly} predict the raw input trajectories (to enable the network to minimize the joint objective), while solving the projections for each vehicle separately; see \cref{fig:multi-robot}.

\item[\ding{51}] \textbf{Arbitrary objective optimization.} 
Our framework can handle any almost everywhere differentiable objective, encoded in \ouralg's loss $\mathcal{L} = \varphi$.
Importantly, we see this example as a proof of concept towards constrained human-preference optimization (e.g., using the approach of \citet{christiano2017deep}). Deploying a traditional solver for this problem is very challenging because the objective functions considered are not available in closed form and are highly non-linear.
\end{itemize} 

We implement this application in a separate codebase \if@submission%
in the supplementary material,
\else%
\url{https://github.com/antonioterpin/glitch},
\fi
demonstrating also the little overhead required to integrate \ouralg into specific applications. We also explore trajectory planning on a longer horizon (up to $750$ steps, amounting to about $9000$ optimization variables and constraints), as well as high-dimensional contexts (mono-channel, $1024\times1024$ images) in \cref{sec:context-coverage}.
Our current formulation focuses only on convex, decoupled-among-vehicles constraints. 
Future works could address collision avoidance constraints through sequential convexification techniques \citep{augugliaro2012generation,malyuta2022convex}.
\section{Conclusions}
\label{sec:conclusion}
\paragraph{Contributions.}
We introduced an output layer that enforces convex constraints satisfaction on the output of an any backbone \gls*{nn} via an operator splitting scheme. The backpropagation is achieved via the implicit function theorem, enabling efficient training. Our work focused on the gritty details of optimizing \ouralg, introducing also simple yet effective techniques such as hyperparameter tuning and matrix equilibration procedures. 
We showed through extensive benchmarks that \ouralg succeeds where existing learning methods fail.
We provide a GPU-ready implementation in \texttt{JAX}, and showcase how our layer can be embedded in an example application, multi-vehicle motion planning.

\paragraph{Limitations.}
One limitation of our work is the requirement of convex constraint sets. Despite the numerous applications involving only convex constraints \citep{boyd2004convex}, and the numerous applications that can be losslessly convexified \citep{malyuta2022convex}, we acknowledge that future work should investigate how to relax this structural assumption.
One potential approach could involve sequential convexification of non-convex constraints, similar to the algorithm
in \citep{lastrucci2025enforce} that addresses non-linear equality constraints\reviewerOne{, or by employing homeomorphisms \citep{liang2024homeomorphic}}.

\paragraph{Outlook.}
We believe that \ouralg holds the potential to substantially impact a wide range of machine learning domains where constraint satisfaction is crucial. Relevant examples include neural \gls*{pde} solvers \citep{raissi2019physics}, 
scheduling 
\citep{baptiste2001constraint}, and robotics \citep{malyuta2022convex}, among others.
We demonstrated the potential of \ouralg in some of these applications in \cref{sec:experiments}, and we believe that applying our method to new applications represents an exciting avenue for future work. We expect the integration of hard constraints into large-scale models to result in more robust performance and more trustworthy machine learning systems.

\section*{Acknowledgments and Disclosure of Funding}
We would like to thank the reviewers for their constructive comments.
This work was supported as a part of NCCR Automation, a National Centre of Competence in Research, funded by the Swiss National Science Foundation (grant number 51NF40\_225155).

\bibliography{iclr2026_conference}

@article{min2024hard,
  title={Hard-Constrained Neural Networks with Universal Approximation Guarantees},
  author={Min, Youngjae and Sonar, Anoopkumar and Azizan, Navid},
  journal={arXiv preprint arXiv:2410.10807},
  year={2024}
}

@inproceedings{terpin2024learning,
  title={Learning Diffusion at Lightspeed},
  author={Terpin, Antonio and Lanzetti, Nicolas and Gadea, Martín and D\"orfler, Florian},
  booktitle = {Advances in Neural Information Processing Systems},
  year={2024},
}

@inproceedings{wang2023linsatnet,
  title={{LinSATNet}: the positive linear satisfiability neural networks},
  author={Wang, Runzhong and Zhang, Yunhao and Guo, Ziao and Chen, Tianyi and Yang, Xiaokang and Yan, Junchi},
  booktitle={International Conference on Machine Learning},
  year={2023},
}

@article{marquez2017imposing,
  title={Imposing hard constraints on deep networks: Promises and limitations},
  author={M{\'a}rquez-Neila, Pablo and Salzmann, Mathieu and Fua, Pascal},
  journal={arXiv preprint arXiv:1706.02025},
  year={2017}
}

@inproceedings{frerix2020homogeneous,
  title={Homogeneous linear inequality constraints for neural network activations},
  author={Frerix, Thomas and Nie{\ss}ner, Matthias and Cremers, Daniel},
  booktitle={Proceedings of the IEEE/CVF Conference on Computer Vision and Pattern Recognition Workshops},
  year={2020}
}

@article{tordesillas2023rayen,
  title={Rayen: Imposition of hard convex constraints on neural networks},
  author={Tordesillas, Jesus and How, Jonathan P and Hutter, Marco},
  journal={arXiv preprint arXiv:2307.08336},
  year={2023}
}

@inproceedings{donti2021dc3,
  title={{DC3}: A learning method for optimization with hard constraints},
  author={Donti, Priya L and Rolnick, David and Kolter, J Zico},
  booktitle={International Conference on Learning Representations},
  year={2021}
}

@inproceedings{amos2017optnet,
  title={Optnet: Differentiable optimization as a layer in neural networks},
  author={Amos, Brandon and Kolter, J Zico},
  booktitle={International Conference on Machine Learning},
  year={2017},
}

@article{agrawal2019differentiable,
  title={Differentiable convex optimization layers},
  author={Agrawal, Akshay and Amos, Brandon and Barratt, Shane and Boyd, Stephen and Diamond, Steven and Kolter, J Zico},
  journal={Advances in neural information processing systems},
  year={2019}
}

@article{diamond2016cvxpy,
  title={{CVXPY}: A Python-embedded modeling language for convex optimization},
  author={Diamond, Steven and Boyd, Stephen},
  journal={Journal of Machine Learning Research},
  year={2016}
}

@article{Stellato_2020,
   title={{OSQP}: an operator splitting solver for quadratic programs},
   journal={Mathematical Programming Computation},
   publisher={Springer Science and Business Media LLC},
   author={Stellato, Bartolomeo and Banjac, Goran and Goulart, Paul and Bemporad, Alberto and Boyd, Stephen},
   year={2020},
}

@InProceedings{boyle1986method,
author="Boyle, James P.
and Dykstra, Richard L.",
title="A Method for Finding Projections onto the Intersection of Convex Sets in {Hilbert} Spaces",
booktitle="Advances in Order Restricted Statistical Inference",
year="1986",
}

@inproceedings{cristian2023end,
  title={End-to-end learning for optimization via constraint-enforcing approximators},
  author={Cristian, Rares and Harsha, Pavithra and Perakis, Georgia and Quanz, Brian L and Spantidakis, Ioannis},
  booktitle={Proceedings of the AAAI Conference on Artificial Intelligence},
  year={2023}
}

@misc{tuor2021neuromancer,
  title={{NeuroMANCER}: Neural modules with adaptive nonlinear constraints and efficient regularizations},
  author={Tuor, Aaron and Drgona, Jan and Skomski, Elliot and Koch, J and Chen, Z and Dernbach, S and Legaard, CM and Vrabie, D},
  url={https://github.com/pnnl/neuromancer},
  year={2021}
}

@book{boyd2004convex,
  title={Convex optimization},
  author={Boyd, Stephen P and Vandenberghe, Lieven},
  year={2004},
  publisher={Cambridge university press}
}

@article{malyuta2022convex,
  title={Convex optimization for trajectory generation: A tutorial on generating dynamically feasible trajectories reliably and efficiently},
  author={Malyuta, Danylo and Reynolds, Taylor P and Szmuk, Michael and Lew, Thomas and Bonalli, Riccardo and Pavone, Marco and A{\c{c}}{\i}kme{\c{s}}e, Beh{\c{c}}et},
  journal={IEEE Control Systems Magazine},
  year={2022},
  publisher={IEEE}
}

@article{raissi2019physics,
  title={Physics-informed neural networks: A deep learning framework for solving forward and inverse problems involving nonlinear partial differential equations},
  author={Raissi, Maziar and Perdikaris, Paris and Karniadakis, George E},
  journal={Journal of Computational physics},
  year={2019},
  publisher={Elsevier}
}

@book{baptiste2001constraint,
  title={Constraint-based scheduling: applying constraint programming to scheduling problems},
  author={Baptiste, Philippe and Le Pape, Claude and Nuijten, Wim},
  year={2001},
publisher={Springer New York}
}

@inproceedings{chen2018approximating,
  title={Approximating explicit model predictive control using constrained neural networks},
  author={Chen, Steven and Saulnier, Kelsey and Atanasov, Nikolay and Lee, Daniel D and Kumar, Vijay and Pappas, George J and Morari, Manfred},
  booktitle={American Control Conference},
  year={2018},
}

@article{bicgstab,
author = {van der Vorst, H. A.},
title = {{Bi-CGSTAB}: A Fast and Smoothly Converging Variant of {Bi-CG} for the Solution of Nonsymmetric Linear Systems},
journal = {SIAM Journal on Scientific and Statistical Computing},
year = {1992},
}

@article{nellikkath2022physics,
  title={Physics-informed neural networks for {AC} optimal power flow},
  author={Nellikkath, Rahul and Chatzivasileiadis, Spyros},
  journal={Electric Power Systems Research},
  year={2022},
  publisher={Elsevier}
}

@article{bengio2021machine,
  title={Machine learning for combinatorial optimization: a methodological tour d’horizon},
  author={Bengio, Yoshua and Lodi, Andrea and Prouvost, Antoine},
  journal={European Journal of Operational Research},
  year={2021},
}

@article{erichson2019physics,
  title={Physics-informed autoencoders for {Lyapunov}-stable fluid flow prediction},
  author={Erichson, N Benjamin and Muehlebach, Michael and Mahoney, Michael W},
  journal={arXiv preprint arXiv:1905.10866},
  year={2019}
}

@inproceedings{park2023self,
  title={Self-supervised primal-dual learning for constrained optimization},
  author={Park, Seonho and Van Hentenryck, Pascal},
  booktitle={Proceedings of the AAAI Conference on Artificial Intelligence},
  year={2023}
}

@article{lastrucci2025enforce,
  title={{ENFORCE}: Exact Nonlinear Constrained Learning with Adaptive-depth Neural Projection},
  author={Lastrucci, Giacomo and Schweidtmann, Artur M},
  journal={arXiv preprint arXiv:2502.06774},
  year={2025}
}

@InProceedings{fioretto2021lagrangian,
author="Fioretto, Ferdinando
and Van Hentenryck, Pascal
and Mak, Terrence W. K.
and Tran, Cuong
and Baldo, Federico
and Lombardi, Michele",
title="Lagrangian Duality for Constrained Deep Learning",
booktitle="Machine Learning and Knowledge Discovery in Databases. Applied Data Science and Demo Track",
year="2021",
publisher="Springer International Publishing",
}

@article{bai2019deep,
  title={Deep equilibrium models},
  author={Bai, Shaojie and Kolter, J Zico and Koltun, Vladlen},
  journal={Advances in neural information processing systems},
  year={2019}
}

@article{pineda2022theseus,
  title={Theseus: A library for differentiable nonlinear optimization},
  author={Pineda, Luis and Fan, Taosha and Monge, Maurizio and Venkataraman, Shobha and Sodhi, Paloma and Chen, Ricky TQ and Ortiz, Joseph and DeTone, Daniel and Wang, Austin and Anderson, Stuart and others},
  journal={Advances in Neural Information Processing Systems},
  year={2022}
}

@article{winston2020monotone,
  title={Monotone operator equilibrium networks},
  author={Winston, Ezra and Kolter, J Zico},
  journal={Advances in neural information processing systems},
  year={2020}
}

@book{dontchev2009implicit,
  title={Implicit functions and solution mappings},
  author={Dontchev, Asen L and Rockafellar, R Tyrrell},
  year={2009},
  publisher={Springer}
}

@article{butler2023efficient,
  title={Efficient differentiable quadratic programming layers: an {ADMM} approach},
  author={Butler, Andrew and Kwon, Roy H},
  journal={Computational Optimization and Applications},
  year={2023},
}

@article{blondel2022efficient,
  title={Efficient and modular implicit differentiation},
  author={Blondel, Mathieu and Berthet, Quentin and Cuturi, Marco and Frostig, Roy and Hoyer, Stephan and Llinares-L{\'o}pez, Felipe and Pedregosa, Fabian and Vert, Jean-Philippe},
  journal={Advances in neural information processing systems},
  year={2022}
}

@inproceedings{king2024metric,
  title={Metric Learning to Accelerate Convergence of Operator Splitting Methods},
  author={King, Ethan and Kotary, James and Fioretto, Ferdinando and Drgo{\v{n}}a, J{\'a}n},
  booktitle={IEEE 63rd Conference on Decision and Control},
  year={2024}
}

@article{sambharya2024learning,
  title={Learning to warm-start fixed-point optimization algorithms},
  author={Sambharya, Rajiv and Hall, Georgina and Amos, Brandon and Stellato, Bartolomeo},
  journal={Journal of Machine Learning Research},
  year={2024}
}

@article{bertsimas2022online,
  title={Online mixed-integer optimization in milliseconds},
  author={Bertsimas, Dimitris and Stellato, Bartolomeo},
  journal={INFORMS Journal on Computing},
  year={2022},
}

@inproceedings{zamzam2020learning,
  title={Learning optimal solutions for extremely fast {AC} optimal power flow},
  author={Zamzam, Ahmed S and Baker, Kyri},
  booktitle={IEEE International Conference on Communications, Control, and Computing Technologies for Smart Grids},
  year={2020}
}

@article{van2025optimization,
  title={Optimization Learning},
  author={Van Hentenryck, Pascal},
  journal={arXiv preprint arXiv:2501.03443},
  year={2025}
}

@article{amos2023tutorial,
  title={Tutorial on amortized optimization},
  author={Amos, Brandon},
  journal={Foundations and Trends{\textregistered} in Machine Learning},
  year={2023},
}

@article{marcucci2023motion,
  title={Motion planning around obstacles with convex optimization},
  author={Marcucci, Tobia and Petersen, Mark and von Wrangel, David and Tedrake, Russ},
  journal={Science robotics},
  year={2023},
  publisher={American Association for the Advancement of Science}
}

@book{bauschke_convex_2017,
  title={Convex Analysis and Monotone Operator Theory in Hilbert Spaces},
  author={Bauschke, H.H. and Combettes, P.L.},
  year={2017},
  publisher={Springer International Publishing}
}

@article{wathen2015preconditioning,
  title={Preconditioning},
  author={Wathen, Andrew J},
  journal={Acta Numerica},
  year={2015},
  publisher={Cambridge University Press}
}

@article{o2016conic,
  title={Conic optimization via operator splitting and homogeneous self-dual embedding},
  author={O’donoghue, Brendan and Chu, Eric and Parikh, Neal and Boyd, Stephen},
  journal={Journal of Optimization Theory and Applications},
  year={2016},
}

@misc{jax2018github,
  author = {James Bradbury and Roy Frostig and Peter Hawkins and Matthew James Johnson and Chris Leary and Dougal Maclaurin and George Necula and Adam Paszke and Jake Vander{P}las and Skye Wanderman-{M}ilne and Qiao Zhang},
  title = {{JAX}: composable transformations of {P}ython+{N}um{P}y programs},
  year = {2018},
}

@article{wachter2006implementation,
  title={On the implementation of an interior-point filter line-search algorithm for large-scale nonlinear programming},
  author={W{\"a}chter, Andreas and Biegler, Lorenz T},
  journal={Mathematical programming},
  year={2006},
}

@article{terpin2022trust,
  title={Trust region policy optimization with optimal transport discrepancies: Duality and algorithm for continuous actions},
  author={Terpin, Antonio and Lanzetti, Nicolas and Yardim, Batuhan and Dorfler, Florian and Ramponi, Giorgia},
  journal={Advances in Neural Information Processing Systems},
  year={2022}
}

@article{balcerak2025energy,
  title={Energy Matching: Unifying Flow Matching and Energy-Based Models for Generative Modeling},
  author={Balcerak, Michal and Amiranashvili, Tamaz and Terpin, Antonio and Shit, Suprosanna and Bogensperger, Lea and Kaltenbach, Sebastian and Koumoutsakos, Petros and Menze, Bjoern},
    journal={Advances in neural information processing systems},
    year={2022}
}

@inproceedings{augugliaro2012generation,
  title={Generation of collision-free trajectories for a quadrocopter fleet: A sequential convex programming approach},
  author={Augugliaro, Federico and Schoellig, Angela P and D'Andrea, Raffaello},
  booktitle={IEEE/RSJ international conference on Intelligent Robots and Systems},
  year={2012},
}

@inproceedings{terpin2022distributed,
  title={Distributed feedback optimisation for robotic coordination},
  author={Terpin, Antonio and Fricker, Sylvain and Perez, Michel and de Badyn, Mathias Hudoba and D{\"o}rfler, Florian},
  booktitle={2022 American Control Conference},
  year={2022},
}

@article{terpin2024dynamic,
  title={Dynamic programming in probability spaces via optimal transport},
  author={Terpin, Antonio and Lanzetti, Nicolas and D{\"o}rfler, Florian},
  journal={SIAM Journal on Control and Optimization},
  year={2024},
}

@article{christiano2017deep,
  title={Deep reinforcement learning from human preferences},
  author={Christiano, Paul F and Leike, Jan and Brown, Tom and Martic, Miljan and Legg, Shane and Amodei, Dario},
  journal={Advances in neural information processing systems},
  year={2017}
}

@article{grontas2024operatorsplittingconvexconstrained,
  title={Operator Splitting for Convex Constrained {Markov} Decision Processes}, 
  author={Panagiotis D. Grontas and Anastasios Tsiamis and John Lygeros},
  year={2024},
  journal={arXiv preprint arXiv:2412.14002},
}

@article{dantzig2002linear,
  title={Linear programming},
  author={Dantzig, George B},
  journal={Operations research},
  year={2002},
}

@article{zeng2024glinsatgenerallinearsatisfiability,
  title={{GLinSAT}: The General Linear Satisfiability Neural Network Layer By Accelerated Gradient Descent},
  author={Zeng, Hongtai and Yang, Chao and Zhou, Yanzhen and Yang, Cheng and Guo, Qinglai},
  journal={Advances in Neural Information Processing Systems},
  year={2024}
}

@article{condat2016fast,
  title={Fast projection onto the simplex and the l1 ball},
  author={Condat, Laurent},
  journal={Mathematical Programming},
  year={2016},
}

@article{maldonado2014imbalanced,
  title={Imbalanced data classification using second-order cone programming support vector machines},
  author={Maldonado, Sebasti{\'a}n and L{\'o}pez, Julio},
  journal={Pattern Recognition},
  year={2014},
}

@article{brodie2009sparse,
  title={Sparse and stable {Markowitz} portfolios},
  author={Brodie, Joshua and Daubechies, Ingrid and De Mol, Christine and Giannone, Domenico and Loris, Ignace},
  journal={Proceedings of the National Academy of Sciences},
  year={2009},
}

@inproceedings{amos2017input,
  title={Input convex neural networks},
  author={Amos, Brandon and Xu, Lei and Kolter, J Zico},
  booktitle={International Conference on Machine Learning},
  year={2017},
}

@article{davis2017faster,
  title={Faster convergence rates of relaxed {Peaceman-Rachford} and {ADMM} under regularity assumptions},
  author={Davis, Damek and Yin, Wotao},
  journal={Mathematics of Operations Research},
  year={2017},
  publisher={INFORMS}
}

@article{hong2017linear,
  title={On the linear convergence of the alternating direction method of multipliers},
  author={Hong, Mingyi and Luo, Zhi-Quan},
  journal={Mathematical Programming},
  year={2017},
  publisher={Springer}
}

@article{pena2021linear,
  title={Linear convergence of the {Douglas-Rachford} algorithm via a generic error bound condition},
  author={Pe{\~n}a, Javier and Vera, Juan C and Zuluaga, Luis F},
  journal={arXiv preprint arXiv:2111.06071},
  year={2021}
}

@article{busseti2019solution,
  title={Solution refinement at regular points of conic problems},
  author={Busseti, Enzo and Moursi, Walaa M and Boyd, Stephen},
  journal={Computational Optimization and Applications},
  volume={74},
  number={3},
  pages={627--643},
  year={2019},
  publisher={Springer}
}

@article{liang2024homeomorphic,
  title={Homeomorphic projection to ensure neural-network solution feasibility for constrained optimization},
  author={Liang, Enming and Chen, Minghua and Low, Steven H},
  journal={Journal of Machine Learning Research},
  volume={25},
  number={329},
  pages={1--55},
  year={2024}
}

@article{kratsios2021universal,
  title={Universal approximation under constraints is possible with transformers},
  author={Kratsios, Anastasis and Zamanlooy, Behnoosh and Liu, Tianlin and Dokmani{\'c}, Ivan},
  journal={arXiv preprint arXiv:2110.03303},
  year={2021}
}

@inproceedings{tabas2022safe,
  title={Safe and efficient model predictive control using neural networks: An interior point approach},
  author={Tabas, Daniel and Zhang, Baosen},
  booktitle={2022 IEEE 61st Conference on Decision and Control (CDC)},
  pages={1142--1147},
  year={2022},
  organization={IEEE}
}

@inproceedings{simon1984lectures,
  title={Lectures on geometric measure theory},
  author={Simon, Leon and others},
  year={1984},
  organization={Centre for Mathematical Analysis, Australian National University Canberra}
}

@article{bolte2021conservative,
  title={Conservative set valued fields, automatic differentiation, stochastic gradient methods and deep learning},
  author={Bolte, J{\'e}r{\^o}me and Pauwels, Edouard},
  journal={Mathematical Programming},
  volume={188},
  number={1},
  pages={19--51},
  year={2021},
  publisher={Springer}
}

@article{bolte2021nonsmooth,
  title={Nonsmooth implicit differentiation for machine-learning and optimization},
  author={Bolte, J{\'e}r{\^o}me and Le, Tam and Pauwels, Edouard and Silveti-Falls, Tony},
  journal={Advances in neural information processing systems},
  volume={34},
  pages={13537--13549},
  year={2021}
}

@article{bolte2024differentiating,
  title={Differentiating nonsmooth solutions to parametric monotone inclusion problems},
  author={Bolte, J{\'e}r{\^o}me and Pauwels, Edouard and Silveti-Falls, Antonio},
  journal={SIAM Journal on Optimization},
  volume={34},
  number={1},
  pages={71--97},
  year={2024},
  publisher={SIAM}
}
\bibliographystyle{iclr2026_conference}

\clearpage
\appendix
\crefalias{section}{appendix}
\crefalias{subsection}{appendix}
\crefname{appendix}{Appendix}{Appendices}
\tableofcontents
\clearpage
\listoffigures
\clearpage
\listoftables
\clearpage
\section{Toy visual example}
\label{sec:experiments:toy-example}
In this section, we instantiate our method on a very simple and interpretable toy problem: learning to predict the outputs of a \gls*{mpc} policy \citep{chen2018approximating}. In high-dimensional scenarios and for long horizons, running \gls*{mpc} in real-time often becomes a computational bottleneck, 
so learning to predict its outputs--even with minor suboptimality, as long as constraints are still satisfied--can bring substantial benefits \citep{chen2018approximating}.

\paragraph{Experimental setting.} We consider a two-dimensional single integrator system,
and define an \gls{mpc} control law given by the solution of
\begin{equation}\label{eq:mpc:toy}
    \underset{u_k \in [-1, 1]^2}{\mathclap{\mathrm{minimize}}} 
    \sum_{k=0}^{N-1} \norm{x_k - \hat{x}}^2 + \norm{u_k}^2
    \qquad
\overset{\hphantom{ y \in \reals^{d}}}{\mathclap{\mathrm{subject~to}}} 
\qquad 
x_0 = \mathrm{x},\,
x_{k+1} = x_k + u_k,\, 
x_k \in [-10, 10]^2.
\end{equation}%
Here, $x_k$ denotes the system state, $u_k$ the control input, and $\hat{x} = [3,\, -12]$ is a target state. 
We sample the context (in this case, the initial condition for the \gls*{mpc}), $\mathrm{x} = x_0$, uniformly in $[-10,10]^2$.
We deploy \ouralg to learn the solution of \eqref{eq:mpc:toy} as a function of $\mathrm{x}$.
We use a self-supervised loss and the backbone \gls{nn} is an \gls*{mlp} with two hidden layers of 200 neurons and \texttt{ReLU} activations. 

\paragraph{Results.}
We superimpose \ouralg's prediction
and the solution of \eqref{eq:mpc:toy} given by a \texttt{Solver}.
Further, to illustrate the importance of \emph{hard} constraints, 
we also plot the prediction of the same \gls*{mlp} trained with soft constraints, i.e., 
adding to the loss the term $\lambda (\texttt{EQCV} + \texttt{INEQCV})$, 
where \texttt{EQCV} and \texttt{INEQCV} is the maximum violation of the equality and 
inequality constraints, respectively, for two values of $\lambda$.

\begin{minipage}{.54\linewidth} 
We represent $\hat{x}$ with a star, $\mathrm{x}$ with a square, and
state constraints with a dashed rectangle.
We tested the trained network in $1000$ instances generated by sampling $\mathrm{x}$ uniformly in $[-10,10]^2$. 
\ouralg achieves an average relative suboptimality of approximately $0.02\%$, and constraints satisfied
in $100\%$ of problem instances to within a tolerance of $10^{-3}$. It is also apparent that different values of $\lambda$ induce different behaviors: Larger values enforce the constraints at the expense of optimality, smaller values do not enforce constraints.
\end{minipage}
\hfill
\begin{minipage}{.45\linewidth}
    \centering
        \begin{tikzpicture}
\begin{axis}[
    width=1.1\linewidth,
    axis lines=none,
    clip=false,
    xtick=\empty, ytick=\empty,
    legend cell align={left},
]
\addplot [red, dashed, forget plot] coordinates {(-0.0,-0.0) (1.0,-0.0) (1.0,1.0) (-0.0,1.0) (-0.0,-0.0)};

\foreach \k in {1} {
    \pgfplotstableread[col sep=comma]{images/toy/trajectory_\k.csv}\data;
    \pgfplotstableread[col sep=comma]{images/toy/trajectory_\k_softtwo.csv}\datasoftone;
    \pgfplotstableread[col sep=comma]{images/toy/trajectory_\k_softone.csv}\datasofttwo;
    \addplot [
        thick,
        black,
        solid,
        mark=x
    ] table [
        x=x,
        y=y,
        col sep=comma
    ] {\data};

    \addplot [
        thick,
        blue,
        mark=o,
        mark size=3pt
    ] table [
        x=xgt,
        y=ygt,
        col sep=comma
    ] {\data};

    \addplot [
        ultra thick,
        black,
        solid,
        mark=o,
        mark size=1pt
    ] table [
        x=x,
        y=y,
        col sep=comma
    ] {\datasoftone};

    \addplot [
        ultra thick,
        brown,
        dotted,
        mark=o,
        mark size=1pt
    ] table [
        x=x,
        y=y,
        col sep=comma
    ] {\datasofttwo};
}
\node at (axis cs:0.65,-0.1) {\Large$\star$};
\node[right] at (axis cs:0.65,-0.1) {$\hat{x}$};

\node[draw,right,minimum width=1pt] at (axis cs:0.125,0.25) {};
\node[left] at (axis cs:0.125,0.25) {$\mathrm{x}$};

\legend{\ouralg, \texttt{Solver}, \texttt{SoftMLP} $\lambda = 1$, \texttt{SoftMLP} $\lambda = 0.001$}

\end{axis}
\end{tikzpicture}
    \end{minipage} 
\clearpage
\section{Additional results}
\label{sec:additional-results}
In this section, we collect the additional results for \cref{sec:experiments} and the ablation studies on some \texttt{DC3} hyperparameters.

\subsection{A comparison with \texorpdfstring{\texttt{cvxpylayers}}{cvxpylayers}.}
We compare \ouralg with \texttt{cvxpylayers} on the small non-convex benchmark, and report the results in \cref{tab:pinetvscvxpy}. 
We use \texttt{cvxpylayers} as an alternative to our custom projection layer, in an analogous manner to the comparison with \texttt{JAXopt} in \cref{sec:experiments:benchmarks}.
Since \texttt{cvxpylayers} runs exclusively on the CPU, we report the results of \ouralg on both CPU and GPU. 
In all cases, we provide a training budget of 25 epochs. For \texttt{cvxpylayers} this corresponds to roughly 10 minutes of wall-clock time, whereas for \ouralg this corresponds to 4 and 28 seconds on the GPU and CPU, respectively. We observe that 
\ouralg attains similar performance in terms of \texttt{RS} and \texttt{CV}, but significantly outperforms \texttt{cvxpylayers} in terms of inference and training time. We omit further results with \texttt{cvxpylayers}, since \texttt{JAXopt} is a more recent and stronger baseline: it implements similar functionalities, but it is executable on the GPU (see \cref{fig:benchmark:non-convex,fig:benchmark:times:non-convex:small_medium}, and \cref{tab:nonconvex_small_medium_times}).
\begin{table}
    \centering
    \begin{tabular}{l|llll}
        \bf Method & \bf \texttt{RS}& \bf \texttt{CV} & \bf Single inference time [s]& \bf Batch inference time [s]
        \\\hline
        \texttt{cvxpylayers} & 0.0036 & 0.00005 & 0.0120 & 2.5917\\
        \rowcolor{blue!10}\ouralg (CPU) & 0.0035 & 0.00000 & 0.0052 & 0.0957\\
        \rowcolor{blue!10}\ouralg (GPU) & 0.0035 & 0.00000 & 0.0065 & 0.0135\\
    \end{tabular}
    \caption{Comparison with \texttt{cvxpylayers} on the small, non-convex benchmark. The \texttt{RS} and \texttt{CV} values reported are the averages over the test set.}
    \label{tab:pinetvscvxpy}
\end{table}

\subsection{Additional analyses for \texorpdfstring{\cref{sec:experiments:benchmarks}}{Section 3.2}.} \label{appendix:subsec:additional_benchmark_results}
\begin{figure}
    \centering
    \begin{minipage}{\linewidth}
    \centering
        \includegraphics[width=\linewidth]{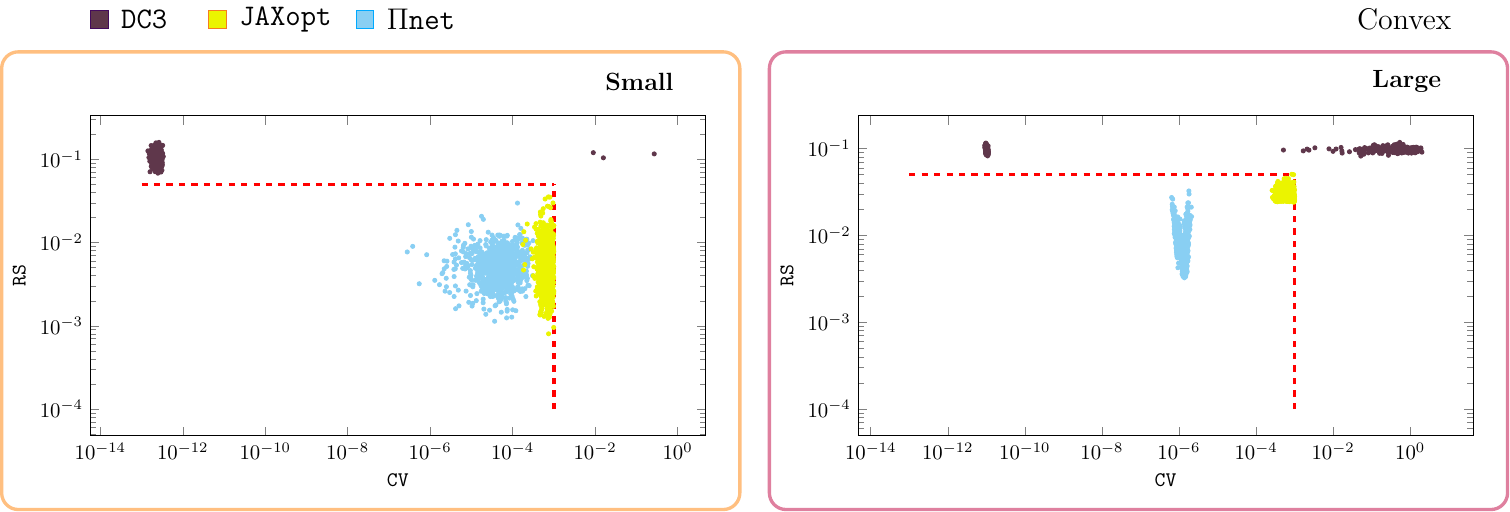}
    \end{minipage}
\caption{
    Scatter plots of \texttt{RS} and \texttt{CV} on the small and large convex problems
    on the test set. 
    The red dashed lines show the thresholds to consider a candidate solution optimal. 
    }
    \label{fig:benchmark_appendix:convex}
\end{figure}

\begin{figure}
    \centering
    \includegraphics[width=\linewidth]{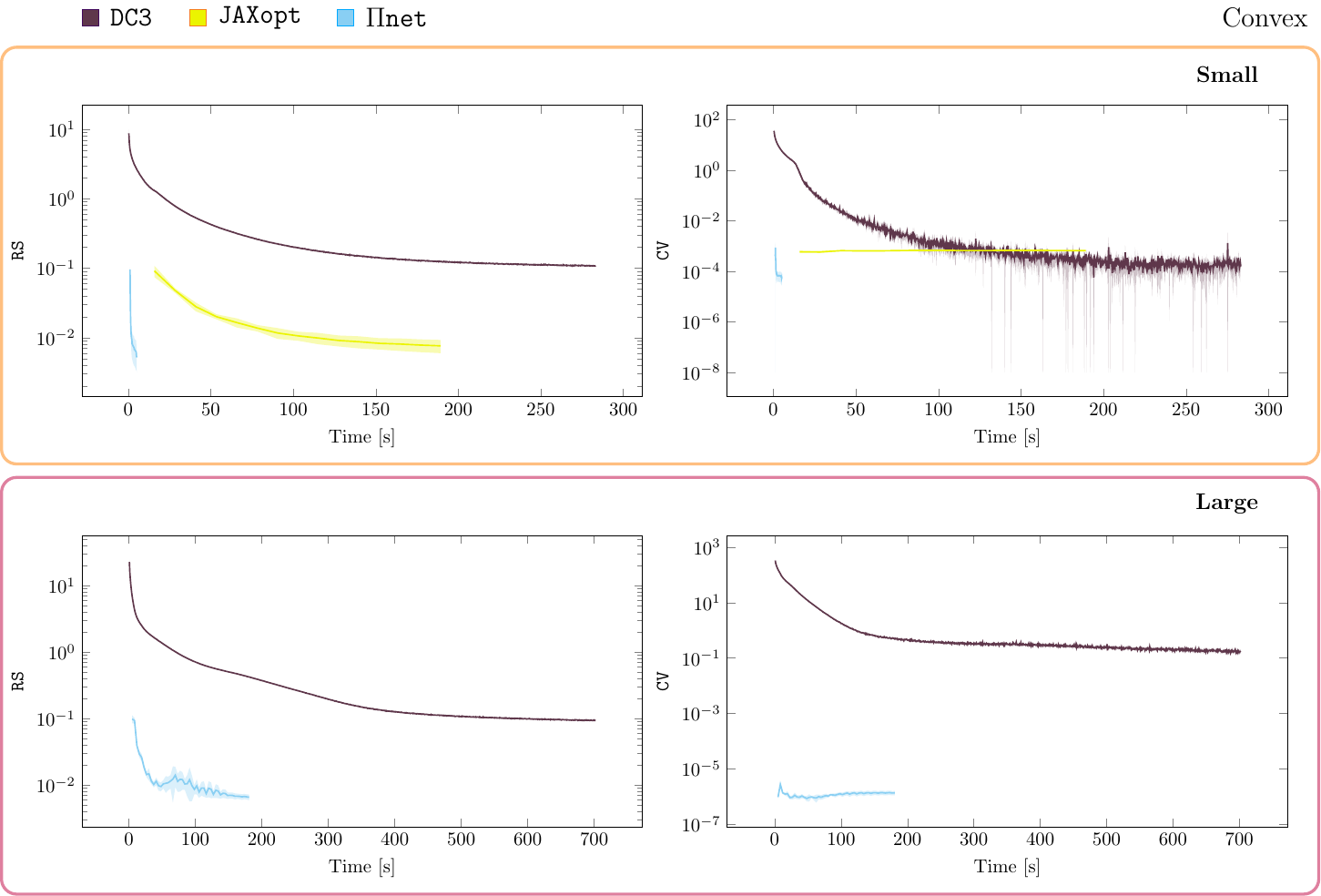}
    \caption{
    Comparison of the learning curves in terms of average \texttt{RS} and \texttt{CV} on the validation set,
    on the small and large convex problems. 
    The solid lines denote the mean and the shaded area the standard deviation across 5 seeds. 
    The learning curves for \texttt{JAXopt} on the large dataset are reported in \cref{fig:jaxopt_convex:learning_curves} because of the orders of magnitude longer training times.
    }
    \label{fig:benchmark:times:convex}
\end{figure}

\begin{figure}
    \centering
    \includegraphics[width=\linewidth]{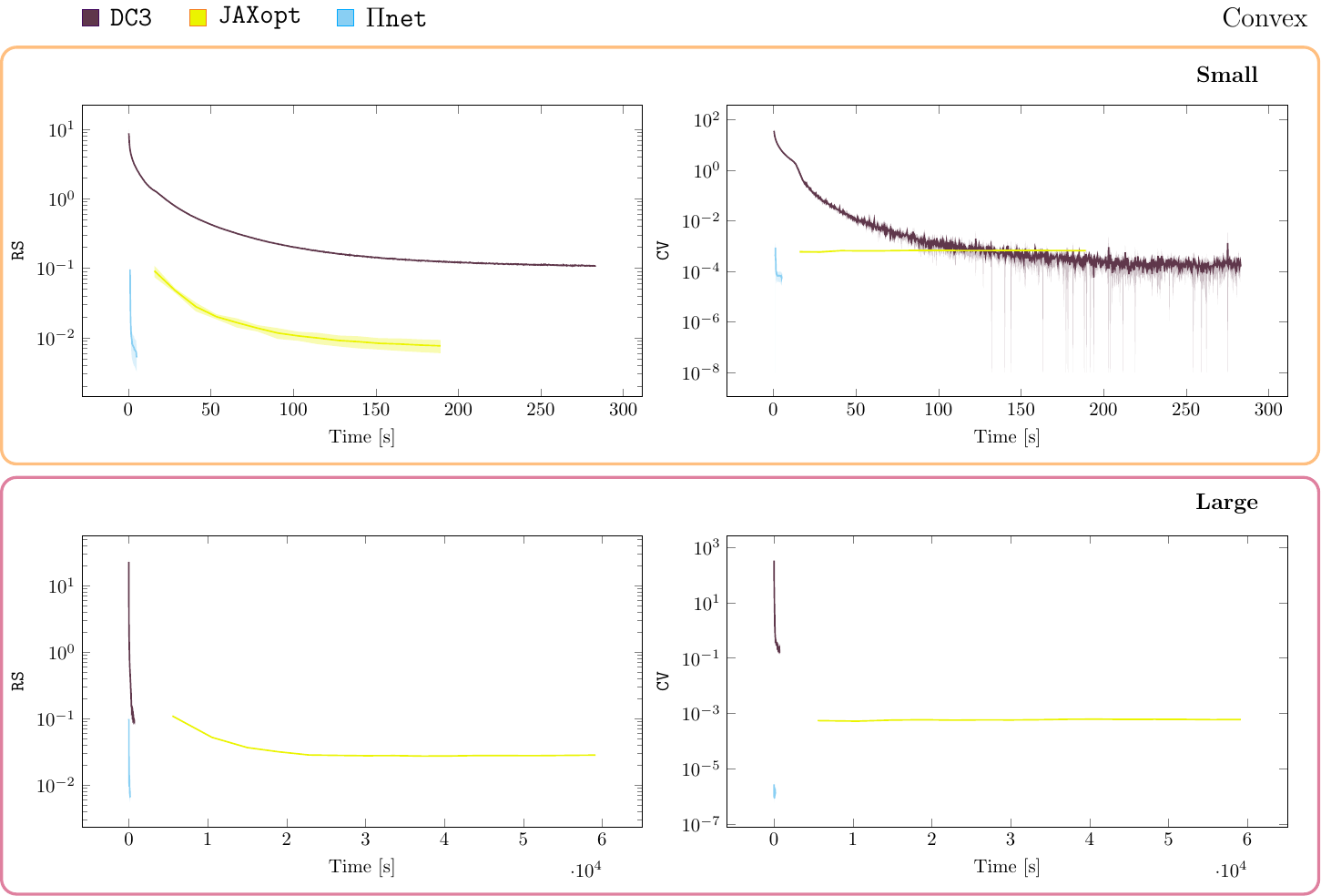}
    \caption{Comparison of the training times in terms of \texttt{RS} and \texttt{CV} for the different methods in the small and large convex problem setting. 
    The solid lines denote the mean and the shaded area the standard deviation across 10 seeds.
    Note the different timescale on the large dataset results, due to the large training time of \texttt{JAXopt}.
    }
    \label{fig:jaxopt_convex:learning_curves}
\end{figure}

\begin{figure}
    \centering
    \includegraphics[width=\linewidth]{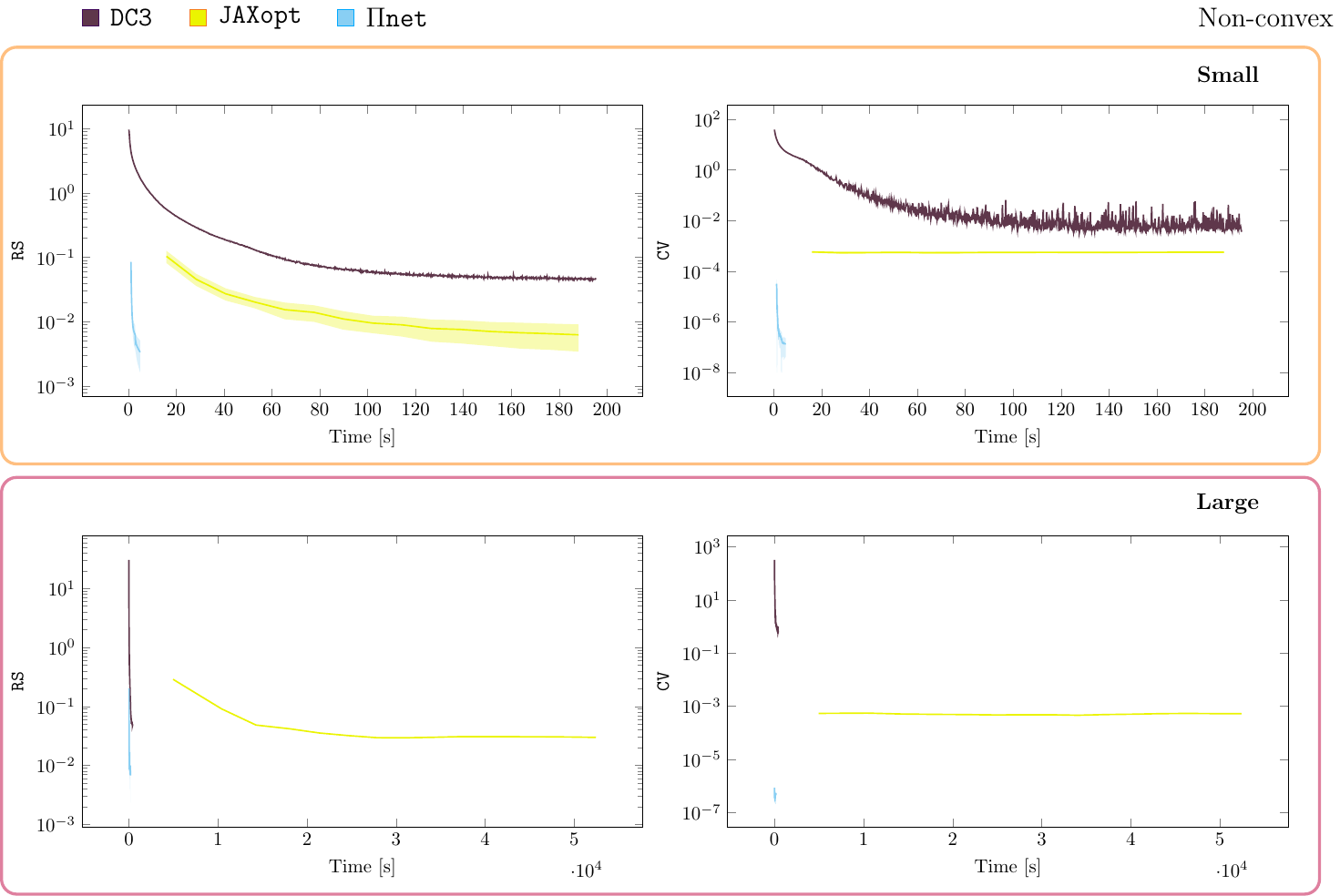}
    \caption{
    Comparison of the learning curves in terms of average \texttt{RS} and \texttt{CV} on the validation set,
    on the small and large non-convex problems. 
    The solid lines denote the mean and the shaded area the standard deviation across 5 seeds. 
    }
    \label{fig:jaxopt_nonconvex:learning_curves}
\end{figure}

\paragraph{Additional results.}
We report the inference times on the non-convex datasets in \cref{tab:nonconvex_small_medium_times}.
We report the omitted benchmark results on the small and large convex datasets of \cref{sec:experiments:benchmarks} in \cref{fig:benchmark_appendix:convex,tab:convex_small_medium_times,fig:benchmark:times:convex}, which further substantiate our claims.
We report the omitted learning curves for \texttt{JAXopt} on the large convex and non-convex datasets in
\cref{fig:jaxopt_convex:learning_curves} and \cref{fig:jaxopt_nonconvex:learning_curves}, respectively.

\begin{table}
    \centering
    \vspace{.5cm}
    
\begin{adjustbox}{width=\textwidth,center}
    \begin{tabular}{l|ccccc|ccccc}
\multirow{2}{*}{\textbf{Method}} & 
\multicolumn{5}{c|}{\textbf{Single inference [s]}} & 
\multicolumn{5}{c}{\textbf{Batch inference [s]}} \\
\cline{2-11}
 & median & LQ & UQ & $\min$ & $\max$ & median & LQ & UQ & $\min$ & $\max$ \\
 \hline
\multicolumn{11}{l}{\textbf{Non-convex small}} \\\hline
\texttt{DC3}    &  
0.0019&0.0018&0.0019&0.0018&0.0026
&
0.0020&0.0019&0.0020&0.0019&0.0023
\\
\texttt{Solver} & 
0.0334&0.0298&0.0497&0.0207&0.1213
&
41.748&40.200&43.198&36.764&47.149
\\
\texttt{JAXopt} & 
0.0134&0.0131&0.0136&0.0122&0.0145
&
0.1371&0.1364&0.1532&0.1357&0.1540
\\
\rowcolor{blue!10}\ouralg   &
0.0056&0.0055&0.0056&0.0054&0.0072
&
0.0130&0.0128&0.0131&0.0123&0.0147
\\
\hline
\multicolumn{11}{l}{\textbf{Non-convex large}} \\\hline
\texttt{DC3}    & 
0.0016&0.0016&0.0017&0.0015&0.0046
&
0.0248&0.0248&0.0248&0.0247&0.0255
\\
\texttt{Solver} &
10.159&9.4739&10.807&7.4350&12.612
&
10720&9538.4&10800&9417.8&16502
\\
\texttt{JAXopt} & 
0.0578&0.0575&0.0590&0.0559&0.0596
&
19.430&19.430&19.430&19.429&19.430
\\
\rowcolor{blue!10}\ouralg   &  
0.0063 & 0.0063 & 0.0064 & 0.0061 & 0.0092
&
0.2804 & 0.2799 & 0.2807 & 0.2794 & 0.2912
\\
\end{tabular}
\end{adjustbox}
\vspace{.125cm}
    \caption{Inference time comparison for single-instance and batch (1024 contexts) settings,
    evaluated on the small and large non-convex problems. 
    The table reports the median, lower quartile (LQ, 25th percentile), upper quartile (UQ, 75th percentile), 
    $\min$ and $\max$
    of the runtime.
    }
    \label{tab:nonconvex_small_medium_times}
\end{table}

\begin{table}
    \centering
    \vspace{.5cm}
    
\begin{adjustbox}{width=\textwidth,center}
    \begin{tabular}{l|ccccc|ccccc}
\multirow{2}{*}{\textbf{Method}} & 
\multicolumn{5}{c|}{\textbf{Single inference [s]}} & 
\multicolumn{5}{c}{\textbf{Batch inference [s]}} \\
\cline{2-11}
 & median & LQ & UQ & $\min$ & $\max$ & median & LQ & UQ & $\min$ & $\max$ \\
 \hline
\multicolumn{11}{l}{\textbf{Convex small}} \\\hline
\texttt{DC3}    &  
0.0033&0.0032&0.0033&0.0031&0.0050
&
0.0033 & 0.0033 & 0.0034&0.0032&0.0044\\
\texttt{Solver} & 
0.0019&0.0018&0.0019&0.0011&0.0083
&
1.9350&1.9264&1.9441&1.8931&2.0282\\
\texttt{Solver${}^{\dagger}$} & 
0.0006&0.0006&0.0006&0.0006&0.0024
&
0.6190&0.6168&0.6217&0.6100&0.6746\\
\texttt{JAXopt} & 
0.0142&0.0138&0.0150&0.0124&0.0165
&
0.1603&0.1590&0.1648&0.1581&0.1794\\
\rowcolor{blue!10}\ouralg   &  
0.0055 & 0.0055 & 0.0056 & 0.0053 & 0.0060
&
0.0130 & 0.0128 & 0.0130 & 0.0125 & 0.0136 \\
\hline
\multicolumn{11}{l}{\textbf{Convex large}} \\\hline
\texttt{DC3}    & 
0.0072&0.0072&0.0073&0.0071&0.0115
&
0.0349&0.0349&0.0351&0.0348&0.0369
\\
\texttt{Solver} & 
0.6660&0.6613&0.6716&0.3159&0.7079
&
680.59&676.24&682.52&662.29&687.84
\\
\texttt{Solver}${}^{\dagger}$ & 
0.0603&0.0589&0.0620&0.0466&0.9501
&
62.022&61.591&62.558&60.064&70.987
\\
\texttt{JAXopt} & 
0.0630&0.0623&0.0641&0.0601&0.0668
&
21.504&21.504&21.505&21.504&21.505\\
\rowcolor{blue!10}
\ouralg   & 
0.0063&0.0063&0.0064&0.0061&0.0145
&
0.2800&0.2797&0.2804&0.2794&0.2851
\\
\hline
\end{tabular}
\end{adjustbox}
\vspace{.125cm}
    \caption{Inference time comparison for single-instance and batch (1024 contexts) settings,
    evaluated on the small and large convex problems. 
    The table reports the median, lower quartile (LQ, 25th percentile), upper quartile (UQ, 75th percentile), 
    $\min$ and $\max$
    of the runtime.
    We report results of the \texttt{Solver}  (i.e., \texttt{OSQP}) in two modes, normal and parametric labeled
    with \texttt{Solver} and \texttt{Solver}${}^{\dagger}$, respectively.
    Parametric mode means that we inform \texttt{OSQP} that we are repeatedly solving problems with the same structure,
    which speeds up solution time by reusing calculations across consecutive calls. 
    We note that this is a feature of \texttt{OSQP} that may or may not be available in other solvers.
    }
    \label{tab:convex_small_medium_times}
\end{table}

\begin{figure}
    \centering
    \begin{minipage}{\linewidth}
    \centering
        \includegraphics[width=\linewidth]{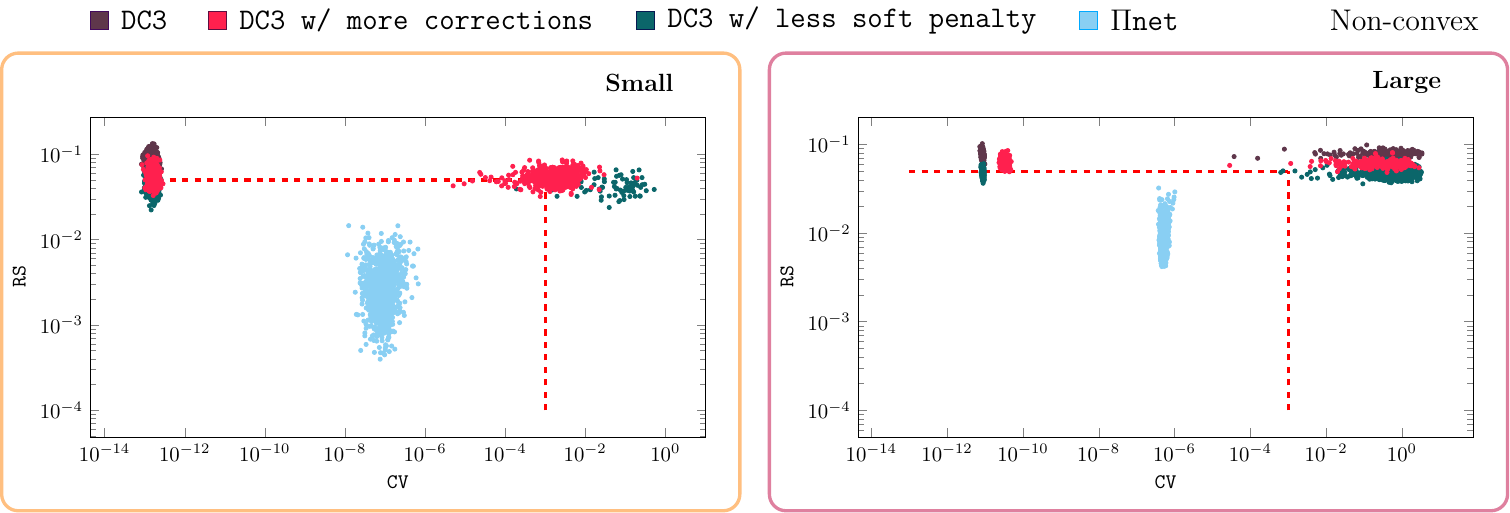}%
    \end{minipage}
    \caption{
    Scatter plots of \texttt{RS} and \texttt{CV} on the small and large non-convex problems on the test set. 
    The red dashed lines show the thresholds to consider a candidate solution optimal. 
    We superimpose the results of the main text with the ones obtained using more correction steps and a smaller soft penalty in the \texttt{DC3} algorithm.
    }
    \label{fig:benchmark:dc3ablations}
\end{figure}

\paragraph{\texttt{DC3}: Hyperparameter ablations.} 
In our benchmarks of \cref{sec:experiments:benchmarks}, we highlighted two weaknesses in the performance of \texttt{DC3}:
(i) in both small and large non-convex problems the \texttt{RS} is unsatisfactory;
(ii) in the large non-convex problem the \texttt{CV} is unsatisfactory.
We investigated whether these shortcomings can be mitigated by choosing hyperparameters 
which are different from the defaults.
In particular, we tried to address 
(i) by decreasing the soft penalty parameter for \texttt{CV} in \texttt{DC3}'s loss function from the default $10.0$ to $2.0$;
(ii) by increasing the number of correction steps from the default $10$ to $50$.
We superimpose the obtained \texttt{RS} and \texttt{CV} in \cref{fig:benchmark:dc3ablations}, 
and report the inference times in \cref{tab:nonconvex_small_medium_times_dc3_more_correction}
(only for the case of more correction steps, since changing the soft penalty does not affect inference times).
Importantly, these results show both the lack of a clear decoupling between the various parameters of the DC3 algorithm and the advantages of our projection scheme; 
see also \cref{sec:sharp-bits}.
\begin{table}
    \centering
    \vspace{.5cm}
    
\begin{adjustbox}{width=\textwidth,center}
    \begin{tabular}{l|ccccc|ccccc}
\multirow{2}{*}{\textbf{Method}} & 
\multicolumn{5}{c|}{\textbf{Single inference [s]}} & 
\multicolumn{5}{c}{\textbf{Batch inference [s]}} \\
\cline{2-11}
 & median & LQ & UQ & $\min$ & $\max$ & median & LQ & UQ & $\min$ & $\max$ \\
\hline
\multicolumn{11}{l}{\textbf{Non-convex small}} \\\hline
\texttt{DC3}    &  
0.0080&0.0080&0.0080&0.0079&0.0087
&
0.0081&0.0081&0.0082&0.0081&0.0083
\\
\texttt{Solver} & 
0.9724&0.6925&1.4587&0.3562&3.6359
&
1258.8&1243.8&1264.3&1185.2&1299.0
\\
\rowcolor{blue!10}\ouralg   &
0.0056&0.0055&0.0056&0.0054&0.0072
&
0.0130&0.0128&0.0131&0.0123&0.0147
\\
\hline
\multicolumn{11}{l}{\textbf{Non-convex large}} \\\hline
\texttt{DC3}    & 
0.0070&0.0069&0.0073&0.0068&0.0105
&
0.1177&0.1177&0.1177&0.1176&0.1180
\\
\texttt{Solver} &
6.8153&6.7161&6.9636&6.2371&12.046
&
6991.4&6986.1&7498.7&6964.0&7523.3
\\
\rowcolor{blue!10}\ouralg   &  
0.0063 & 0.0063 & 0.0064 & 0.0061 & 0.0092
&
0.2804 & 0.2799 & 0.2807 & 0.2794 & 0.2912
\\
\end{tabular}
\end{adjustbox}
\vspace{.125cm}
    \caption{Inference time comparison for single-instance and batch-instance (1024 problems) settings across different methods,
    evaluated on small and large non-convex problems. 
    The table reports median runtime along with statistical descriptors: 
    lower quartile (LQ, 25th percentile), upper quartile (UQ, 75th percentile), 
    $\min$ and $\max$ of the runtime.
    \texttt{DC3} uses more than the default correction steps.
    }
    \label{tab:nonconvex_small_medium_times_dc3_more_correction}
\end{table}

\subsection{Additional details and results for \texorpdfstring{\cref{sec:multi-robot}}{the multi-robot planning application}}
\label{appendix:multi-robot}
We detail the problem setup for the multi-vehicle motion planning application in \cref{sec:multi-robot} and present additional results. Recall that we denote with $p_i[t] \in \reals^m$ the generalized coordinates of vehicle $i$ at the discrete times $t \in \{1, \ldots, T + 1\}$, and with $v_i[t]$ and $a_i[t]$ its generalized velocity and acceleration. Its simple discretized dynamics read $v_i[t + 1] = v_i[t] + h a_i[t]$, $p_i[t + 1] = p_i[t] + h v_i[t] + \frac{h^2}{2}a_i[t]$. 
We have constraints on each of these variables and, thus, we consider as optimization variable $y = \begin{bmatrix}
    p^\top & v^\top & a^\top
\end{bmatrix}^\top$, with
\begin{align*} 
p &= \begin{bmatrix}
p_1[0]^\top & \cdots & p_N[0]^\top & \cdots & p_1[T]^\top & \cdots & p_N[T]^\top
\end{bmatrix}^\top,
\\
v &= \begin{bmatrix}
    v_1[0]^\top & \cdots & v_N[0]^\top & \cdots & v_1[T]^\top & \cdots & v_N[T]^\top
\end{bmatrix}^\top,
\\
a &= \begin{bmatrix}
    a_1[0]^\top & \cdots & a_N[0]^\top & \cdots & a_1[T - 1]^\top & \cdots & a_N[T - 1]^\top
\end{bmatrix},
\end{align*}
where $N$ is the number of vehicles. Below, we use $m = 2$. We want the network to generate trajectories that go from a given set of initial positions $\bar{p}_1[0], \ldots, \bar{p}_N[0]$ to a set of final positions $\bar{p}_1[T + 1], \ldots, \bar{p}_N[T + 1]$. This is the \emph{context} of the optimization problem, i.e.,
\[
\mathrm{x} = \begin{bmatrix}
    \bar{p}_1[0]^\top & \cdots & \bar{p}_N[0]^\top & \cdots & \bar{p}_1[T + 1]^\top & \cdots & \bar{p}_N[T + 1]^\top
\end{bmatrix}^\top.
\]

The constraints on the system are:
\begin{itemize}[leftmargin=*]
    \item \emph{Dynamic constraints}. The optimization variables $p, v, a$ are related by the system dynamics, via an equality constraint of the type $A_{\text{dyn}} y = 0$.
    \item \emph{Initial and final positions constraints}. We ensure that the optimal $y$ satisfies the given initial and terminal position constraints via the equality constraint $A_{\text{if}} y = b(\mathrm{x})$. 
    Importantly, we observe that the context affects the constraint via the vector $b(\mathrm{x})$,
    which in this case is simply $\mathrm{x}$.
    \item \emph{Workspace, velocity and acceleration constraints}. These constraints impose box constraints $l_p \leq p \leq u_p, l_v \leq v \leq u_v, l_a \leq a \leq u_a$.
    \item \emph{Jerk constraints}. Jerk constraints limit how abrupt the change in accelerations can be. These are affine inequality constraints of the type $l \leq (a_i[t + 1] - a_i[t]) / h \leq u$, which we can compactly write as $l_{\text{jerk}} \leq C y \leq u_{\text{jerk}}$ for appropriate $C, l_{\text{jerk}}, u_{\text{jerk}}$.
\end{itemize}

\begin{figure}
    \centering
    \hfill
    \begin{minipage}{.32\linewidth}
        \centering
        \includegraphics[width=\linewidth]{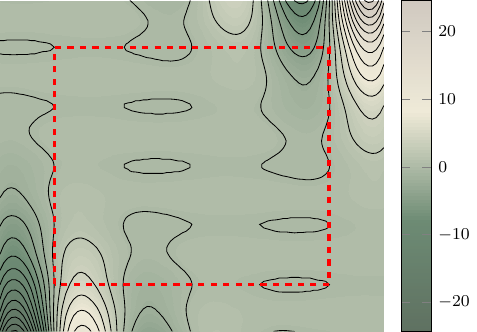}
    \end{minipage}
    \hfill
    \begin{minipage}{.32\linewidth}
        \centering
        \begin{tikzpicture}
    \node[inner sep=0] (img) 
      {\includegraphics[width=\linewidth]{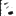}};
    \draw[red, dashed, line width=1pt] 
      (img.south west) rectangle (img.north east);
  \end{tikzpicture}
    \end{minipage}
    \hfill
    \begin{minipage}{.32\linewidth}
        \centering
        \includegraphics[width=\linewidth]{images/multi-vehicle/data/5_25_1_1_1/plot.pdf}
    \end{minipage}
    \hfill
    \caption{Preference and coverage. On the left we depict the landscape of the potential $\psi(\cdot)$ with the working space delimited by the red, dashed rectangle. In the middle we report the image used to compute the coverage score of the trajectory on the right.}
    \label{fig:objective-viz}
\end{figure}

The objective is to minimize
\[
\varphi(y) = \texttt{effort}(y) + \lambda \cdot \texttt{preference}(y) + \nu\cdot \texttt{coverage}(y)
\]
where $\texttt{effort}(y)$ describes the input effort of the solution $y$, 
$\texttt{preference}(y)$ describes the fitness of $y$ with respect to a spatial potential $\psi$, 
and $\texttt{coverage}(y)$ describes the fraction of the workspace that the agents cover over time, 
and $\lambda, \nu \geq 0$ are tuning parameters. In our experiment, we use them as binary variables to show the effects of adding certain terms to the objective function. The different terms are defined as follows:
\begin{itemize}[leftmargin=*]
    \item \emph{Input effort} (\texttt{effort}). The input effort is 
    \[
    \texttt{effort}(y)= \sum_{i = 1}^N\sum_{t = 0}^{T - 1} \norm{a_i[t]}^2.
    \]
    \item \emph{Individual contribution} (\texttt{preference}). Each vehicle tries to minimize the cumulative value 
    along the path of the scaled Ishigami potential \citep{terpin2024learning}
    \[
    \psi\left(\frac{z}{1.25}\right) = 0.05 \left(\sin(z_1) + 7 \sin(z_2)^2 + \frac{1}{10} \left(\frac{z_1 + z_2}{2}\right)^4 \sin(z_1)\right), 
    \]
    depicted on the left of \cref{fig:objective-viz}. That is, 
    \[
    \texttt{preference}(y) = \sum_{i = 1}^N\sum_{t = 0}^T \psi(p_i[t]).
    \]
    \item \emph{Fleet contribution} (\texttt{coverage}). 
    We define the coverage over time as the fraction of the space $[(p_{\min})_1,(p_{\max})_1]\times[(p_{\min})_2,(p_{\max})_2]$ 
    the vehicles sweep over during their trajectory. 
    We map the position of the $i^\mathrm{th}$ vehicle into a continuous pixel‐space of size $H\times W = 16\times 16$ via
\[
u_i = \frac{(p_i[t])_1 - (p_{\min})_1}{(p_{\max})_1-(p_{\min})_1}W,
\qquad
v_i = H - \frac{(p_i[t])_2 - (p_{\min})_2}{(p_{\max})_2-(p_{\min})_2}H.
\]
On each pixel $(u, v)$ we place a bivariate Gaussian
\[
G_i(u, v)
=
A \exp\left(
-\frac{1}{2(1-\rho^2)}
\left[
\frac{(u - u_i)^2}{\sigma_x^2}
+
\frac{(v - v_i)^2}{\sigma_y^2}
-
\frac{2\rho(u - u_i)(v - v_i)}{\sigma_x\sigma_y}
\right]
\right),
\]
where $A = 200,\sigma_x = \sigma_y = 1,\rho = 0$ are hyperparameters. We sum (and clip) the $N$ kernels to obtain a coverage image,
\[
I(u, v) = \min\left(\sum_{i=1}^N G_i(u,v), 255\right).
\]
We display $I$ for a sampled trajectory in \cref{fig:objective-viz}. Finally, the coverage score is computed as
\[
\texttt{coverage}(y)
=-
\frac{1}{HW}\sum_{p=1}^H\sum_{q=1}^W
\frac{I(p,q)}{255},
\]
measuring the fraction of pixels covered by the Gaussian footprints in a differentiable manner.

\end{itemize}

Overall, the optimization problem reads:
\begin{equation*}
\label{eq:multi-vehicle:optimization}
    \underset{ y \in \reals^{d}}{\mathclap{\mathrm{minimize}}} 
    \qquad \varphi(y)
    \qquad
\overset{\hphantom{ y \in \reals^{d}}}{\mathclap{\mathrm{subject~to}}} \qquad
\begin{bmatrix}
    A_{\text{if}}
    \\
    A_{\text{dyn}}
\end{bmatrix}
y = \begin{bmatrix}
    b(\mathrm{x})
    \\
    0
\end{bmatrix},
\quad
\begin{bmatrix}
    l_p\\l_v\\l_a
\end{bmatrix} \leq y \leq \begin{bmatrix}
    u_p\\u_v\\u_a
\end{bmatrix},
\quad 
l_{\text{jerk}}
\leq C y \leq u_{\text{jerk}}.
\end{equation*}

\begin{figure}
    \centering
    \begin{minipage}{\linewidth}
        \hfill
        \begin{minipage}{.3\linewidth}
        \centering
        \includegraphics[width=\linewidth]{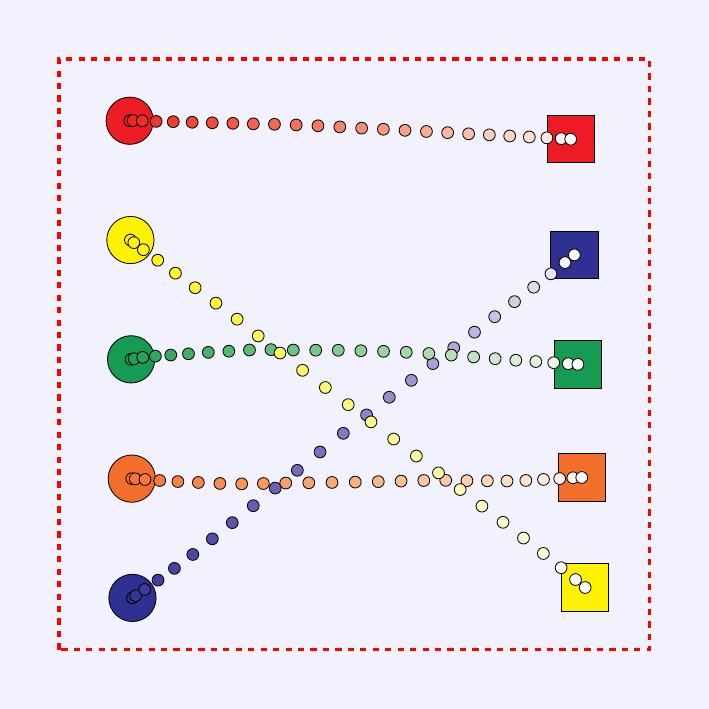}
        \end{minipage}
        \hfill
        \begin{minipage}{.3\linewidth}
        \centering
        \includegraphics[width=\linewidth]{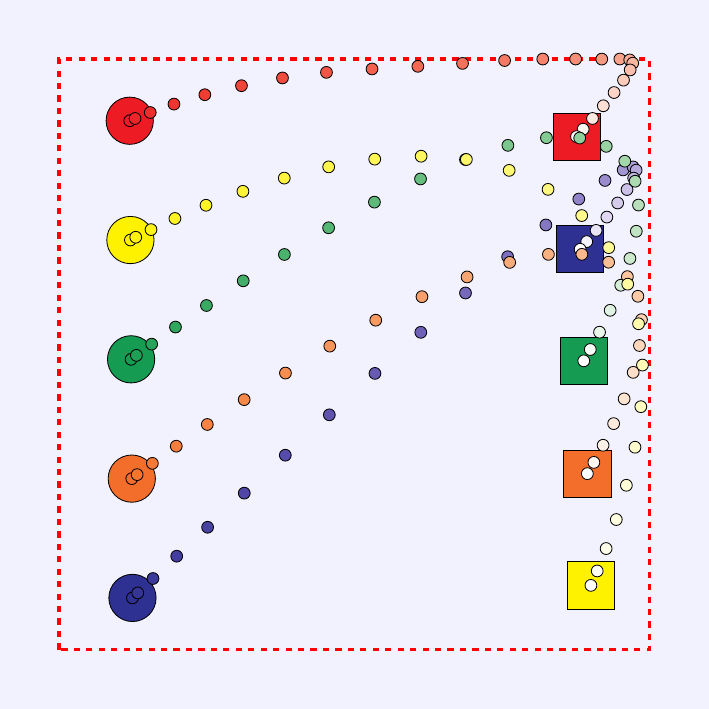}
        \end{minipage}
        \hfill
        \begin{minipage}{.3\linewidth}
        \centering
        \includegraphics[width=\linewidth]{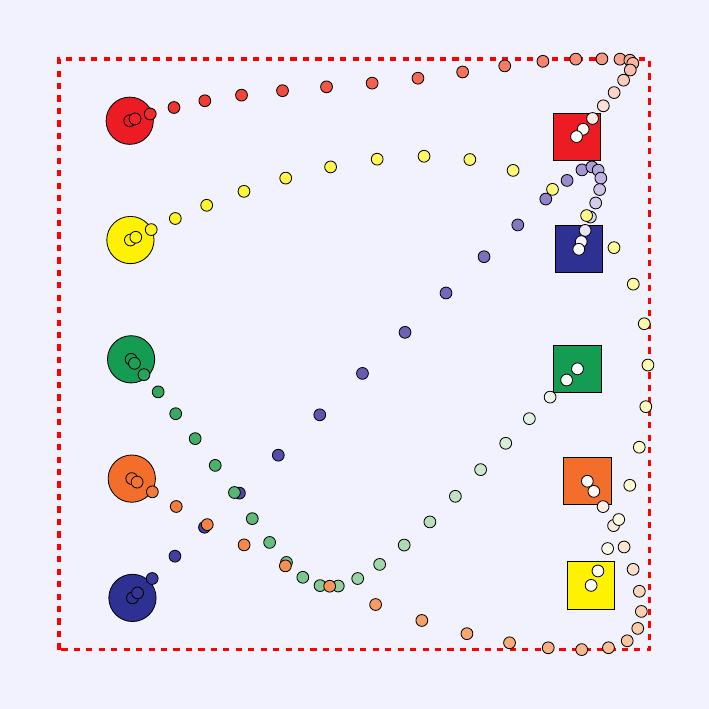}
        \end{minipage}
        \hfill
    \end{minipage}
    \begin{minipage}{\linewidth}
        \hfill
        \begin{minipage}{.3\linewidth}
        \centering
        \includegraphics[width=\linewidth]{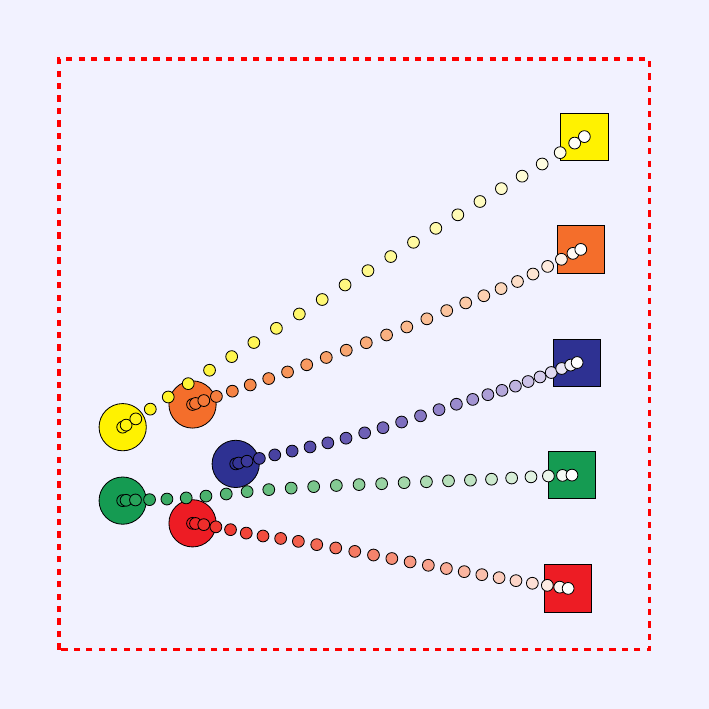}
        \end{minipage}
        \hfill
        \begin{minipage}{.3\linewidth}
        \centering
        \includegraphics[width=\linewidth]{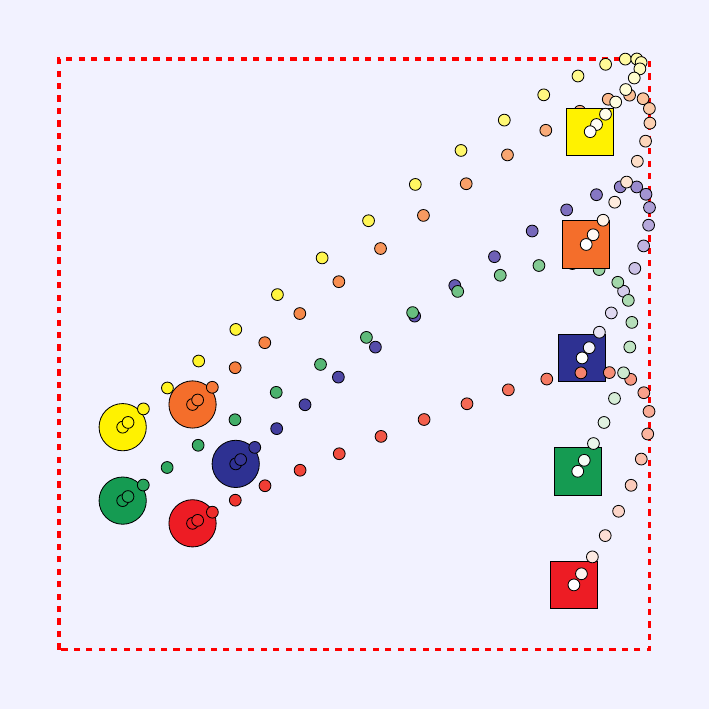}
        \end{minipage}
        \hfill
        \begin{minipage}{.3\linewidth}
        \centering
        \includegraphics[width=\linewidth]{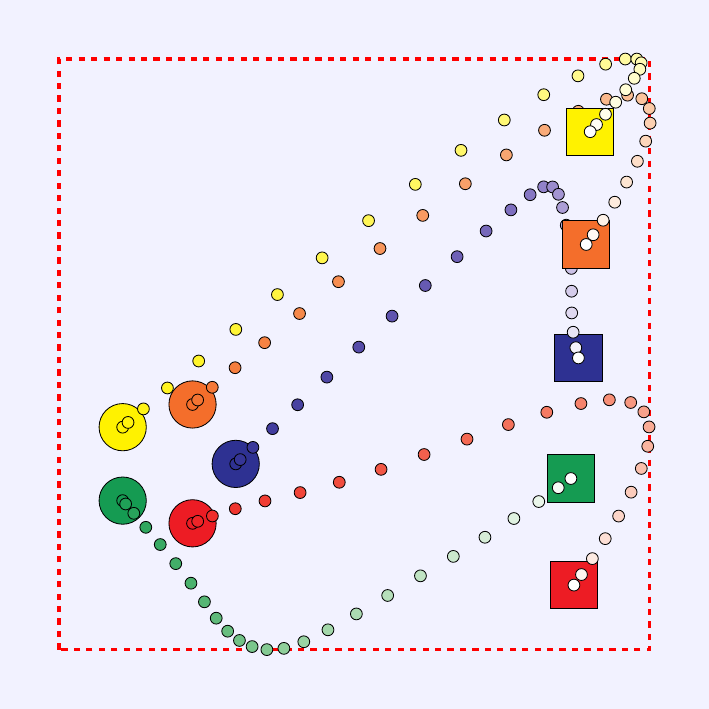}
        \end{minipage}
        \hfill
    \end{minipage}
    \begin{minipage}{\linewidth}
        \hfill
        \begin{minipage}{.3\linewidth}
        \centering
        \includegraphics[width=\linewidth]{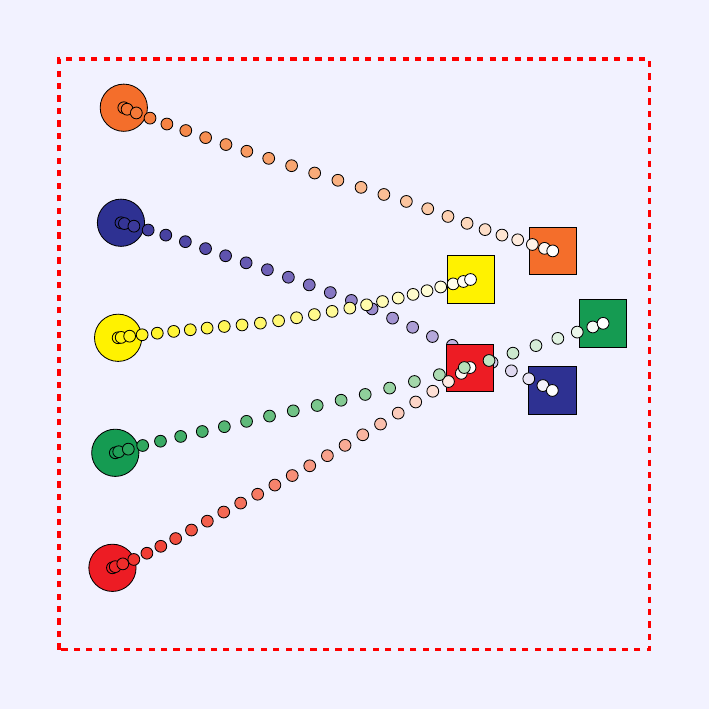}
        \end{minipage}
        \hfill
        \begin{minipage}{.3\linewidth}
        \centering
        \includegraphics[width=\linewidth]{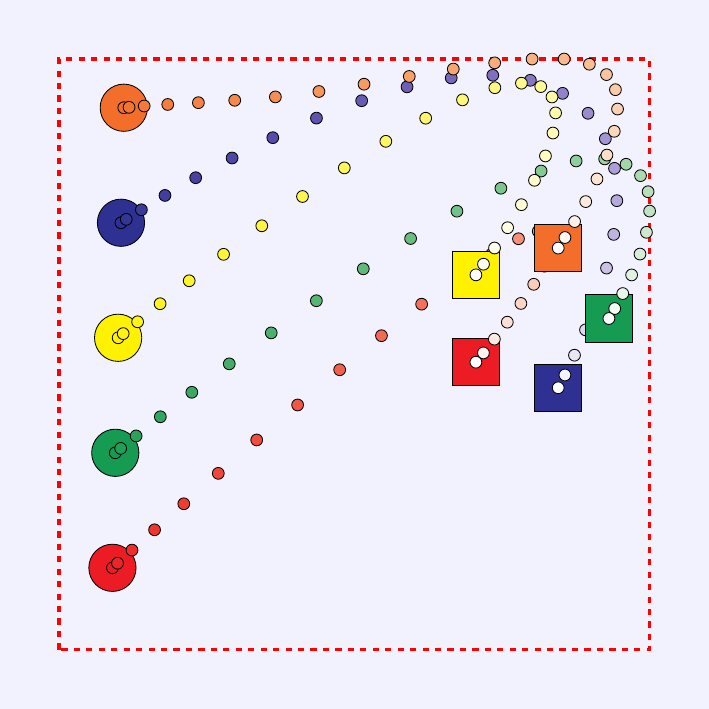}
        \end{minipage}
        \hfill
        \begin{minipage}{.3\linewidth}
        \centering
        \includegraphics[width=\linewidth]{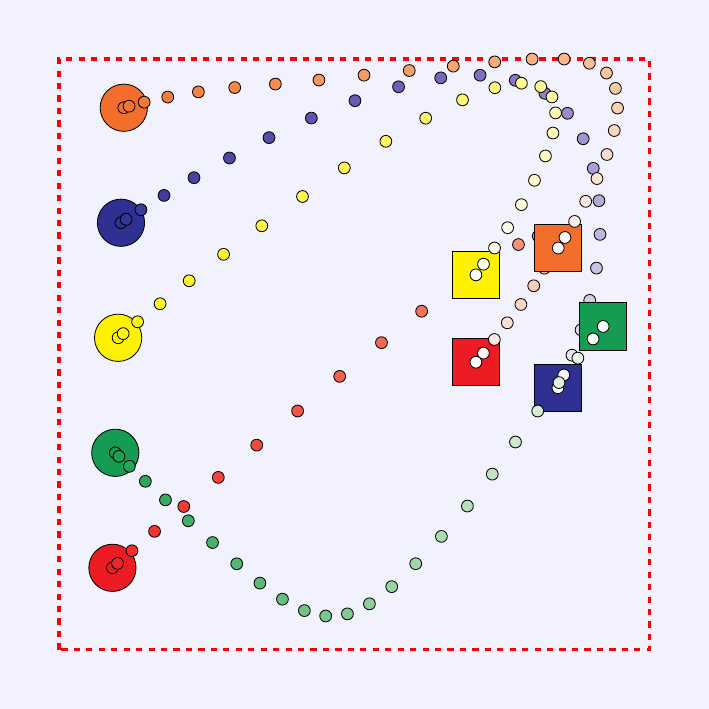}
        \end{minipage}
        \hfill
    \end{minipage}
    \begin{minipage}{\linewidth}
        \hfill
        \begin{minipage}{.3\linewidth}
        \centering
        \includegraphics[width=\linewidth]{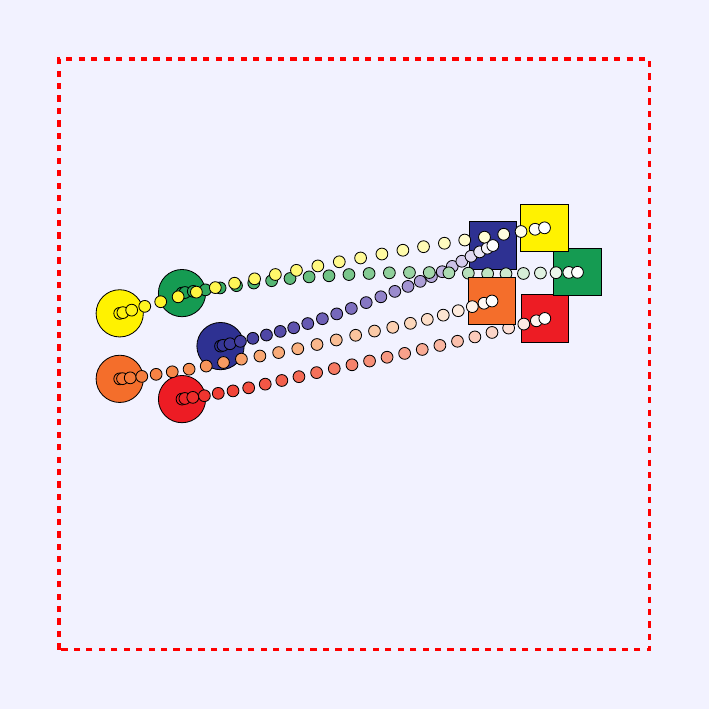}
        \end{minipage}
        \hfill
        \begin{minipage}{.3\linewidth}
        \centering
        \includegraphics[width=\linewidth]{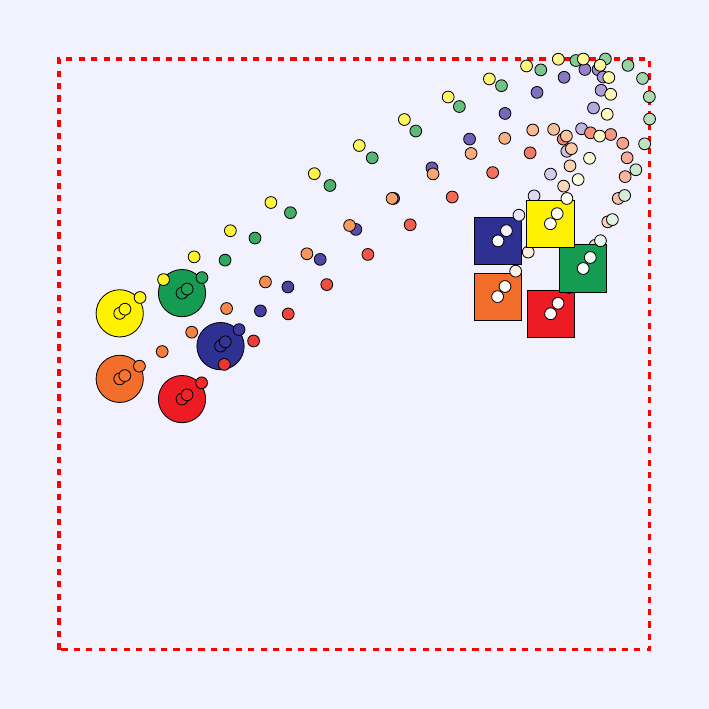}
        \end{minipage}
        \hfill
        \begin{minipage}{.3\linewidth}
        \centering
        \includegraphics[width=\linewidth]{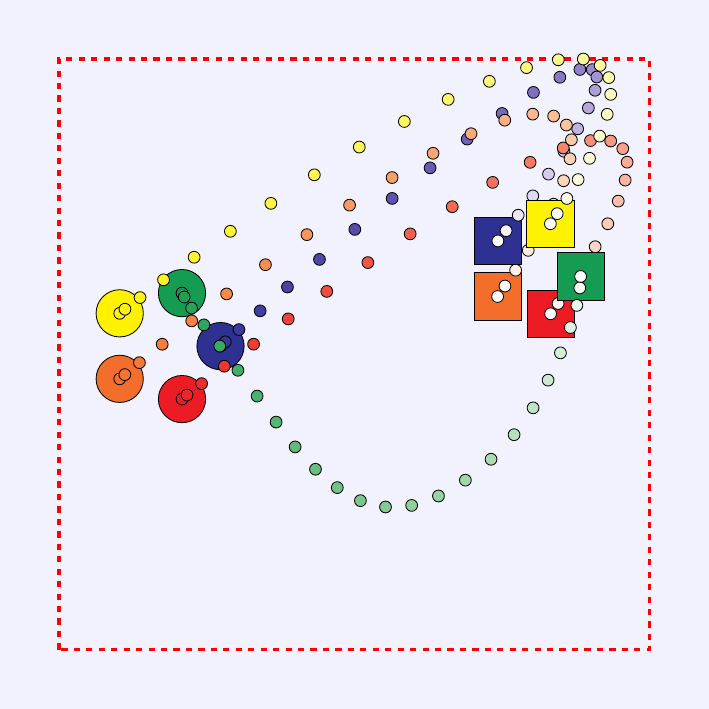}
        \end{minipage}
        \hfill  
    \end{minipage}
    \begin{minipage}{\linewidth}
        \hfill
        \begin{minipage}{.3\linewidth}
        \centering
        \includegraphics[width=\linewidth]{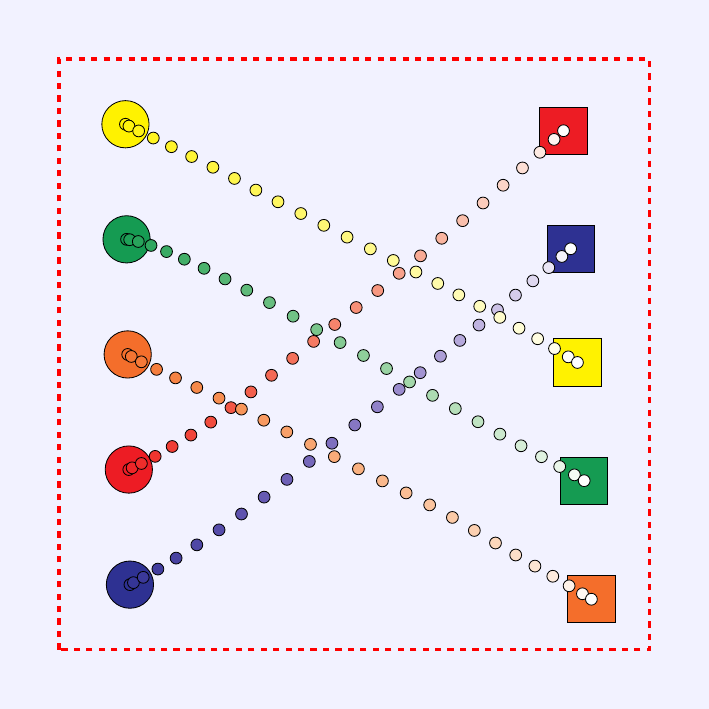}
        \end{minipage}
        \hfill
        \begin{minipage}{.3\linewidth}
        \centering
        \includegraphics[width=\linewidth]{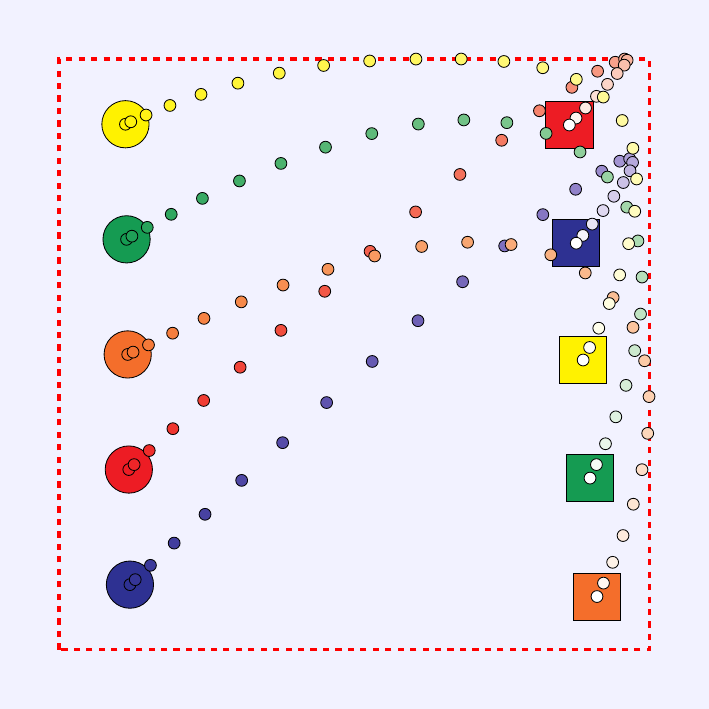}
        \end{minipage}
        \hfill
        \begin{minipage}{.3\linewidth}
        \centering
        \includegraphics[width=\linewidth]{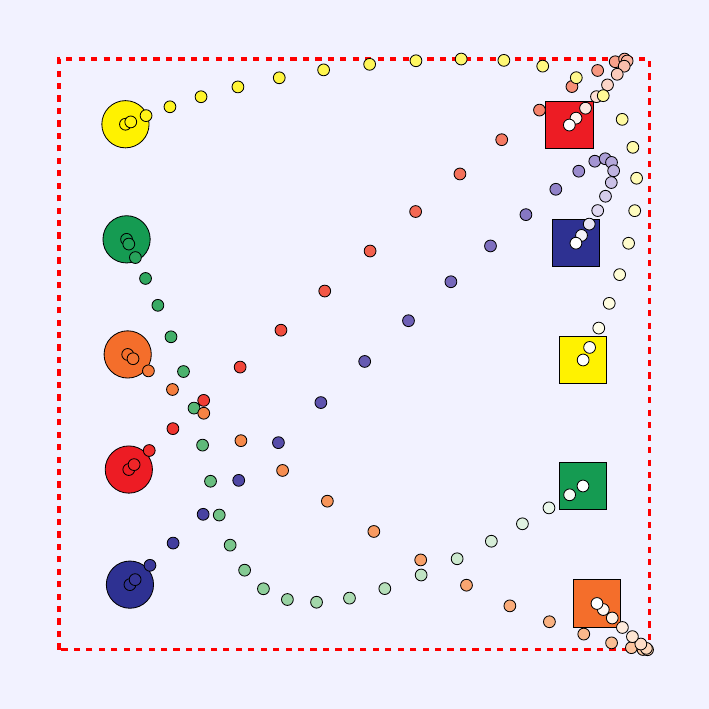}
        \end{minipage}
        \hfill
    \end{minipage}
    \caption{Collection of synthesized $5$-vehicles trajectories. Each row relates to a different instance of initial and final configurations, and from left to right we report the generated trajectory with a network trained with $(\lambda, \nu) = (0,0), (\lambda, \nu) = (1,0)$ and $(\lambda, \nu) = (1,1)$, respectively.}
    \label{fig:multi-robot:5:25}
\end{figure}

\begin{figure}
    \centering
    \begin{minipage}{\linewidth}
        \hfill
        \begin{minipage}{.3\linewidth}
        \centering
        \includegraphics[width=\linewidth]{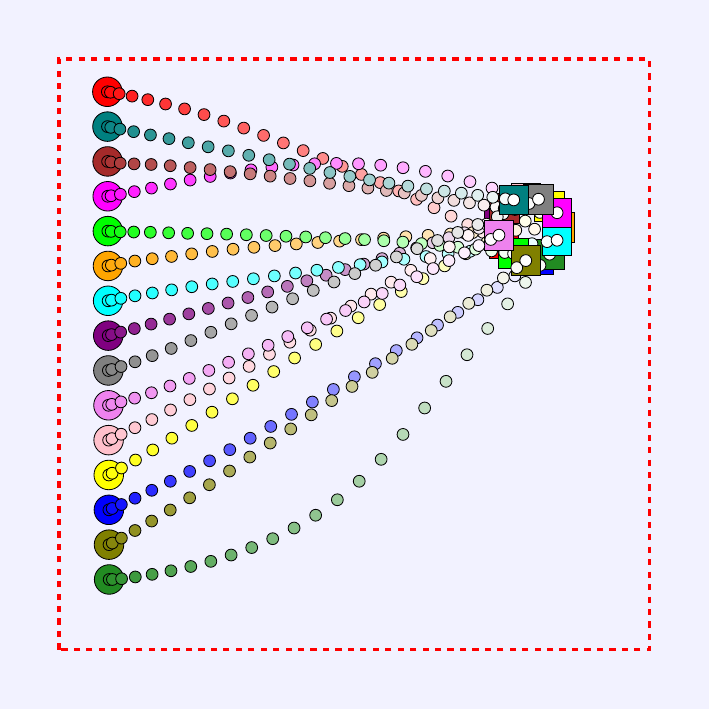}
        \end{minipage}
        \hfill
        \begin{minipage}{.3\linewidth}
        \centering
        \includegraphics[width=\linewidth]{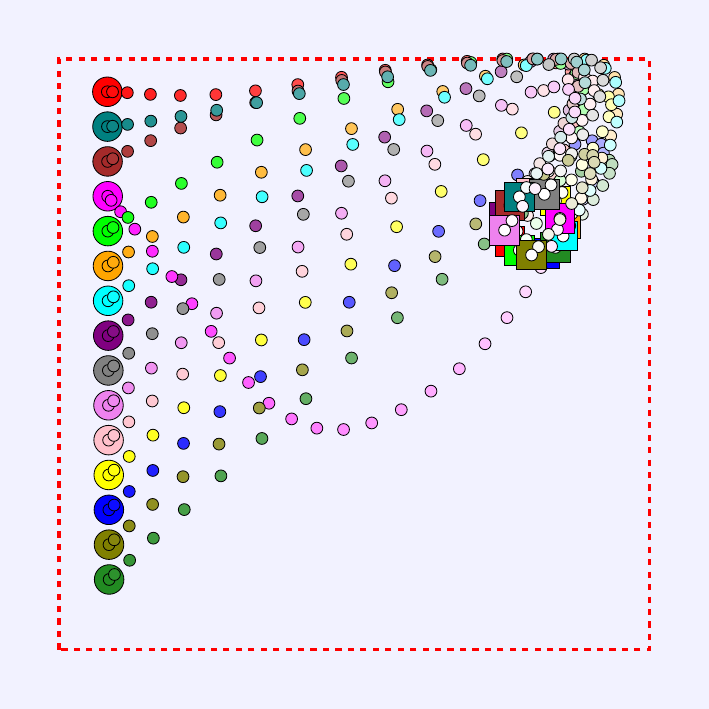}
        \end{minipage}
        \hfill
        \begin{minipage}{.3\linewidth}
        \centering
        \includegraphics[width=\linewidth]{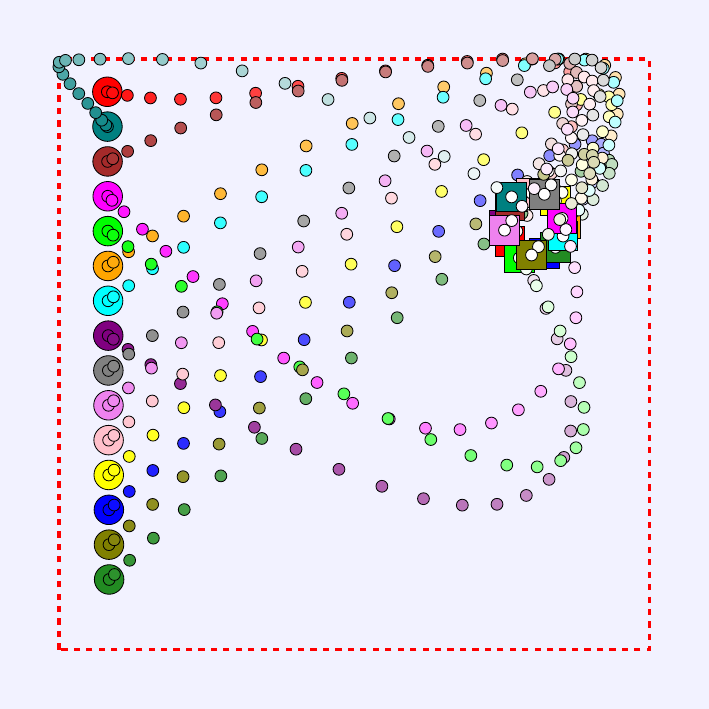}
        \end{minipage}
        \hfill
    \end{minipage}
    \begin{minipage}{\linewidth}
        \hfill
        \begin{minipage}{.3\linewidth}
        \centering
        \includegraphics[width=\linewidth]{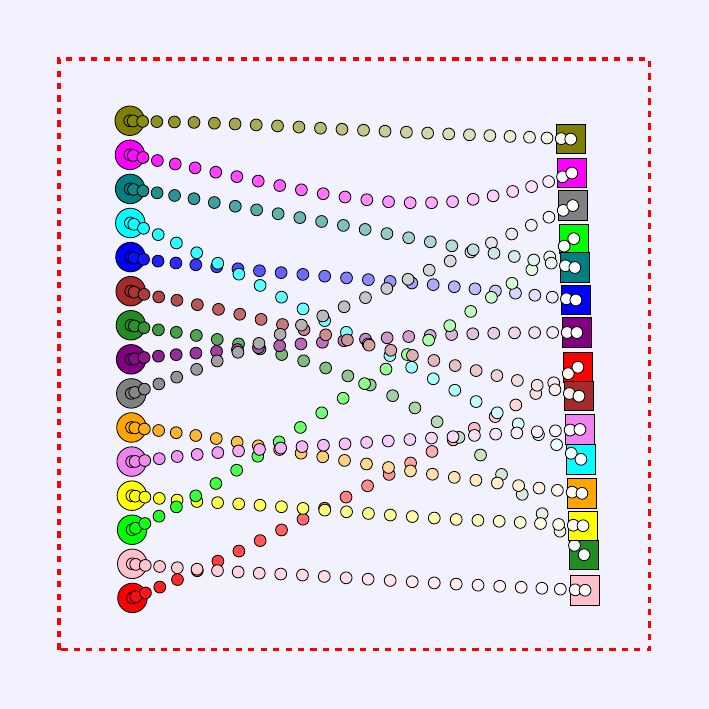}
        \end{minipage}
        \hfill
        \begin{minipage}{.3\linewidth}
        \centering
        \includegraphics[width=\linewidth]{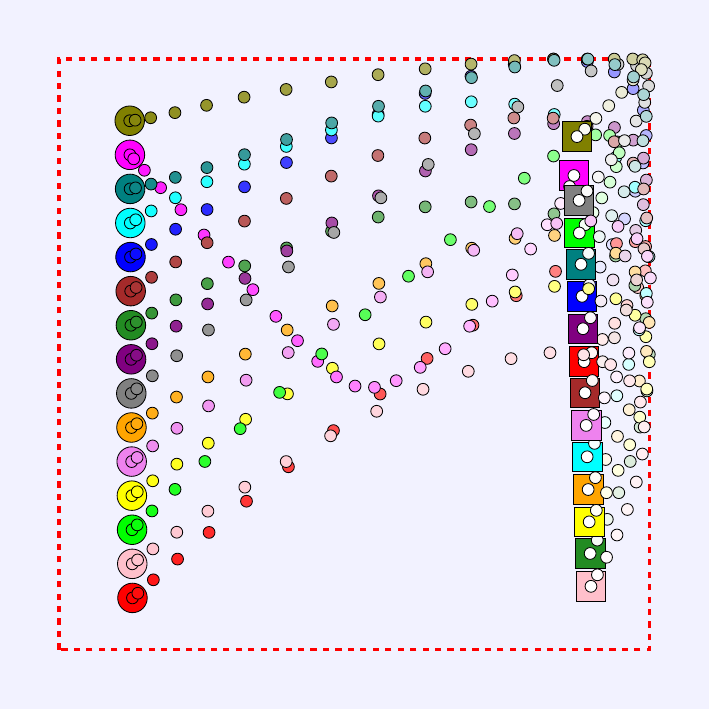}
        \end{minipage}
        \hfill
        \begin{minipage}{.3\linewidth}
        \centering
        \includegraphics[width=\linewidth]{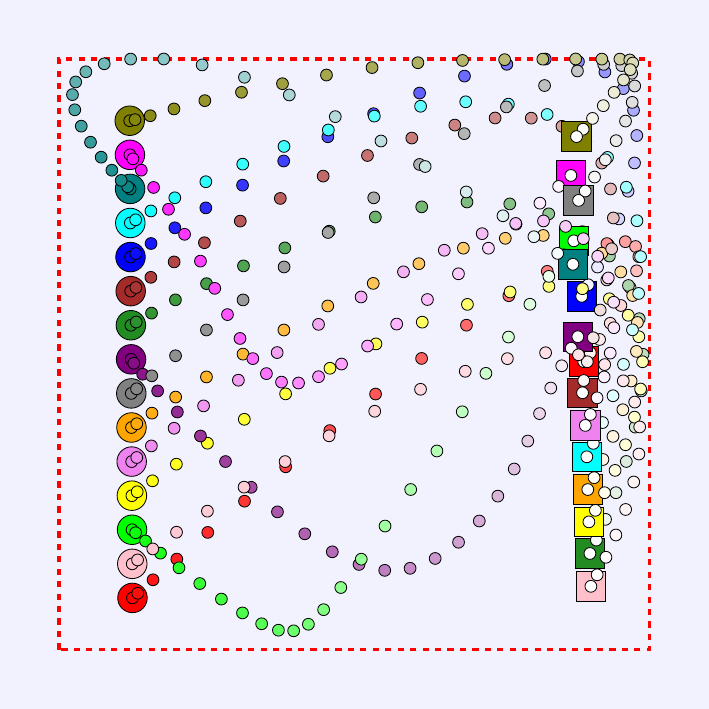}
        \end{minipage}
        \hfill
    \end{minipage}
    \begin{minipage}{\linewidth}
        \hfill
        \begin{minipage}{.3\linewidth}
        \centering
        \includegraphics[width=\linewidth]{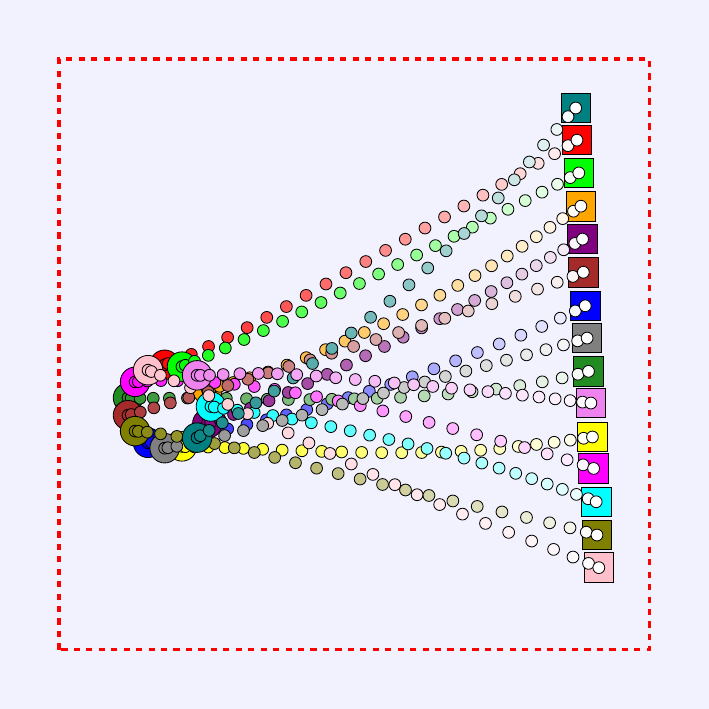}
        \end{minipage}
        \hfill
        \begin{minipage}{.3\linewidth}
        \centering
        \includegraphics[width=\linewidth]{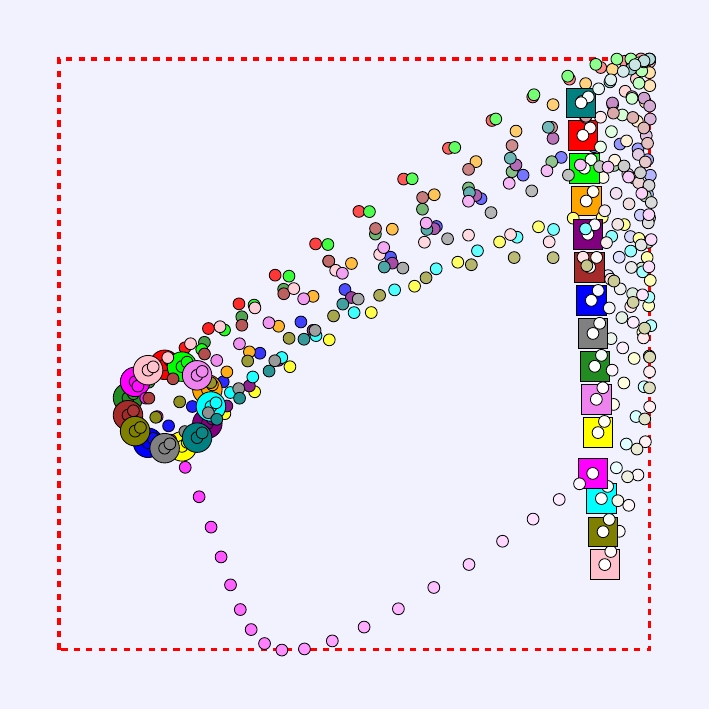}
        \end{minipage}
        \hfill
        \begin{minipage}{.3\linewidth}
        \centering
        \includegraphics[width=\linewidth]{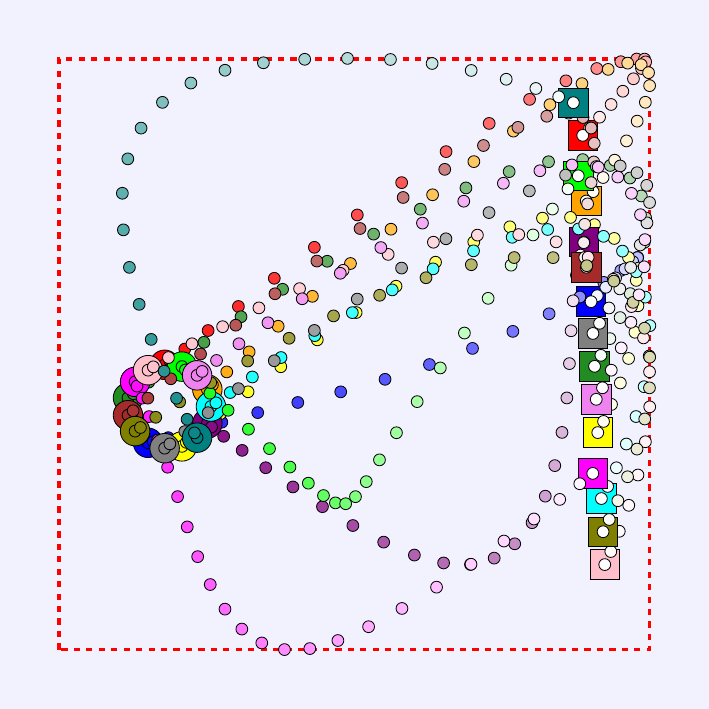}
        \end{minipage}
        \hfill
    \end{minipage}
    \begin{minipage}{\linewidth}
        \hfill
        \begin{minipage}{.3\linewidth}
        \centering
        \includegraphics[width=\linewidth]{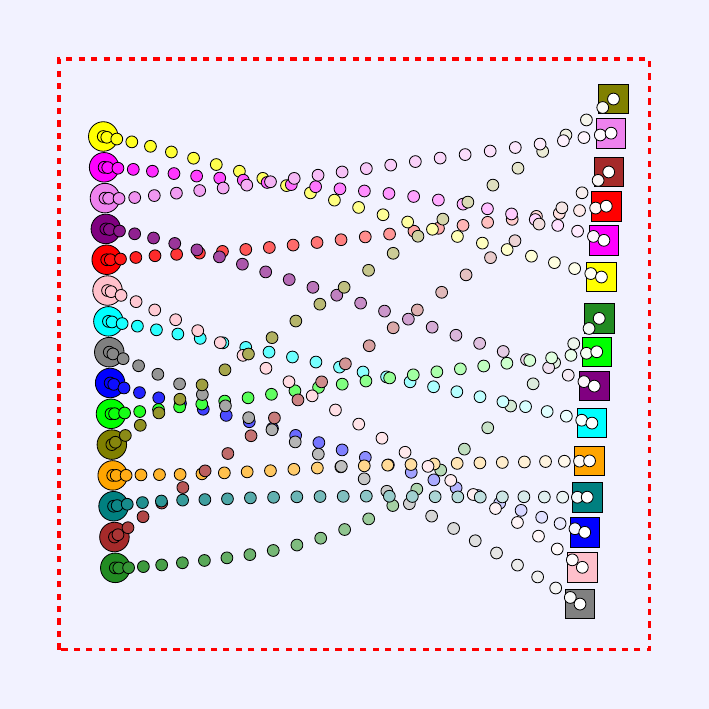}
        \end{minipage}
        \hfill
        \begin{minipage}{.3\linewidth}
        \centering
        \includegraphics[width=\linewidth]{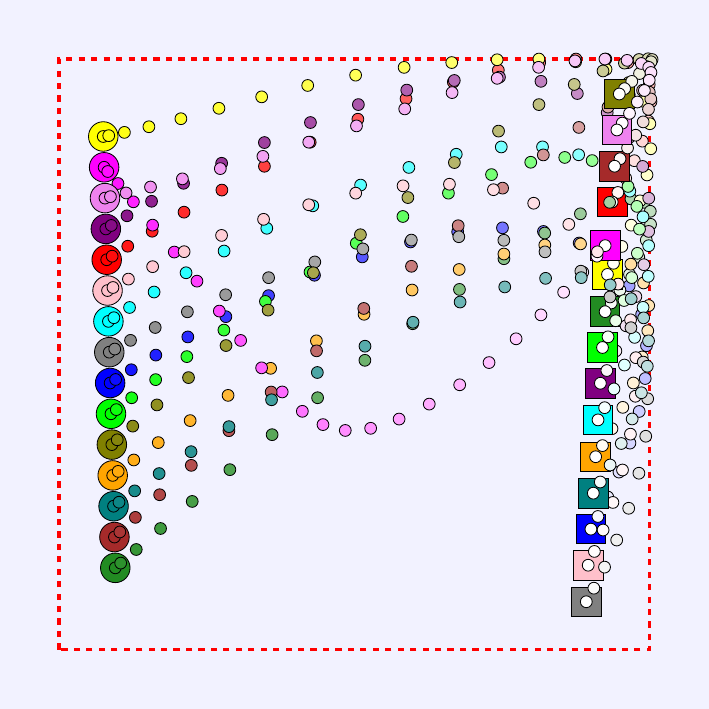}
        \end{minipage}
        \hfill
        \begin{minipage}{.3\linewidth}
        \centering
        \includegraphics[width=\linewidth]{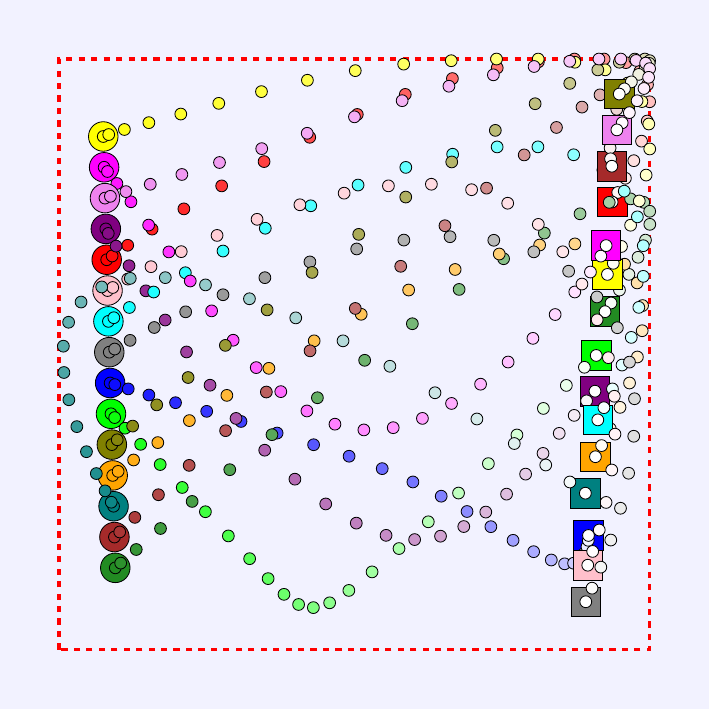}
        \end{minipage}
        \hfill  
    \end{minipage}
    \begin{minipage}{\linewidth}
        \hfill
        \begin{minipage}{.3\linewidth}
        \centering
        \includegraphics[width=\linewidth]{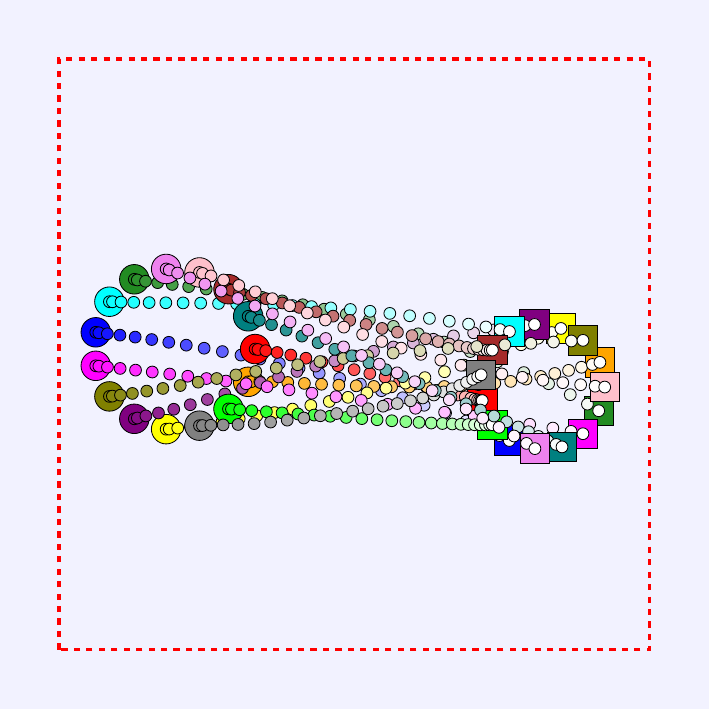}
        \end{minipage}
        \hfill
        \begin{minipage}{.3\linewidth}
        \centering
        \includegraphics[width=\linewidth]{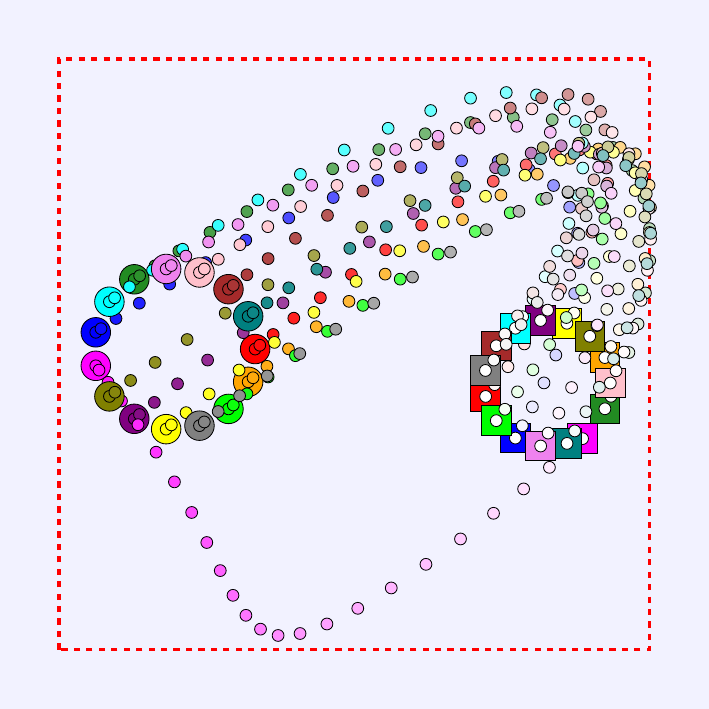}
        \end{minipage}
        \hfill
        \begin{minipage}{.3\linewidth}
        \centering
        \includegraphics[width=\linewidth]{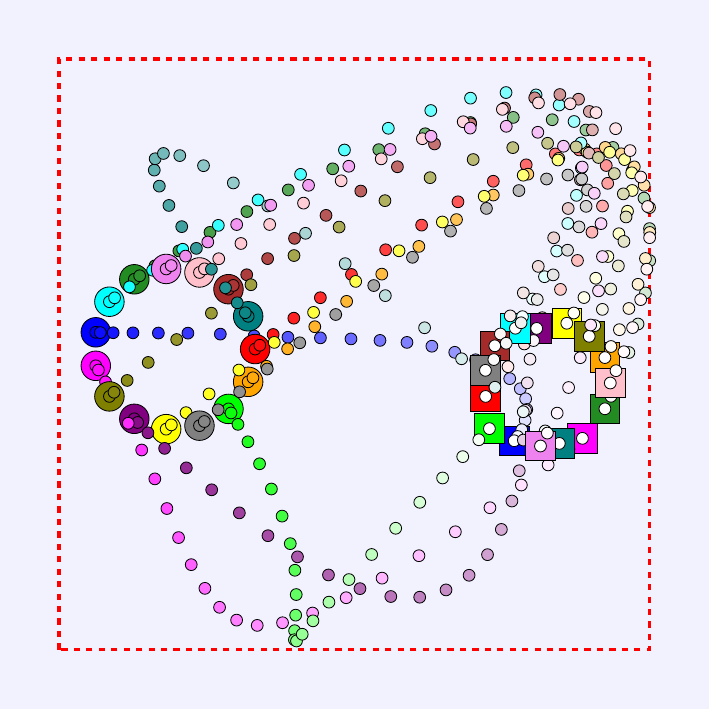}
        \end{minipage}
        \hfill
    \end{minipage}
    \caption{Collection of synthesized $15$-vehicles trajectories. Each row relates to a different instance of initial and final configurations, and from left to right we report the generated trajectory with a network trained with $(\lambda, \nu) = (0,0), (\lambda, \nu) = (1,0)$ and $(\lambda, \nu) = (1,1)$, respectively.}
    \label{fig:multi-robot:15:25}
\end{figure}

In \cref{fig:multi-robot:5:25,fig:multi-robot:15:25} we report various trajectories generated with \ouralg with different number of vehicles $N \in \{5, 15\}$ and horizon $T = 25$. These additional qualitative results show the effectiveness of \ouralg in synthesizing trajectories that optimize non-convex, fleet-level preferences also for large fleets and long horizons: in the largest setting reported here, there are $n = 3030$ optimization variables ($d = 2280$).

\subsection{Contextual coverage and long-horizon planning}
\label{sec:context-coverage}
Building on top of the trajectory planning application, we provide some preliminary evidence that \ouralg is capable of handling:
\begin{enumerate}
    \item very high-dimensional context size (in the millions of variables); and
    \item a large number of optimization variables and constraints (in the tens of thousands).
\end{enumerate}

For this, we focus on a single vehicle with the same constraints as in the previous section, increase the horizon length, and consider as cost function the average along the trajectory of the value assigned (via bilinear interpolation) to the position of the vehicle by a discrete map $\mathrm{m}$ of resolution $\texttt{res}\times\texttt{res}$.
Thus, the context $\mathrm{x}$ corresponds to the initial and final location as well as the map $\mathrm{m}$. We generate the map $\mathrm{m}$ by sampling uniformly at random one of four locations 
\[c \in \{(-2.5, -2.5), (-2.5, 2.5), (2.5, -2.5), (2.5, 2.5)\}.\]
Then, the $\texttt{res}\times\texttt{res}$ pixels comprising the map $\mathrm{m}$ correspond to a uniform discretization of the grid $[-5,5]\times[-5,5]$, with values resulting from a Gaussian with center $c$ and a standard deviation of $2$. This simple construction (in the sense that the single parameter controlling the map $\mathrm{m}$ is which $c$ was used) allows to test how \ouralg deals with backbone networks that have as input high-dimensional contexts, while keeping fixed the number of optimization variables.

For this, we use a convolutional neural network with three layers, $16$ features and downsampling factor $D$ for each layer, followed by a linear layer and a softmax, resulting in $4$ logits. These are appended to the vector input (the initial and final positions) as input to the same \gls*{mlp} used in the previous experiments.
We note that the convolutional neural network and the \gls*{mlp} are trained jointly as the backbone of \ouralg.
We consider two cases:
\begin{enumerate}[leftmargin=*]
\item $H = 100$, $\texttt{res} = 1024$, $D = 4$. 
In this case, we evaluate if \ouralg enables training end-to-end of neural networks with constraints even when the context is very high-dimensional (for a comparison, \citet{donti2021dc3} consider a context of dimension $50$).
\item $H = 750$, $\texttt{res} = 64$, $D = 2$. In this case, we evaluate \ouralg for a context of size similar to that typical of, e.g., reinforcement learning settings and a number of optimization variables and constraints two orders of magnitude larger than state-of-the-art parametric optimization algorithms (for instance, \citet{donti2021dc3} consider $100$ optimization variables, and $50$ equality and inequality constraints).
\end{enumerate}

\begin{figure}
    \centering
    \begin{minipage}{0.49\linewidth}
    \begin{tcolorbox}[
      colback=white,
      colframe=orange!20,
      arc=1mm,
      boxsep=0pt,
      top=2pt,
      left=5pt,
      right=5pt,
      bottom=2pt,
      toptitle=3pt,
      bottomtitle=2pt,
      coltitle=black,
      title={$H = 100$, $\texttt{res} = 1024$}
    ]
      \includegraphics[width=.48\linewidth]{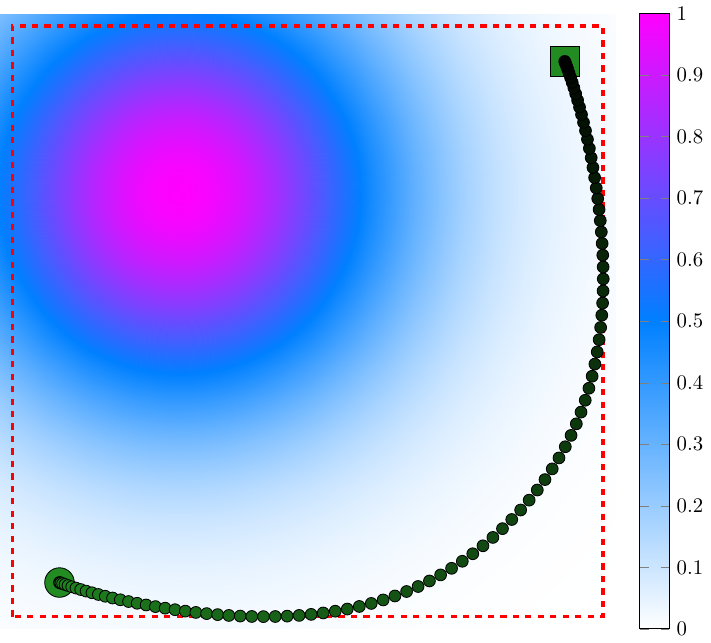}
      \hfill
      \includegraphics[width=.48\linewidth]{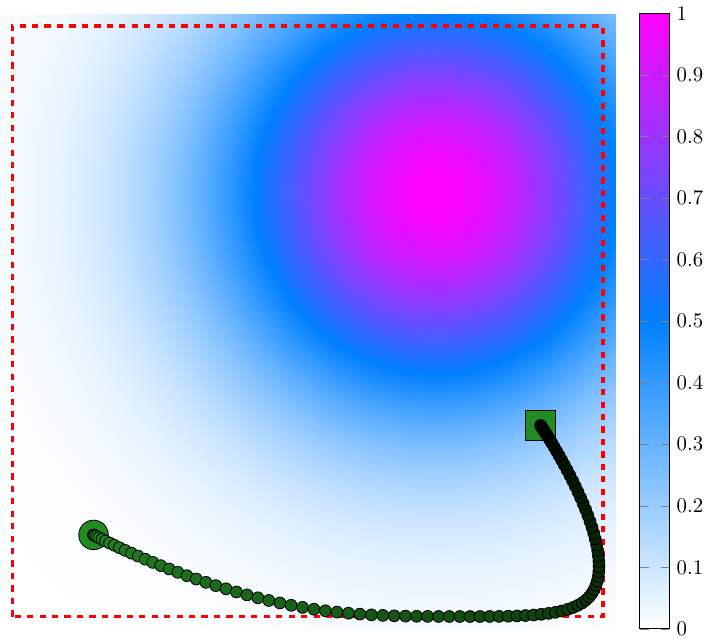}
    \end{tcolorbox}
  \end{minipage}
  \hfill
  \begin{minipage}{0.49\linewidth}
    \begin{tcolorbox}[
      colback=white,
      colframe=blue!20,
      arc=1mm,
      boxsep=0pt,
      top=2pt,
      left=5pt,
      right=5pt,
      bottom=2pt,
      toptitle=3pt,
      bottomtitle=2pt,
      coltitle=black,
      title={$H = 750$, $\texttt{res} = 64$}
    ]
      \includegraphics[width=.48\linewidth]{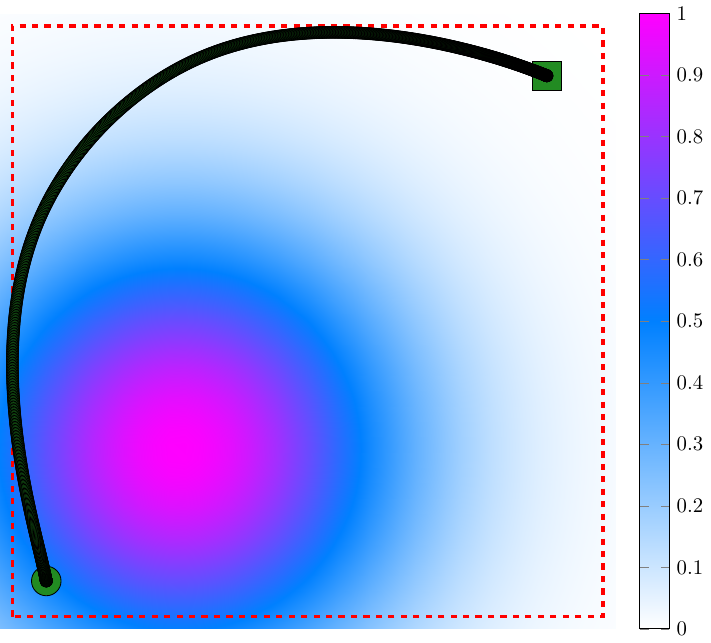}
      \hfill
      \includegraphics[width=.48\linewidth]{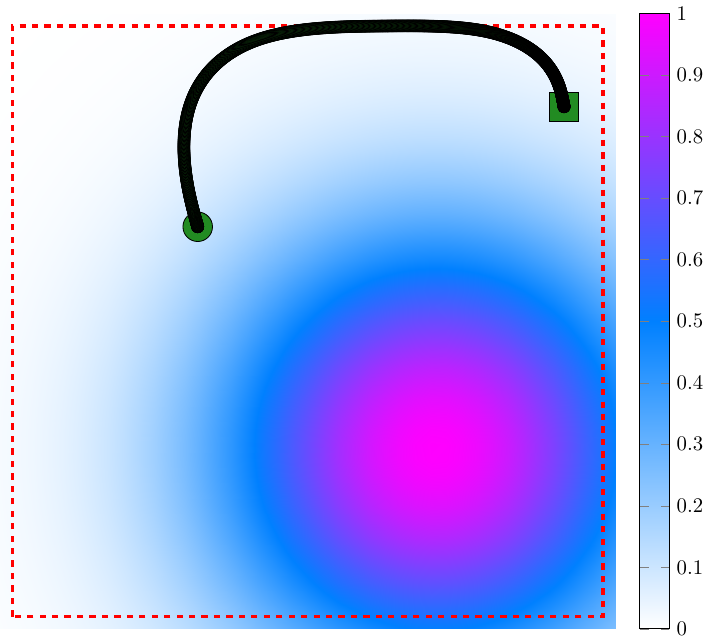}
    \end{tcolorbox}
  \end{minipage}
    \caption{Examples of resulting trajectories for the experiments in \cref{sec:context-coverage}.}
    \label{fig:context-coverage}
\end{figure}

In both settings, \ouralg successfully learns to solve the planning task, as qualitatively shown in \cref{fig:context-coverage}. Despite the simplicity of the maps used, we believe that these examples show the promise of using \ouralg for high-dimensional optimization problems with very large contexts.

\subsection{Second order cone constraints.}
\label{sec:soc}

In this section, we provide evidence that \ouralg can be used to solve parametric problems with second-order cone constraints. By doing so, we also elaborate on how one can introduce additional constraint types into \ouralg effortlessly.

\paragraph{Experimental setup.}
We focus on problems of the form:
\begin{equation}
\label{eq:soc}
\underset{ y = (y_1, y_2) }{\mathrm{minimize}}
\quad c^\top y_1
    \quad
\mathrm{subject~to} \quad
A y_1 + y_2
= b, ~
y_2 \in \mathcal{K}_1',
\end{equation}
where $A \in \reals^{d_2 \times d_1}, c, y_1 \in \reals^{d_1}, y_2, b \in \reals^{d_2},
\mathcal{K}_1' = \{
y_2 \in \mathbb{R}^{d_2} \,|\, \norm{y_{2,:-1}}_2 \leq y_{2,-1}
\}
$ is a second-order cone, and $y_{2,:-1}$ denotes the first $d_2 - 1$ entries of $y_2$, and $y_{2,-1}$ the last one.
We make the following two observations:
\begin{itemize}[leftmargin=*]
    \item First, the feasible set is non-linear and possibly unbounded. Thus, methods like GLinSAT \citep{zeng2024glinsatgenerallinearsatisfiability} are not applicable.
    \item Second, one cannot train an unconstrained network and then apply the projection layer, since the optimization objective is linear and, thus, the network would diverge. See also \cref{sec:constraints-training}.
\end{itemize}

To generate the problem data, we follow the procedure described by \citet[Section 6.6]{o2016conic}, reported here for completeness and to clarify the elements of the random parametric problems we consider. First, we generate a random matrix $A \in \reals^{d_2 \times d_1}$ sampling uniformly at random in $[-1,1]^{d_2\times d_1}$. The matrix $A$ is fixed for all the problems of a training run, i.e., it is not a context/parameter of the problem. In other words, $A$ is context-independent. 
Given $A$, we generate a batch of $B$ problems as follows:
\begin{enumerate}[leftmargin=*]
    \item We sample $B$ random vectors $z$ uniformly in $[-1,1]^{d_2}$ and $B$ random primal solutions $y_1^\star \in \reals^{d_1}$.
    \item We project $z$ onto the second-order cone to obtain $y_2^\star = \Pi_{\mathcal{K}_1'}(z)$.
    \item We set $b = Ay_1^\star + y_2^\star$ and $c = -A^\top (y_2^\star - z)$.
\end{enumerate}
The optimal value of the problem is then given by $c^\top y_1^\star$, and the context $\mathrm{x}$ (i.e., the input to the \gls*{nn}), is $\mathrm{x} = (b, c)$. \reviewerOne{For \ouralg, we} employ the same \gls*{mlp} used in \cref{sec:experiments:toy-example} followed by the \ouralg projection layer, so that the optimal output should yield $\hat{y} = (\hat{y}_1, \hat{y}_2)$ with $c^\top \hat{y}_1 \approx c^\top y_1^\star$. For the experiments in this section, we use \reviewerOne{$(d_1, d_2) \in \{(25, 25), (500, 500)\}$ (we refer to these configurations as small and large, respectively) and $B = 128$}. We train the network for $1000$ epochs with a different batch at each epoch, and then generate an additional one for evaluation at the end of the training, on which we calculate the statistics. \reviewerOne{We compare the performance of our method against:
\begin{itemize}[leftmargin=*]
    \item \texttt{cvxpylayers} \cite{agrawal2019differentiable}, which we use as a replacement of our custom projection layer. This is the same procedure we adopted in \cref{sec:experiments:benchmarks} for \texttt{JAXopt}. Here, we consider \texttt{cvxpylayers} because \texttt{JAXopt} does not support second-order cone constraints.
    \item \texttt{SCS} \cite{o2016conic}, a traditional first-order solver for second-order cone programs. To make our comparison with \texttt{SCS} as fair and comprehensive as possible, we perform two benchmarks. On one, we interface \texttt{SCS} through the very popular parser \texttt{CVXPY} \citep{diamond2016cvxpy} using its Disciplined Parametrized Programming functionality, as many users would do. On the other benchmark, we interface \texttt{SCS} directly and exploited its native parametric programming functionality, as more advanced users would.
\end{itemize}
\begin{remark}
It is important to note that with these benchmarks we are exploiting the full parametric capabilities of existing solvers, which is not always the case by end-users or even other hard-constrained \gls*{nn} benchmarks.
\end{remark}
Similarly to the linearly-constrained benchmarks in \cref{sec:experiments:benchmarks}, we consider learning curves, \texttt{RS}, \texttt{CV}, and inference times.
}

\paragraph{Integrating \texorpdfstring{\eqref{eq:soc}}{the second-order cone constraints} into \texorpdfstring{\ouralg}{Pinet}.}
Equivalently, we can write \eqref{eq:soc} as:
\begin{equation}
\label{eq:soc:2}
\underset{ y }{\mathrm{minimize}}
\quad \begin{bmatrix}c^\top & 0\end{bmatrix} y
    \quad
\mathrm{subject~to} \quad
\begin{bmatrix}A & I\end{bmatrix} y = b, \quad y \in \mathcal{K}_1,
\end{equation}
with $\mathcal{K}_1 = \reals^n \times \mathcal{K}_1'$, $\mathcal{K}_1'$ being the second-order cone, and without $\mathcal{K}_2$ since we do not use any auxiliary variable (i.e., $n = d = d_1 + d_2$). 
We recall here that $\mathcal{K}_1$ and $\mathcal{K}_2$ refer to the problem formulation in \cref{sec:projection-layer}.
In view of \cref{appendix:extra-sets}, this amounts to a lifted formulation of second-order cones and is thus without loss of generality. Now, \eqref{eq:soc:2} is in our standard formulation and we can apply \cref{alg:os-project}. In particular, with the notation used in \cref{alg:os-project}, we only need to define the projection
\[
t_{k + 1} = \Pi_{\mathcal{K}_1}\left(\frac{2z_{k + 1,1} - s_{k,1} + 2\sigma y_{\text{raw},1}}{1 + 2\sigma}\right).
\]
Denoting with $t_{k +1, :d_1}, z_{k + 1,1, :d_1}, s_{k,1, :d_1}, y_{\text{raw}, :d_1}$ the first $d_1$ entries of the vectors $t_{k + 1}, z_{k + 1,1}, s_{k,1}, y_{\text{raw},1}$ and with $t_{k + 1, d_1:}, z_{k + 1,1, d_1:}, s_{k,1, d_1:}, y_{\text{raw}, d_1:}$ the remaining $d_2$ entries, we have 
\begin{align*}
    & t_{k + 1, :d_1} = \frac{2z_{k + 1,1, :d_1} - s_{k,1, :d_1} + 2\sigma y_{\text{raw},1, :d_1}}{1 + 2\sigma}
    \quad\text{and}\quad \\
    & t_{k + 1, d_1:} = \Pi_{\mathcal{K}_1'}\left(\frac{2z_{k + 1,2, d_1:} - s_{k,2, d_1:} + 2\sigma y_{\text{raw},2, d_1:}}{1 + 2\sigma}\right).
\end{align*}
The projection $\Pi_{\mathcal{K}_1'}(\cdot)$ onto the second-order cone admits a closed form expression; see, e.g., the work of \citet{busseti2019solution}.

\begin{figure}
    \centering
    \includegraphics[width=\linewidth]{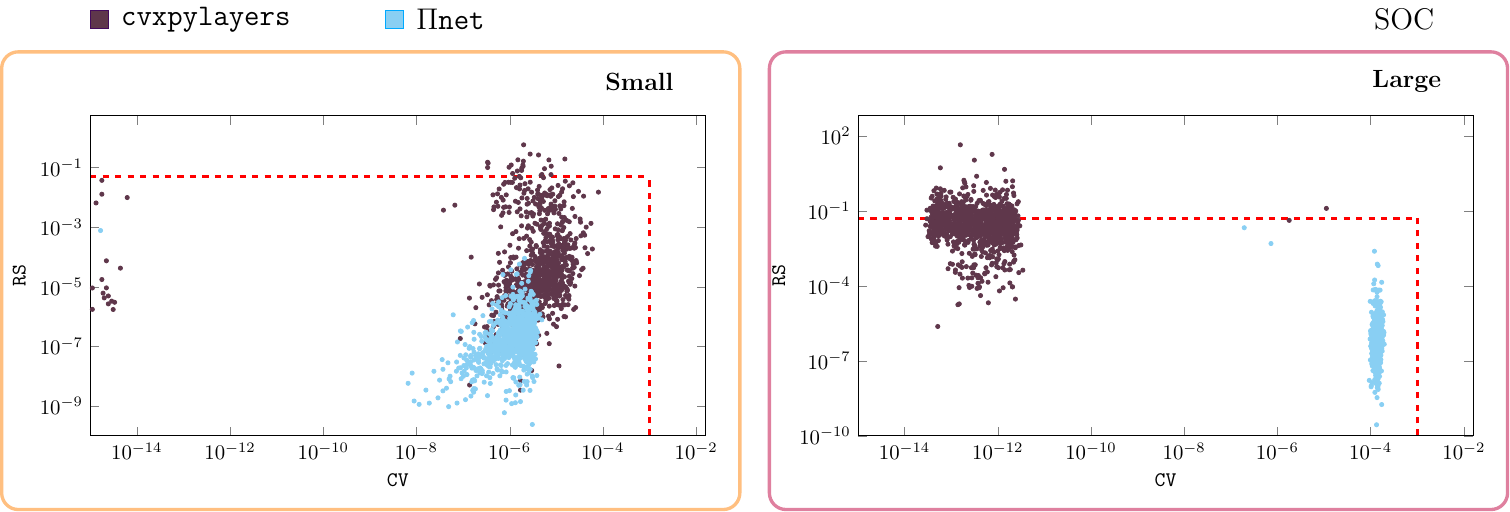}
    \caption{\reviewerOne{Scatter plots of \texttt{RS} and \texttt{CV} on the second-order cone programs on the test set. 
    The red dashed lines show the thresholds to consider a candidate solution optimal.}}
    \label{fig:benchmark:soc:rs-cv}
\end{figure}

\begin{figure}
    \centering
    \includegraphics[width=\linewidth]{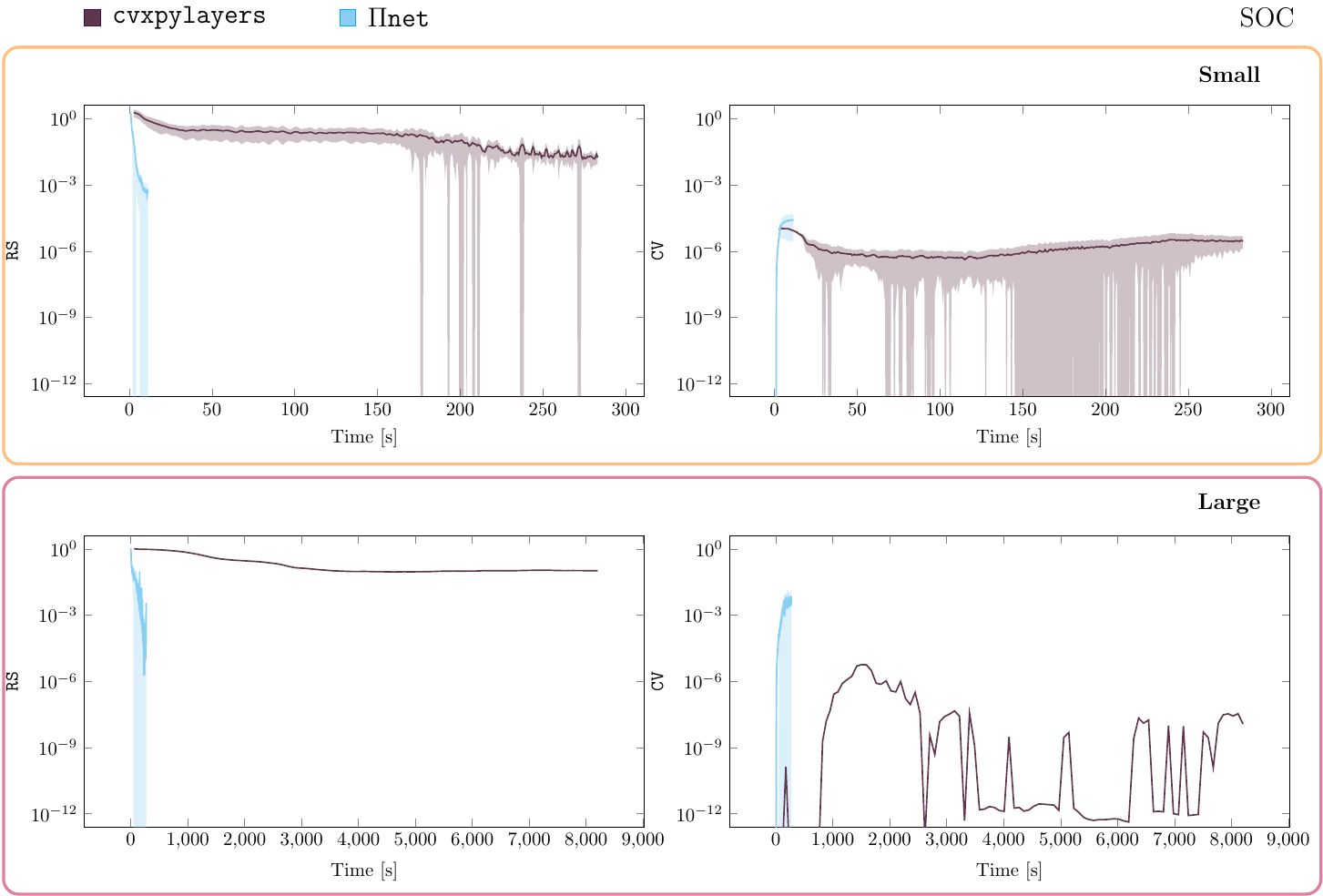}
    \caption{\reviewerOne{Comparison of the learning curves in terms of average \texttt{RS} and \texttt{CV} on the validation set,
    on the second-order cone programs. 
    The solid lines denote the mean and the shaded area the standard deviation across 5 seeds.}}
    \label{fig:benchmark:soc:training-curves}
\end{figure}

\paragraph{Results.} \reviewerOne{In \cref{fig:benchmark:soc:rs-cv} we report the \texttt{RS} and \texttt{CV} for each problem instance in the test set.
Analogously to the linearly constrained benchmarks in \cref{sec:experiments:benchmarks}, we consider a candidate solution to be optimal if the condition $\texttt{CV} \leq 10^{-3}$ and $\texttt{RS} \leq 5\%$
is satisfied. We report training curves in \cref{fig:benchmark:soc:training-curves} and inference times in \cref{tab:benchmark:soc:times}.}

\begin{table}
    \centering
    \begin{adjustbox}{width=\textwidth,center}
        \begin{tabular}{l|ccccc|ccccc}
\multirow{2}{*}{\textbf{Method}} & 
\multicolumn{5}{c|}{\textbf{Single inference [s]}} & 
\multicolumn{5}{c}{\textbf{Batch inference [s]}} \\
\cline{2-11}
 & median & LQ & UQ & $\min$ & $\max$ & median & LQ & UQ & $\min$ & $\max$ \\
 \hline
\multicolumn{11}{l}{\textbf{Small second-order cone programs}} \\\hline
\texttt{CVXPY} & 0.00076 & 0.00074 & 0.00078 & 0.00071 & 0.00084 
& 0.59085 & 0.58784 & 0.59399 & 0.57668 & 0.70665 \\
\texttt{SCS} & 0.00005 & 0.00004 & 0.00005 & 0.00004 & 0.00005
& 0.04612 & 0.04580 & 0.04652 & 0.04483 & 0.04926
\\
\texttt{cvxpylayers} & 0.00857 & 0.00813 & 0.01270 & 0.00706 & 0.06567 
 & 1.82553 & 1.68570 & 4.47413 & 1.65853 & 5.73859
\\
\rowcolor{blue!10}\ouralg & 0.00905 & 0.00884 & 0.00928 & 0.00848 & 0.01137 
& 0.02265 & 0.02245 & 0.02292 & 0.02202 & 0.02415 \\
\hline
\multicolumn{11}{l}{\textbf{Large second-order cone programs}} \\\hline
\texttt{CVXPY} & 0.06721 & 0.06658 & 0.06743 & 0.06578 & 0.06899 
& 62.70289 & 62.65441 & 62.76826 & 62.46146 & 63.31168 \\
\texttt{SCS} & 0.01072 & 0.01049 & 0.01078 & 0.00976 & 0.01155 
& 10.97265 & 10.80431 & 11.10839 & 10.68313 & 26.18852 \\
\texttt{cvxpylayers} & 2.36724 & 2.29826 & 2.41810 & 2.18466 & 4.04885
& 342.35211 & 341.85465 & 343.06660 & 341.65177 & 344.11561 \\
\rowcolor{blue!10}\ouralg & 0.01219 & 0.01187 & 0.01356 & 0.01113 & 0.04383 
& 0.90982 & 0.90932 & 0.91134 & 0.90844 & 0.94499 \\
\end{tabular}
\end{adjustbox}
    \caption{\reviewerOne{Inference time comparison for single-instance and batch-instance (1024 problems) settings across different methods,
    evaluated on the second-order cone programs. 
    The table reports median runtime along with statistical descriptors: 
    lower quartile (LQ, 25th percentile), upper quartile (UQ, 75th percentile), 
    $\min$ and $\max$ of the runtime.}}
    \label{tab:benchmark:soc:times}
\end{table}

\reviewerOne{Compared to \texttt{cvxpylayers}, \ouralg can train and perform inference significantly faster, even orders of magnitude for large problems. Moreover, \ouralg provides solutions with both lower \texttt{RS} and \texttt{CV}.
Compared to \texttt{SCS}, \ouralg is faster at inference for a batch of small problems, and both a single and a batch of large problems. Instead, \texttt{SCS} is faster for single small problems.
}

\clearpage
\section{The sharp bits}
\label{sec:sharp-bits}
Numerical optimization is an art of details, and this section describes the sharp bits of \ouralg, which are fundamental to make \glspl*{hcnn} work reliably.
\subsection{Matrix equilibration}
\label{sec:equilibration}
The precision and convergence rates of both the forward (\cref{sec:forward-pass}) and backward (\cref{sec:backward-pass}) passes 
in our projection layer are heavily influenced, via \eqref{eq:equality_constraint} and \eqref{eq:linear-system}, 
by the \emph{condition number} of the matrix $A(\mathrm{x})$ in the equality constraint 
$A(\mathrm{x}) \begin{bmatrix}
    y \\ y_\text{aux}
\end{bmatrix} = b(\mathrm{x})$ \citep{wathen2015preconditioning}. The condition number of a matrix $A$ (we drop the dependency on $\mathrm{x}$ for clarity of exposition) is defined as the ratio of its maximum singular value to its minimum singular value, and the procedure of \emph{decreasing} this ratio for improving numerical performance is called \emph{preconditioning} or \emph{equilibration}. 

To achieve this, we seek a \emph{diagonal scaling} \citep{wathen2015preconditioning}. 
The idea is that to solve $A\begin{bmatrix}
    y \\ y_\text{aux}
\end{bmatrix} = b$, one can instead solve $D_r A D_c \begin{bmatrix}
    \tilde{y} \\ \tilde{y}_\text{aux}
\end{bmatrix} = D_rb$ for invertible matrices $D_r \in \mathbb{R}^{m\times m},D_c \in \mathbb{R}^{n\times n}$ 
and recover the solution as $\begin{bmatrix}
    y \\ y_\text{aux}
\end{bmatrix} = D_c \begin{bmatrix}
    \tilde{y} \\ \tilde{y}_\text{aux}
\end{bmatrix} $. 
We implement a modified version of Ruiz's equilibration \citep{wathen2015preconditioning} which is outlined in \cref{algorithm:equilibration}.
In our numerical experiments we use $K = 25$, $\varepsilon = 10^{-3}$ and the Gauss-Seidel update mode.

\begin{figure}[ht]
  \begin{myalgorithm}[Modified Ruiz's equilibration] \label{algorithm:equilibration}%
        \textbf{Inputs}: $A \in \reals^{m \times n}$, maximum iterations $K$, tolerance $\varepsilon$,
            update mode (Gauss-Seidel or Jacobi) \\
            $D_r \gets I_m$,
            $D_c \gets I_n$ \\
            $A_\text{scaled} \gets A$ \\
            $
            \textbf{For $k=1$ to $K$:}  \\
            \left\lfloor \\
            \begin{array}{l}
                \textbf{If Gauss-Seidel update:} \\
                \left|
                \begin{array}{l}
                    \text{Compute row norms $d_{r,i} = \| (A_{\text{scaled}})_{i,:} \|_2$ for all $i$} \\
                    \text{$D_r \gets \mathrm{diag}(1 / \sqrt{d_{r,1}}, \dots, 1 / \sqrt{d_{r,m}}) \cdot D_r$} \\
                    \text{$A_{\text{scaled}} \gets \mathrm{diag}(1 / \sqrt{d_{r,1}}, \dots, 1 / \sqrt{d_{r,m}}) \cdot A_{\text{scaled}}$} \\
                    \text{Compute column norms $d_{c,i} = \| (A_{\text{scaled}})_{:,i} \|_2$ for all $i$} \\
                    \text{$D_c \gets D_c \cdot \mathrm{diag}(1 / \sqrt{d_{c,1}}, \dots, 1 / \sqrt{d_{c,n}})$} \\
                    \text{$A_{\text{scaled}} \gets A_{\text{scaled}} \cdot \mathrm{diag}(1 / \sqrt{d_{c,1}}, \dots, 1 / \sqrt{d_{c,n}})$}
                \end{array}
                \right. \\
                \textbf{Else:} \\
                \left|
                \begin{array}{l}
                    \text{Compute row norms $d_{r,i}  = \| (A_{\text{scaled}})_{i,:} \|_2$ for all $i$} \\
                    \text{Compute column norms $d_{c,i}  = \| (A_{\text{scaled}})_{:,i} \|_2$ for all $i$} \\
                    \text{$D_r \gets \mathrm{diag}(1 / \sqrt{d_{r,1}}, \dots, 1 / \sqrt{d_{r,m}}) \cdot D_r$} \\
                    \text{$D_c \gets D_c \cdot \mathrm{diag}(1 / \sqrt{d_{c,1}}, \dots, 1 / \sqrt{d_{c,n}})$} \\
                    \text{$A_{\text{scaled}} \gets \mathrm{diag}(1 / \sqrt{d_{r,1}}, \dots, 1 / \sqrt{d_{r,m}}) 
                    \cdot A_{\text{scaled}} \cdot \mathrm{diag}(1 / \sqrt{d_{c,1}}, \dots, 1 / \sqrt{d_{c,n}})$}
                \end{array}
                \right. \\[.5em] 
                \text{Compute row norms $d_{r,i}  = \| (A_{\text{scaled}})_{i,:} \|_2$ for all $i$} \\
                \text{Compute column norms $d_{c,i}  = \| (A_{\text{scaled}})_{:,i} \|_2$ for all $i$} \\
                \textbf{If } (1 - \min(d_{r,i}) / \max(d_{r,i})) < \varepsilon \text{ and } (1 - \min(d_{c,i}) / \max(d_{c,i})) < \varepsilon\textbf{:} \\
                \left| 
                \begin{array}{l}
                    \textbf{Return } D_r, D_c
                \end{array}
                \right.
            \end{array} \\[.5em]
            \right. \\
            \textbf{Return } D_r, D_c
        $
  \end{myalgorithm}
\end{figure}
Next, we describe how the matrix equilibration affects the Douglas-Rachford algorithm
in the case of polytopic constraint sets. 
To start, we observe that $\mathcal{K}$ is a box constraint and, thus, can be written as 
\[
\mathcal{K} 
= 
\mathcal{K}_1\times\mathcal{K}_2 
=
\mathcal{K}_{1,1}\times\ldots\times\mathcal{K}_{1,d} 
\times  
\mathcal{K}_{2,1}\times\ldots\times\mathcal{K}_{2,n-d}.
\] 
Then, we rewrite \eqref{eq:projection_in_lifted_form} in terms of the new coordinates:
\[
\begin{bmatrix}
    \tilde{y}\\\tilde{y}_{\text{aux}}
\end{bmatrix}
=
D_c^{-1}
\begin{bmatrix}
    {y}\\{y}_{\text{aux}}
\end{bmatrix}
=
\begin{bmatrix}
    D_{c,1}^{-1} & \\
    & D_{c,2}^{-1}
\end{bmatrix}
\begin{bmatrix}
    {y}\\{y}_{\text{aux}}
\end{bmatrix}.
\]
That is, since 
\[
\begin{aligned}
\begin{bmatrix}
y\\y_{\text{aux}}
\end{bmatrix}
\in \mathcal{A}
\iff
\begin{bmatrix}
    \tilde{y}\\\tilde{y}_{\text{aux}}
\end{bmatrix}
\in \tilde{\mathcal{A}}
&=
\left\{
v \st D_r A D_c v = D_r b
\right\}
=
\left\{
v \st \tilde{A} v = \tilde{b}
\right\}
\\
\begin{bmatrix}
y\\y_{\text{aux}}
\end{bmatrix}
\in \mathcal{K}
\iff
\begin{bmatrix}
    \tilde{y}\\\tilde{y}_{\text{aux}}
\end{bmatrix}
\in \tilde{\mathcal{K}}
&=
d_{c,1}^{-1}\mathcal{K}_{1,1}
\times
\ldots
\times
d_{c,d}^{-1}\mathcal{K}_{1,d}
\\&\times
d_{c,d+1}^{-1}\mathcal{K}_{2,1}
\times
\ldots
\times
d_{c,n}^{-1}\mathcal{K}_{2,n-d}
\\&=
\tilde{\mathcal{K}}_{1,1}
\times
\ldots
\times
\tilde{\mathcal{K}}_{1,d}
\times
\tilde{\mathcal{K}}_{2,1}
\times
\ldots
\times
\tilde{\mathcal{K}}_{2,n-d}
\\&= \tilde{\mathcal{K}}_1 \times \tilde{\mathcal{K}}_2,
\end{aligned}
\]
we seek to solve
\begin{equation} \label{eq:projection_in_lifted_form:2}
    (\Pi_{\mathcal{C}}(y_{\text{raw}}), y_{\text{aux}}^{\star}) 
    = D_c\cdot\underset{ \tilde{y}, \tilde{y}_{\text{aux}} }{\mathrm{argmin}}
    \bigg\{\norm{D_{c,1}\tilde{y} - y_{\text{raw}}}^2
        + \mathcal{I}_{\tilde{\mathcal{A}}}\left(\begin{bmatrix}
    \tilde{y}\\\tilde{y}_{\text{aux}}
\end{bmatrix}\right)
+
\mathcal{I}_{\tilde{\mathcal{K}}}\left(
\begin{bmatrix}
    \tilde{y}\\\tilde{y}_{\text{aux}}
\end{bmatrix}
\right)
\bigg\}.
\end{equation}%

Then, we split the objective function as follows
\[
\tilde{g}\left(\begin{bmatrix}
    \tilde{y}\\\tilde{y}_{\text{aux}}
\end{bmatrix}\right)
=
\mathcal{I}_{\tilde{\mathcal{A}}}\left(\begin{bmatrix}
    \tilde{y}\\\tilde{y}_{\text{aux}}
\end{bmatrix}\right)
\quad\text{and}\quad
\tilde{h}\left(\begin{bmatrix}
    \tilde{y}\\\tilde{y}_{\text{aux}}
\end{bmatrix}\right) = \norm{D_{c,1}\tilde{y} - y_{\text{raw}}}^2
    + \mathcal{I}_{\tilde{\mathcal{K}}}\left(\begin{bmatrix}
    \tilde{y}\\\tilde{y}_{\text{aux}}
\end{bmatrix}\right)
\]
and by applying the Douglas-Rachford algorithm we obtain
the fixed-point iteration
\begin{subequations} \label{eq:dra_for_projection:2}
\begin{empheq}[left = {k = 0, 1, \ldots, K-1}\quad \empheqlfloor]{align}
\tilde{z}_{k + 1} &= 
\begin{bmatrix}
    \tilde{z}_{k,1}\\\tilde{z}_{k,2}
\end{bmatrix}
= \mathrm{prox}_{\sigma \tilde{g}}(s_k) = \Pi_{\tilde{\mathcal{A}}}(\tilde{s}_k)
\label{eq:equality_constraint:2}
\\
\tilde{t}_{k + 1} &= \mathrm{prox}_{\sigma \tilde{h}}(2\tilde{z}_{k + 1} - \tilde{s}_k) \overset{}{=} \begin{bmatrix}
    \diamondsuit
    \\
    \Pi_{\tilde{\mathcal{K}}_2}(2\tilde{z}_{k + 1,2} - \tilde{s}_{k,2})
\end{bmatrix}
\label{eq:inequality_projection:2}
\\
\tilde{s}_{k + 1} &= \begin{bmatrix}
\tilde{s}_{k + 1, 1}\\\tilde{s}_{k + 1,2}
\end{bmatrix} = \tilde{s}_k + \omega (\tilde{t}_{k + 1} - \tilde{z}_{k + 1})
\label{eq:dra_governing_update:2}
\end{empheq}
\end{subequations}
where $\diamondsuit = \begin{bmatrix}
    \ldots  & \diamondsuit_i & \ldots
\end{bmatrix}^\top$, with (using the notation $(v)_i$ to indicate the $i^{\mathrm{th}}$ entry of the vector $v$)
\[
\diamondsuit_i = 
\Pi_{\tilde{\mathcal{K}}_{1,i}}\left(\frac{
2\sigma d_{c,i}y_{\text{raw}}
+
2(\tilde{z}_{k+1,1})_i - (\tilde{s}_{k,1})_i
}{1 + 2\sigma d_{c,i}^2}\right)
.
\]
The equilibration effectively only changes \eqref{eq:inequality_projection:2}. 
To prove the update in \eqref{eq:inequality_projection:2}, we note that equilibration does not affect the solution of the proximal in  \eqref{eq:inequality_projection:2}
with respect to the auxiliary variables:
\begin{align*}
\mathrm{prox}_{\sigma \tilde{h}}\left(
    2\tilde{z}_{k + 1} - \tilde{s}_{k}
\right)
&=
\underset{\tilde{v}, \tilde{v}_{\text{aux}}}{\argmin}
\;
\norm{D_{c,1} \tilde{v} - y_{\text{raw}}}^2
+
\mathcal{I}_{\tilde{\mathcal{K}}}\left(\begin{bmatrix}
    \tilde{v}\\\tilde{v}_{\text{aux}}
\end{bmatrix}\right)
+
\frac{1}{2\sigma}
\norm{\begin{bmatrix}
    \tilde{v}\\\tilde{v}_{\text{aux}}
\end{bmatrix}
-
2\tilde{z}_{k + 1} + \tilde{s}_k
}^2
\\
&=
\begin{bmatrix}
  \underset{\tilde{v} \in \tilde{\mathcal{K}}_1}{\argmin}
\;
\norm{D_{c,1}\tilde{v} - y_{\text{raw}}}^2
+
\frac{1}{2\sigma}
\norm{\tilde{v}
-
2\tilde{z}_{{k + 1},1} + \tilde{s}_{k,1}
}^2  
\\
\underset{\tilde{v}_{\text{aux}} \in \tilde{\mathcal{K}}_2}{\argmin}
\;
\frac{1}{2\sigma}
\norm{
    \tilde{v}_{\text{aux}}
-
2\tilde{z}_{{k + 1},2} + \tilde{s}_{k,2}
}^2
\end{bmatrix}
\\
&=
\begin{bmatrix}
\diamondsuit
\\
\Pi_{\tilde{\mathcal{K}}_2}(2\tilde{z}_{{k + 1},2} - \tilde{s}_{k,2})
\end{bmatrix}.
\end{align*}
In particular, since
\[
l \leq y_{\text{aux}} \leq u 
\iff 
D_{c,2}^{-1}l \leq D_{c,2}^{-1}y_{\text{aux}} \leq D_{c,2}^{-1}u
\iff
D_{c,2}^{-1}l \leq \tilde{y}_{\text{aux}} \leq D_{c,2}^{-1}u
\]
the projection onto $\tilde{\mathcal{K}}_2$ remains a projection onto a box,
but we modify the upper and lower bounds of the box according to $D_{c,2}$.
We can write $\diamondsuit $, by completing the square, as
\[
\underset{\tilde{v} \in \tilde{\mathcal{K}}_1}{\argmin}
\;
\norm{\left(I_d + 2\sigma D_{c,1}^2\right)\tilde{v} - 
\left(
2\sigma D_{c,1}y_{\text{raw}}
+
2\tilde{z}_{{k + 1},1} - \tilde{s}_{k,1}
\right)
}^2
\]
and, thus,
\begin{align*} 
\diamondsuit_i &=
\underset{\tilde{v} \in \tilde{\mathcal{K}}_{1, i}}{\argmin}
\;
\norm{\left(1 + 2\sigma d_{c,i}^2\right)\tilde{v} - 
\left(
2\sigma d_{c,i}y_{\text{raw}}
+
2(\tilde{z}_{{k + 1},1})_i - (\tilde{s}_{k,1})_i
\right)
}^2
\\&=
\underset{\tilde{v} \in \tilde{\mathcal{K}}_{1, i}}{\argmin}
\;
\norm{\tilde{v} - 
\frac{
2\sigma d_{c,i}y_{\text{raw}}
+
2(\tilde{z}_{{k + 1},1})_i - (\tilde{s}_{k,1})_i
}{1 + 2\sigma d_{c,i}^2}
}^2
\\&=
\Pi_{\tilde{\mathcal{K}}_{1,i}}\left(\frac{
2\sigma d_{c,i}y_{\text{raw}}
+
2(\tilde{z}_{{k + 1},1})_i - (\tilde{s}_{k,1})_i
}{1 + 2\sigma d_{c,i}^2}\right).
\end{align*}

Although we described the case with box constraints, our changes can be adapted for other constraint classes such as second-order cone constraints 
by equilibrating equally variables that are coupled by the constraints. We leave these modifications for future work.

\subsection{Auto-tune procedure}
\label{sec:autotuning}
During the auto-tuning procedure, we will tune the hyperparameters \texttt{sigma} and \texttt{n\_iter\_fwd}.
For the remaining ones, we set \texttt{omega} and \texttt{n\_iter\_bwd} to their respective default values $1.7$ and $25$.
For simplicity, we set \texttt{n\_iter\_test} to be equal to \texttt{n\_iter\_fwd}.

For the auto-tuning, we consider a subset of the validation set consisting of $150$ contexts $\mathrm{x}$ out of the $1024$
used in the benchmark problems of \cref{sec:experiments:benchmarks}.
Then, we generate $150$ corresponding points-to-be-projected from a standard normal distribution $\mathcal{N}(0, I)$, 
which will serve as a proxy for the infeasible output of the backbone \gls{nn}.
First, to tune \texttt{sigma} we generate 100 logarithmically-spaced number between $10^{-3}$ and $5.05$.
Then, we compute the projection for each \texttt{sigma} on the $150$ context--infeasible point pairs, by running
$100$ iterations of the projection layer.
We evaluate the quality of each \texttt{sigma} by computing the resulting maximum constraint violation and
average relative suboptimality of the projection, i.e., $\norm{z_K - y_\text{raw}}/\norm{\Pi_{\mathcal{C}(\mathrm{x})}(y_\text{raw}) - y_\text{raw}}$.
We consider any \texttt{sigma} for which both metrics are below certain thresholds to be candidate hyperparameters.
We choose among the candidate hyperparameters the one with the minimum constraint violation.

We use the same procedure to choose \texttt{n\_iter\_fwd} from the set $\{50, 100, \ldots, 350, 400\}$,
where the projection is evaluated using the previously determined \texttt{sigma}.

\subsection{Ablations.}
\label{sec:the-sharp-bits:ablation}
We evaluate the effectiveness of the matrix equilibration and auto-tuning by comparing the performance
of our method with and without the different components on the large non-convex problem of \cref{sec:experiments:benchmarks}.  
We test 3 different configurations: default hyperparameters and no equilibration; auto-tuned hyperparameters and no equilibration;
auto-tuned hyperparameters and equilibration.
We respectively refer to these configurations as \texttt{Default}, \texttt{Auto} and \ouralg.
On this problem setting, performing auto-tuning takes roughly 2 minutes, while the equilibration takes less than 1 second.
The (tuned) hyperparameters for each configuration are as follows:
\begin{align*}
    \texttt{Default} & \leftarrow \texttt{sigma} = 1.0, ~\texttt{n\_iter\_fwd} = 100 \\
    \texttt{Auto} & \leftarrow \texttt{sigma} = 3.28, ~\texttt{n\_iter\_fwd} = 350 \\
    \text{\ouralg} & \leftarrow \texttt{sigma} = 0.161, ~\texttt{n\_iter\_fwd} = 50
\end{align*}
and the remaining ones are chosen as discussed previously.
We note that the large difference in the value of \texttt{sigma} between \texttt{Auto} and \ouralg
is due to the equilibration that changes the problem scaling.

\begin{figure}
    \centering
    \begin{minipage}{0.5\linewidth}
    \centering
        \includegraphics[width=\linewidth]{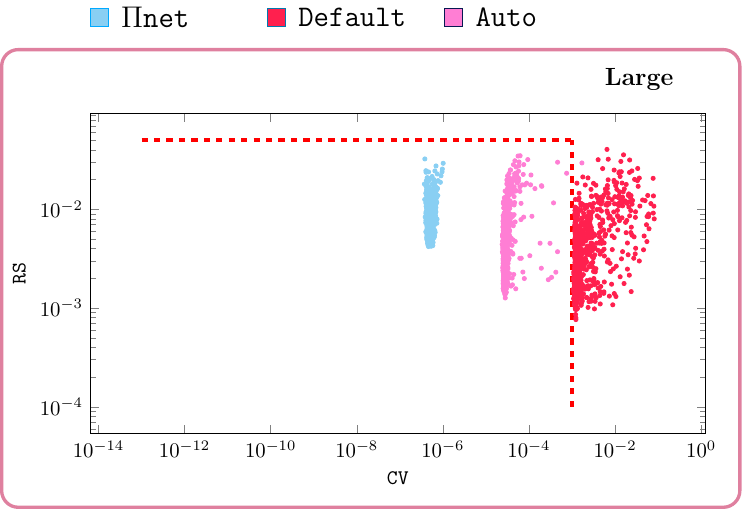}
    \end{minipage}
    \caption{Visualization of the ablation results in \cref{sec:the-sharp-bits:ablation}. 
    Scatter plots of \texttt{RS} and \texttt{CV} for the methods on the large non-convex problems
    on the test set. 
    The red dashed lines show the thresholds that we require to consider a candidate solution optimal. 
    }
    \label{fig:autotune:performances}
\end{figure}

\begin{figure}
    \centering
    \includegraphics[width=\linewidth]{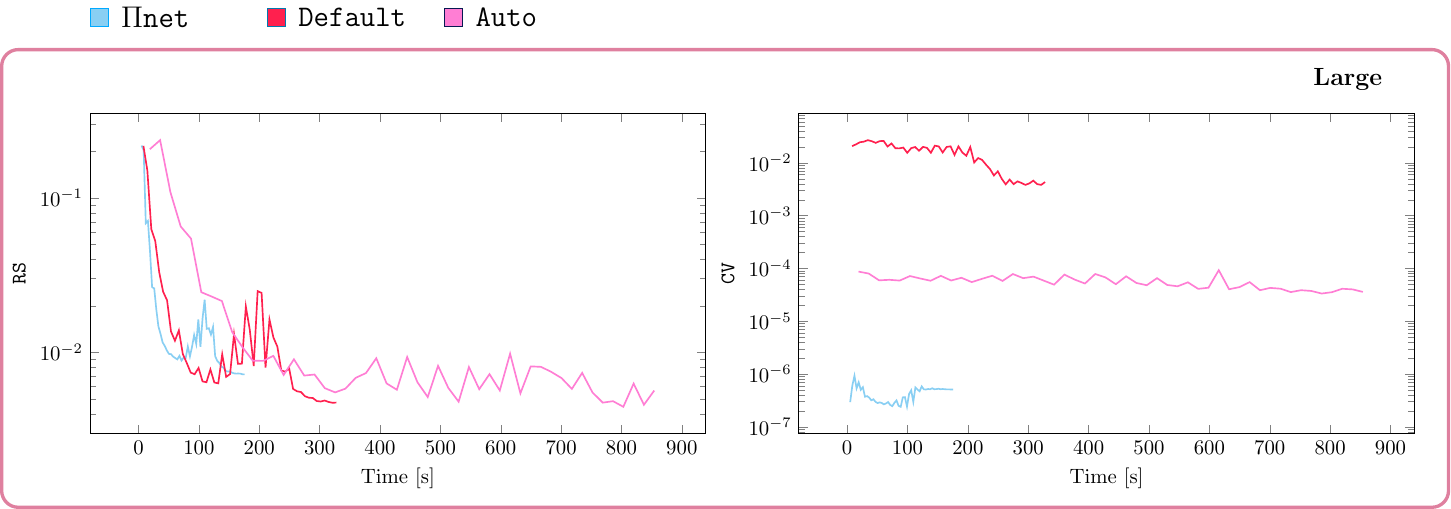}
    \caption{Comparison of the training times in terms of \texttt{RS} and \texttt{CV} for the different ablation configuration on the large non-convex problem setting. 
    For simplicity, we report the learning curves only for a single seed.
    }
    \label{fig:autotune:learning-curves}
\end{figure}

We report in \cref{fig:autotune:performances} the \texttt{RS} and \texttt{CV} by all 3 configurations,
training curves are shown in \cref{fig:autotune:learning-curves} for 50 training epochs and for a single seed,
and inference times are given in \cref{tab:autotune:times}.
From \cref{fig:autotune:performances} we can readily deduce the substantial improvement in terms
of \texttt{CV} that both auto-tuning and equilibration offers.
Additionally, \cref{fig:autotune:learning-curves} and \cref{tab:autotune:times} highlight the clear benefits
of our preprocessing steps in terms of both training time and inference time.
These benefits are due primarily to the fact that a well-tuned and equilibrated projection layer
requires significantly fewer iterations to achieve a satisfactory \texttt{CV}, specifically
$50$ iterations instead of $100$ and $350$ for default and only auto-tuned, respectively.

\begin{table}
    \centering
    \vspace{.5cm}
    
\begin{adjustbox}{width=\textwidth,center}
    \begin{tabular}{l|ccccc|ccccc}
\multirow{2}{*}{\textbf{Method}} & 
\multicolumn{5}{c|}{\textbf{Single inference [s]}} & 
\multicolumn{5}{c}{\textbf{Batch inference [s]}} \\
\cline{2-11}
 & median & LQ & UQ & $\min$ & $\max$ & median & LQ & UQ & $\min$ & $\max$ \\
\hline
\multicolumn{11}{l}{\textbf{Non-convex large}} \\\hline
\texttt{Default}    & 
0.0080&0.0080&0.0081&0.0075&0.0090
&
0.5491&0.5490&0.5493&0.5487&0.5495
\\
\texttt{Auto} &
0.0141&0.0139&0.0143&0.0135&0.0150
&
1.8900&1.8899&1.8901&1.8896&1.8903
\\
\rowcolor{blue!10}\ouralg   &  
0.0063 & 0.0063 & 0.0064 & 0.0061 & 0.0092
&
0.2804 & 0.2799 & 0.2807 & 0.2794 & 0.2912
\\
\end{tabular}
\end{adjustbox}
\vspace{.125cm}
    \caption{Inference time comparison for single-instance and batch-instance (1024 problems) settings across different ablation
    configurations, evaluated on the large non-convex problem. 
    The table reports median runtime along with statistical descriptors: 
    lower quartile (LQ, 25th percentile), upper quartile (UQ, 75th percentile), 
    $\min$ and $\max$ of the runtime.
    }
    \label{tab:autotune:times}

\end{table}

\begin{figure}
    \centering
    \begin{minipage}{.7\linewidth}
        \includegraphics[width=\linewidth]{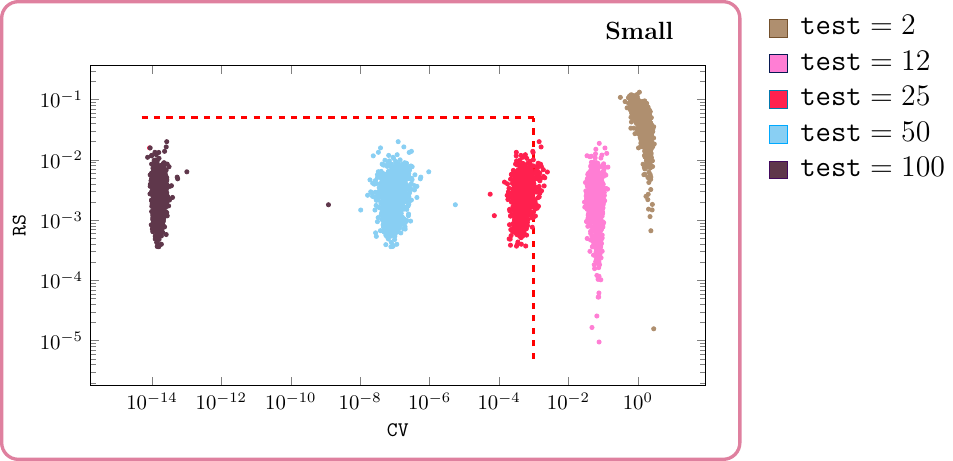}
    \end{minipage}
    \caption{\reviewerTwo{Scatter plot of \texttt{RS} and \texttt{CV} on the small non-convex test problems for a \ouralg network trained with 50 iterations and evaluated with different numbers of iterations at test time. The red dashed lines indicate the thresholds used to consider a candidate solution optimal.}}
    \label{fig:ablations_n_iter_test_train50}
\end{figure}

\reviewerTwo{In our experiments, we have extensively tested the effects of using a different number of forward iterations during testing (\texttt{n\_iter\_test}) compared to the training (\texttt{n\_iter\_fwd}), with both more and fewer iterations at test time. Using more iterations works well and stably, and is meaningful to achieve reduced constraint violation. Using less iterations needs more care. Slightly less iterations is often possible and can improve inference time. Significantly less may cause issues since the iterates might not be close to a fixed-point anymore. Of course, ``slightly'' and ``significantly'' here depend on the problem at hand. For this, in \cref{fig:ablations_n_iter_test_train50} we report the \texttt{RS} and \texttt{CV} obtained by training \ouralg with $\texttt{n\_iter\_fwd} = 50$, and then deploying with different number of iterations during testing. Perhaps interestingly, note that the \texttt{RS} is unaffected by the number of iterations during testing, whereas the \texttt{CV} gradually decreases. This is true as long as we have enough iterations: as the number of iterations during test becomes too small (e.g., 2) also the \texttt{RS} is affected: The output is not close to the projection anymore.}

\begin{figure}
    \centering
    \begin{minipage}{.7\linewidth}
        \begin{tcolorbox}[
      colback=white,
      colframe=blue!20,
      arc=1mm,
      boxsep=0pt,
      top=2pt,
      left=5pt,
      right=5pt,
      bottom=2pt,
      toptitle=3pt,
      bottomtitle=2pt,
      coltitle=black,
      title=\centering{$\texttt{n\_iter\_fwd} = 50$}
    ]
    \includegraphics[width=\linewidth]{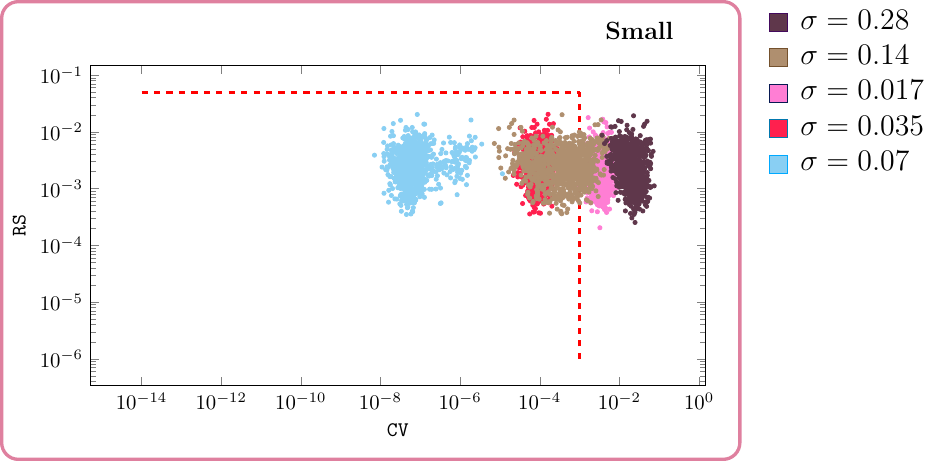}
    \end{tcolorbox}
    \end{minipage}
    
    \vspace{.5cm}
    
    \begin{minipage}{.7\linewidth}
            \begin{tcolorbox}[
      colback=white,
      colframe=orange!20,
      arc=1mm,
      boxsep=0pt,
      top=2pt,
      left=5pt,
      right=5pt,
      bottom=2pt,
      toptitle=3pt,
      bottomtitle=2pt,
      coltitle=black,
      title=\centering{$\texttt{n\_iter\_fwd} = 100$}
    ]
    \includegraphics[width=\linewidth]{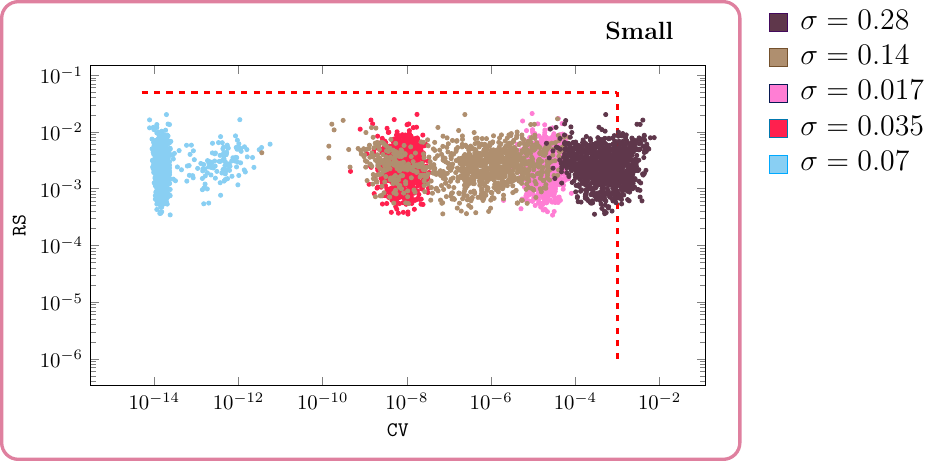}
    \end{tcolorbox}
    \end{minipage}
    \caption{\reviewerFour{Scatter plot of \texttt{RS} and \texttt{CV} on the small non-convex test problems for a \ouralg network trained with different values of $\sigma$ and (top) $50$, (bottom) $100$ forward iterations. The red dashed lines indicate the thresholds used to consider a candidate solution optimal.}}
    \label{fig:sigma_ablations}
\end{figure}

\begin{figure}
    \centering
    \begin{minipage}{.7\linewidth}
        \includegraphics[width=\linewidth]{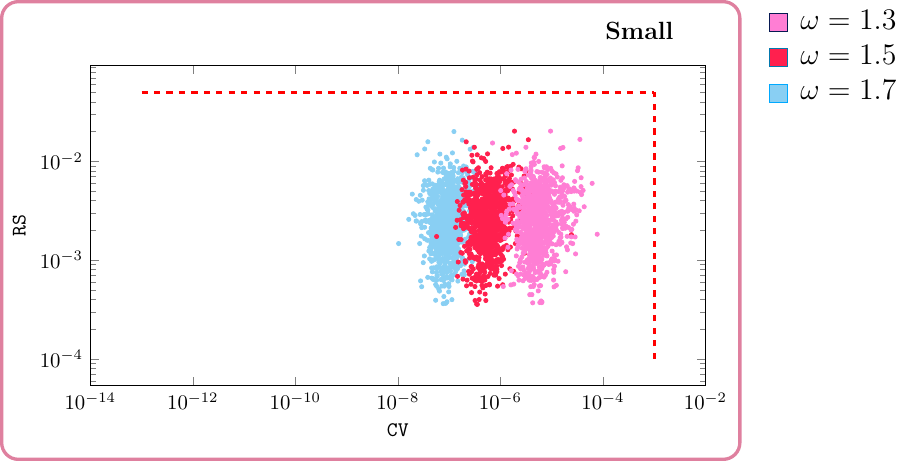}
    \end{minipage}
    \caption{\reviewerFour{Scatter plot of \texttt{RS} and \texttt{CV} on the small non-convex test problems for a \ouralg network trained with different values of $\omega$. The red dashed lines indicate the thresholds used to consider a candidate solution optimal.}}
    \label{fig:ablation_omega}
\end{figure}

\reviewerFour{Finally, in \cref{fig:sigma_ablations,fig:ablation_omega} we report an ablation study on the parameters $\sigma$ and $\omega$. These results, together with the ones on the number of forward and backward iterations (cf. \cref{fig:ablations_n_iter_test_train50} and \cref{appendix:convergence_of_dra:rates,appendix:theory}), substantiate the claim of little sensitivity of \ouralg to hyperparameter tuning. In particular, \cref{fig:sigma_ablations,fig:ablation_omega} show that different values of the parameters $\sigma$ and $\omega$ yield qualitatively similar behaviors. For $\omega$, we see clearly that different values effectively only change the convergence rates: With more iterations, all values of $\omega$ achieve sufficiently low \texttt{CV}.
}

\subsection{Why enforcing constraints during training?}
\label{sec:constraints-training}
\ouralg enforces the constraints during training, which warrants a discussion: What changes if one trains an unconstrained network and enforces the constraints only during inference? We answer this question with illustrative examples.

\paragraph{Some optimization problems may cause divergence of the network if trained unconstrained.} The first example shows that, without constraints, the training process may even diverge, as the optimization problems we are interested in may not even be meaningful in the absence of constraints.
Consider the parametric (in $\mathrm{x}$) optimization problem:

\begin{minipage}{.45\linewidth}
    \[
\underset{ y \in \mathbb{R} }{\mathrm{minimize}} 
\quad \mathrm{x} y \quad
\mathrm{subject~to} \quad 0 \leq y \leq 1.
\]
The optimal solution is 
\[
y^\star(\mathrm{x}) = \begin{cases}
    0 \quad & \text{if } \mathrm{x} < 0\\
    1 \quad & \text{if } \mathrm{x} > 0\\
    \text{any }y \in [0,1]\quad&\text{otherwise}.
\end{cases}
\]
\end{minipage}
\begin{minipage}{.55\linewidth}
\centering
\includegraphics[width=\linewidth]{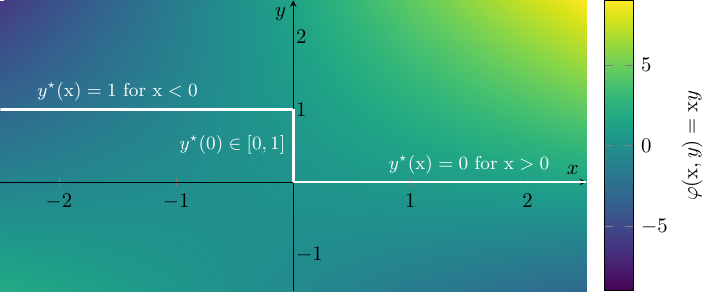}
\end{minipage}

However, if we would train an unconstrained network to predict $\hat{y}(\mathrm{x})$, its values would result in $\hat{y}(\mathrm{x}) \to +\infty$ for $\mathrm{x} > 0$ and $\hat{y}(\mathrm{x}) \to -\infty$ for $\mathrm{x}< 0$. 

It is easy to see how one may build higher-dimensional examples along the same lines of the presented one. Importantly, this example shows that, without constraints, the problem may be altered significantly to the point that it is not even possible to train an unconstrained network.

\paragraph{The projection of the unconstrained optimizer is often suboptimal.}
Our second example shows that applying the projection layer to the solution of the unconstrained problem does not yield, in general, the solution to the constrained problem.

\begin{minipage}{.6\linewidth}
As a consequence, one can expect a strong decrease in performance when training the network unconstrained (i.e., to solve the unconstrained optimization problem) and only at inference adding a projection layer (i.e., projecting the solution of the unconstrained problem onto the feasible set).

Consider the parametric (in $\mathrm{x}$) optimization problem depicted on the right.
Here, one may train an unconstrained network to optimize the objective function, finding the optimizer $\hat{y}(\mathrm{x}) = \mathrm{x} + 2$. However, projecting this onto the constraints $y \leq \mathrm{x}$ would result in $y = \mathrm{x}$, which is suboptimal compared to the constrained optimizer $y^\star(x) = \mathrm{x} - 2$. Importantly, this simple example shows an instance in which training without constraints may downgrade the performances of the network.
\end{minipage}\hfill
\begin{minipage}{.35\linewidth}
    \centering
    \vspace{-.5cm}
    \includegraphics[width=\linewidth]{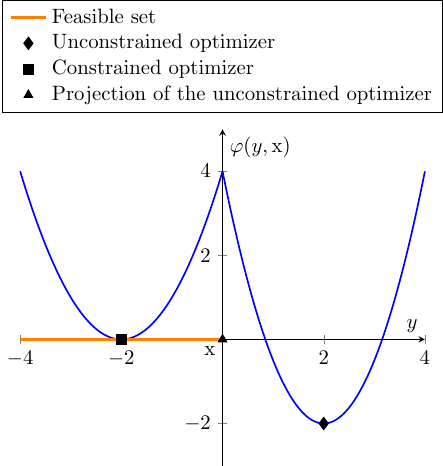}
\end{minipage}

\paragraph{An empirical case-study.}
In the third example, we train an unconstrained network on the small non-convex benchmark (using soft penalty terms) 
and add the projection layer only during inference on the test set (we refer to this as \ouralg-\texttt{inf}). 
The resulting average \texttt{RS} on the test set for \ouralg-\texttt{inf} is 0.02178, whereas for \ouralg it is 0.00216
using roughly the same computational budget (8 seconds of wall-clock time).
The constraint violation is the same for both methods, since they use our projection layer during inference.
Training with \ouralg (i.e., enforcing constraints during training) achieves one order of magnitude better \texttt{RS}.
In fact, if we increase the computational budget, \ouralg continues to reduce the \texttt{RS} down to 0.0007 for 50 seconds of training.
On the contrary, \ouralg-\texttt{inf} cannot reduce the \texttt{RS} further even with more training (or more hyperparameter tuning).
Therefore, we observe in practice what our previous examples outlined:
enforcing the constraints during training allows the network to anticipate the projection 
and further improve its predictions.

\paragraph{Constraints as an inductive bias.} There is a subtle difference between constraining the architecture of a neural network (e.g., by ensuring that the neural network is a convex function for any values of the weights, as done by \citet{amos2017input}) and constraining the output to satisfy some properties without altering the ``backbone'' network architecture. In \ouralg, one can use the best existing architecture unaltered, but training it with the inductive bias that the output will be modified so to satisfy the constraints. For this, one may argue that enforcing the constraints during training improves the performance of the network at inference time, rather than downgrading it. We thus believe that an approach like \ouralg may, in fact, change the common perspective on the difficulty of training hard-constrained networks. 

\begin{remark}
    It is perhaps worth clarifying the distinction between our work and the work of \citet{amos2017input}:
    \begin{enumerate}[leftmargin=*]
        \item Input convex neural networks learn a {scalar-valued} function, which is guaranteed to be a convex function of its input for any appropriate choices of the network weights (see \citet[Propositions 1, 2]{amos2017input}). Then, during inference, a minimum of this function is computed. Recall that the set of minimizers of a convex function is a convex set, and in this sense \ouralg and input convex neural networks are similar.
    \item On the contrary, \ouralg learns a {vector-valued} mapping which is guaranteed to lie on a convex set chosen a-priori for any choice of the weights.
    \end{enumerate}
\end{remark}

\subsection{\texorpdfstring{\ouralg}{Pinet} as an implicit layer}
\label{sec:pinet-implicit-layer}
The techniques used to implement \ouralg (operator-splitting and backpropagation via the implicit function theorem) are fundamental and widely adopted in the literature. In this sense, \ouralg is a special case of implicit layers \citep{agrawal2019differentiable,butler2023efficient}. However, our key ideas are related to the type of problem we are solving (a projection) and how the structure of this problem can be exploited to derive an efficient formulation of the optimization algorithm. The improvement over the state of the art (as exemplified by our results in \cref{sec:experiments}) is achieved by focusing on a problem setting that is sufficiently general yet rich in structure, and by adopting the right optimization techniques (e.g., which split to perform to deploy the Douglas-Rachford algorithm). The key ideas, in this sense, are:
\begin{enumerate}[leftmargin=*]
    \item Using a single additional layer instead of multiple ones.
    \item Using a projection layer.
    \item Adapting the Douglas-Rachford algorithm to best exploit the resulting problem structure.
\end{enumerate}

\clearpage
\section{Derivation details}
\label{appendix:extra-calculations}
In this section, we report the derivations omitted in the text.

\subsection{Derivation of \texorpdfstring{\cref{alg:os-project}}{Algorithm 1}.}
\label{appendix:extra-calculations:inequality_projection}
To derive \cref{alg:os-project} from \eqref{eq:dra_for_projection}, we will explicitly write the
proximal operators in \eqref{eq:dra_for_projection}.
For the $z$-update, we recall that $g = \mathcal{I}_{\mathcal{A}} $ and $\sigma > 0$,
from which it immediately follows that $\mathrm{prox}_{\sigma g} = \Pi_{\mathcal{A}}$.

For the $t$-update, we recall that:
\begin{equation*}
    h\left(\begin{bmatrix}
    y\\y_{\text{aux}}
\end{bmatrix}\right) = \norm{y - y_{\text{raw}}}^2
    + \mathcal{I}_{\mathcal{K}}\left(\begin{bmatrix}
    y\\y_{\text{aux}}
\end{bmatrix}\right)
\end{equation*}
and using the definition of the proximal operator yields:
\begin{align*}
\mathrm{prox}_{\sigma h}\left(
    2z_{k + 1} - s_{k}
\right)
&=
\underset{y, y_{\text{aux}}}{\argmin}
\;
\norm{y - y_{\text{raw}}}^2
+
\mathcal{I}_{\mathcal{K}}\left(\begin{bmatrix}
    y\\y_{\text{aux}}
\end{bmatrix}\right)
+
\frac{1}{2\sigma}
\norm{\begin{bmatrix}
    y\\y_{\text{aux}}
\end{bmatrix}
-
2z_{k + 1} + s_k
}^2
\\
&=
\begin{bmatrix}
  \underset{y \in \mathcal{K}_1}{\argmin}
\;
\norm{y - y_{\text{raw}}}^2
+
\frac{1}{2\sigma}
\norm{y
-
2z_{{k + 1},1} + s_{k,1}
}^2  
\\
\underset{y_{\text{aux}} \in \mathcal{K}_2}{\argmin}
\;
\frac{1}{2\sigma}
\norm{
    y_{\text{aux}}
-
2z_{{k + 1},2} + s_{k,2}
}^2
\end{bmatrix}
\\
&=
\begin{bmatrix}
\Pi_{\mathcal{K}_1}(\frac{2z_{{k + 1},1} - s + 2\sigma y_{\text{raw}}}{1 + 2\sigma}) 
\\
\Pi_{\mathcal{K}_2}(2z_{{k + 1},2} - s_{k,2})
\end{bmatrix}.
\end{align*}

\subsection{Convergence proof of \texorpdfstring{\eqref{eq:dra_for_projection}}{the fixed point iteration}.}
\label{appendix:extra-calculations:convergence_of_dra}
In this subsection, we show that \eqref{eq:dra_for_projection} converges to a solution of \eqref{eq:projection_in_lifted_form}.
To do so, we first introduce the following mild assumption on the feasibility of the problem.
\begin{assumption} \label{assumption:strict_feasibility}
    $\mathcal{A} \cap \mathrm{ri} \, \mathcal{K} \neq \emptyset $, where $\mathrm{ri}$ is the relative interior.
\end{assumption}
This assumption corresponds to strict feasibility of \eqref{eq:projection_in_lifted_form}.
In fact, if $ \mathcal{K} $ is a polyhedron, then we can relax \cref{assumption:strict_feasibility}
to $\mathcal{A} \cap \mathcal{K} \neq \emptyset $.

To obtain our convergence result, we simply note that iteration \eqref{eq:dra_for_projection} is the 
Douglas-Rachford algorithm applied to problem \eqref{eq:projection_in_lifted_form} and,
under \cref{assumption:strict_feasibility}, we invoke the Corollaries 27.6 and 28.3 in \citep{bauschke_convex_2017}
which yield the desired result.

\subsection{Convergence rates of \texorpdfstring{\eqref{eq:dra_for_projection}}{the fixed point iteration}.}
\label{appendix:convergence_of_dra:rates}
\begin{figure}
    \centering
    \includegraphics[width=.75\linewidth]{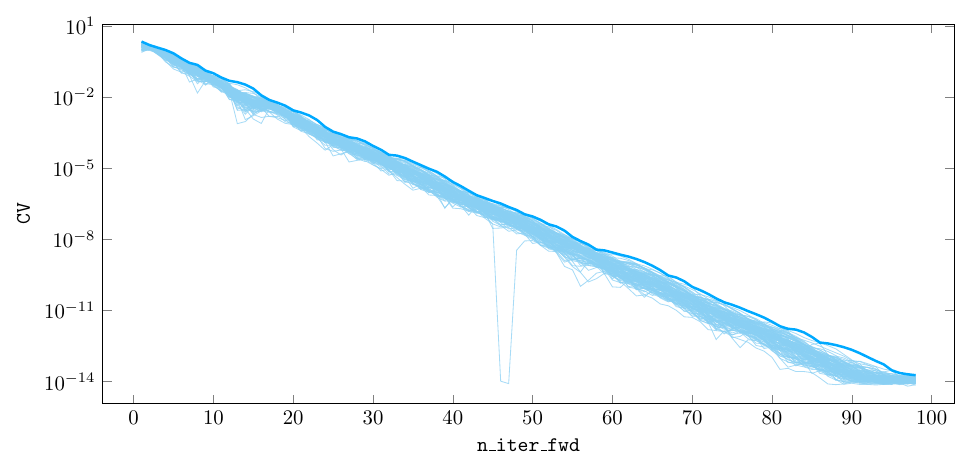}
    \caption{Constraint violation for $100$ problem instances (light blue) in the test set of the small non-convex benchmark as the number of iterations in the forward pass \texttt{n\_iter\_fwd} (i.e., \texttt{n\_iter\_train} increases, as well as the maximum \texttt{CV} among all instances (dark blue).}
    \label{fig:convergence-rates}
\end{figure}

The convergence rate of the Douglas-Rachford algorithm is a well-studied topic in the literature, which we report here for completeness. For the case of a convex \gls*{qp}, it has been shown that the Douglas-Rachford algorithm attains a linear convergence rate \cite[Subsection 3.2]{pena2021linear}, namely an $\varepsilon$-optimal solution is computed in $\mathcal{O}(\log(1/\varepsilon))$ iterations. This result applies to our algorithm in the case of polyhedral constraints, and indeed we have observed these rates in our numerical experiments (see also \cref{fig:convergence-rates}). Other technical conditions for linear convergence are given in \cite[Assumption A]{hong2017linear}; their applicability to our setting depends on the specific constraint set $\mathcal{C}$. In the general setting, the Douglas-Rachford algorithm typically attains a $\mathcal{O}(1/k)$ convergence rate (where $k$ is the number of iterations), hence requiring $\mathcal{O}(1/\varepsilon)$ iterations \citep{davis2017faster}.

We visualize these convergence rates in the plot in \cref{fig:convergence-rates}, showing how for increasing iterations we get a rapidly decreasing \texttt{CV}. Specifically, for one of the benchmarks, we provide the \texttt{CV} for 100 instances in the test set for an increasing number of iterations. 
In particular, we observe that within very few iterations the constraints are satisfied up to high accuracy (e.g., $10^{-5}$ as in \citep{Stellato_2020}).

\subsection{Derivation of the backward pass}
\label{appendix:extra-calculations:derivation_of_backward}
Applying the implicit function theorem to $s_{\infty}(y_{\text{raw}}) = \Phi(s_{\infty}(y_{\text{raw}}), y_{\text{raw}})$ yields the linear equation:
\begin{equation} \label{eq:implicit_function_theorem}
    \left(I - \frac{\partial \Phi(s, y_{\text{raw}})}{\partial s}\bigg|_{s = s_\infty(y_{\text{raw}})}\right)
    \frac{\partial s_\infty(y_\text{raw})}{\partial y_\text{raw}} \bigg|_{y_\text{raw} = f(\mathrm{x};\theta)}
    =
    \frac{\partial \Phi(s, y_\text{raw})}{\partial y_\text{raw}}\bigg|_{y_\text{raw} = f(\mathrm{x};\theta)},
\end{equation}
whose unknown is the Jacobian matrix $\frac{\partial s_\infty(y_\text{raw})}{\partial y_\text{raw}} \big|_{y_\text{raw} = f(\mathrm{x};\theta)}$.

Now, to compute the \gls{vjp}
\begin{equation*}
    v^{\top} \frac{\partial s_\infty(y_\text{raw})}{\partial y_\text{raw}} \big|_{y_\text{raw} = f(\mathrm{x};\theta)}
\end{equation*}
where $v \in \reals^{n+d}$, we assume that the linear system \eqref{eq:linear-system} admits a solution, namely $\xi(y_\text{raw}, v)$.
Next, we pre-multiply \eqref{eq:implicit_function_theorem} by $\xi(y_\text{raw}, v)^{\top}$:
\begin{align*}
    \xi(y_\text{raw}, v)^{\top}
    \left(I - \frac{\partial \Phi(s, y_{\text{raw}})}{\partial s}\bigg|_{s = s_\infty(y_{\text{raw}})}\right)
    \frac{\partial s_\infty(y_\text{raw})}{\partial y_\text{raw}} \bigg|_{y_\text{raw} = f(\mathrm{x};\theta)}
    & =
    \xi(y_\text{raw}, v)^{\top}
    \frac{\partial \Phi(s, y_\text{raw})}{\partial y_\text{raw}}\bigg|_{y_\text{raw} = f(\mathrm{x};\theta)} 
    \\
    \implies v^{\top}
    \frac{\partial s_\infty(y_\text{raw})}{\partial y_\text{raw}} \bigg|_{y_\text{raw} = f(\mathrm{x};\theta)}
    & =
    \xi(y_\text{raw}, v)^{\top}
    \frac{\partial \Phi(s, y_\text{raw})}{\partial y_\text{raw}}\bigg|_{y_\text{raw} = f(\mathrm{x};\theta)} 
\end{align*}
where the implication follows by the definition of $\xi(y_\text{raw}, v)$.
This gives rise to our backward pass outlined in \eqref{eq:vjp} and \eqref{eq:linear-system}.

\subsection{Computational complexity of \texorpdfstring{\ouralg}{Pinet}.}
\label{sec:computational-complexity}
Here we derive the computational complexity of \ouralg for both the forward and backward pass.
\paragraph{Forward.} The complexity of the projection operations in \cref{alg:os-project} are determined by\footnote{The complexity also depends on the batch size, but the operations can be fully parallelized over a batch and, thus, we focus on the per-element complexity.} the sets $\mathcal{A}(\mathrm{x})$, $\mathcal{K}_1 (\mathrm{x})$ and $\mathcal{K}_2 (\mathrm{x})$, by the cost of instantiating them given the context $\mathrm{x}$, which may depend on the dimensionality $p$ of $\mathrm{x}$, and by the dimensions of the input $n$, $d$ and $n - d$. We denote the complexity of these operations as $g_{\mathrm{x}}(n, p), g_{\mathcal{A}}(n), g_{\mathcal{K}_1}(d)$ and $g_{\mathcal{K}_2}(n-d)$. The overall complexity is:
\[
\mathcal{O}\left(
g_{\mathrm{x}}(n, p) + K(g_{\mathcal{A}}(n) + g_{\mathcal{K}_1}(d) + g_{\mathcal{K}_2}(n-d) + n)
\right),
\]
where $K = \mathrm{n\_iter\_fwd}$ is the number of forward iterations, and the term $n$ is due to the sum\footnote{On a GPU, this entry-wise vector sum can be parallelized, but here we focus on the number of operations.} in the governing update \eqref{eq:dra_governing_update}.

Next, we analyze the complexity of $g_{\mathrm{x}}(n, p), g_{\mathcal{A}}(n), g_{\mathcal{K}_1}(d)$ and $g_{\mathcal{K}_2}(n-d)$:
\begin{itemize}[leftmargin=*]
    \item $g_{\mathrm{x}}(n, p)$ is problem specific. In the examples presented in this work, the complexity is linear in $p$ and amounts to stacking the vectors describing the problem instance (e.g., the initial and final position of the vehicles in \cref{sec:multi-robot}) into the equality constraints. In general, one may expect $g_{\mathrm{x}}(n, p)$ to be dominated by the other contributions. If the matrix $A$ is context dependent, an additional cost is the calculation of the pseudo-inverse, which we use to compute the projection onto the hyperplane: $g_{\mathrm{x}}(n, p) = n^3$; see also the next item. When the matrix $A$ is independent of the context, the pseudo-inverse can be computed only once at the instantiation of the method, and its computational complexity does not affect the forward pass.
    \item $g_{\mathcal{A}}(n)$ is the cost of the projection on the hyperplane identified by a matrix $A \in \reals^{(d + d')\times n}$ and a vector $b \in \reals^{d + d'}$. The parameter $d'$ is problem-dependent, but usually proportional to the number of inequality constraints; see \cref{appendix:extra-sets}. For this reason, we carry out the complexity analysis for $A \in \reals^{n\times n}$. For the projection, we need the pseudo-inverse of $A$, which we compute once for all iterations. Thus, we account for this complexity in $g_{\mathrm{x}}(n, p)$. Then, the complexity of computing the projection is dominated by matrix-vector multiplications: $g_{\mathcal{A}}(n) = n^2$.
    \item $g_{\mathcal{K}_1}(d)$ and $g_{\mathcal{K}_2}(n-d)$ are the cost of the projection onto a box after a linear combination of the input. Thus, $g_{\mathcal{K}_1}(d) + g_{\mathcal{K}_2}(n-d) = n$.
\end{itemize}
Overall, the complexity of the forward pass is $\mathcal{O}(n^2)$ for context-independent $A$ and $\mathcal{O}(n^3)$ for context-dependent $A$. On a GPU and for context-independent $A$ matrices, the effective compute time is possibly linear in $n$.
\paragraph{Backward.}
For the backward pass, we need to:
\begin{itemize}[leftmargin=*]
    \item Compute the $v$ for the linear system in \eqref{eq:linear-system}, for which we need only the first \gls*{vjp} in \eqref{eq:projection_vjp_expanded}. The complexity is $\mathcal{O}(d(n + d)) = \mathcal{O}(n^2)$.
    \item Solve the linear system in \eqref{eq:linear-system} using \texttt{bicgstab} \citep{bicgstab}. The complexity of this operation is $\mathcal{O}(K'(n + d)d) = \mathcal{O}(K'n^2)$, where $K' = \texttt{n\_iter\_backward}$ is the number of iterations used for \texttt{bicgstab}.
    \item We compute the \gls*{vjp} in \eqref{eq:vjp}. By direct inspection of the computational graph of \eqref{eq:equality_constraint}-\eqref{eq:dra_governing_update}, complexity of this step is $\mathcal{O}(d(d + n)) = \mathcal{O}(n^2)$.
\end{itemize}
Overall, the backpropagation through our projection layer has complexity $\mathcal{O}(n^2)$, with the constant dominated by the number of iterations used for \texttt{bicgstab}.
\clearpage
\section{More examples of constraint sets}
\label{appendix:extra-sets}
In this section, we describe how several classes of constraints that can be described within our framework admit an efficient projection. 
    We stress that the decomposition $\mathcal{C} = \Pi_d(\mathcal{A} \cap \mathcal{K})$ 
    is not an assumption.
    One can always decompose a convex set in this way, e.g., by considering the trivial decomposition 
    $\mathcal{A} = \mathcal{C}$ and $\mathcal{K} = \mathbb{R}^d$.
    Instead, determining $\mathcal{A}$ and $\mathcal{K}$ is a design choice which we leverage to make the
    projections $\Pi_\mathcal{A}$ and $\Pi_\mathcal{K}$ computationally efficient.
    We note two important points regarding this design choice:
    \begin{itemize}[leftmargin=*]
    \item The only assumption is that $\Pi_\mathcal{A}$ and $\Pi_\mathcal{K}$ and their \gls*{vjp} are computable.
    Being computationally efficient is an added benefit of our decomposition, but is not necessary.
    \item Below, we show that many practically-relevant constraints $\mathcal{C}$ can
    be decomposed in a computationally-efficient manner: polyhedra, second-order cones, sparsity constraints, simplices,
    and the intersections and Cartesian products all admit efficient decompositions.
    In fact, this list is not exhaustive; see, e.g., \citep{condat2016fast,boyd2004convex}.
    \end{itemize}
We believe the implementation and adoption of these constraints in practical applications represents an important direction for future work.

\paragraph{Polytopic sets} are often employed to enforce constraints in robotics \citep{augugliaro2012generation}, numerical solutions to \gls*{pde} \citep{raissi2019physics}, and non-convex relaxations for trajectory planning \citep{malyuta2022convex}, among others. They are expressed as
\[
\{ y \in \reals^d \st E y= q, l \leq Cy \leq u \},
\]
for some $E, q, l, C, u$ of appropriate dimensions. 
We introduce the auxiliary variable $y_{\text{aux}} = Cy \in \reals^{n_\text{ineq}}$
with dimension $n-d = n_\text{ineq}$.
Then, we respectively define $\mathcal{A}, \, \mathcal{K} \subseteq \reals^{n}$ as the following affine subspace and box
\begin{align*}
\mathcal{A} & = \bigg\{
\begin{bmatrix}
    y\\
    y_{\text{aux}}
\end{bmatrix}
\, \bigg| \,
\underbrace{\begin{bmatrix}
    E & 0\\C & -I
\end{bmatrix}}
_{=A} 
\begin{bmatrix}
    y \\ y_\text{aux}
\end{bmatrix} 
=
\underbrace{\begin{bmatrix}
    q \\0
\end{bmatrix}}
_{=b}
\bigg\},
\quad
\mathcal{K} = \bigg\{ \begin{bmatrix}
    y\\
    y_{\text{aux}}
\end{bmatrix} \, \bigg| \,\,
y \in \reals^d, ~ l \leq y_\text{aux} \leq u
\bigg\}.
\end{align*}
Importantly, $\mathcal{C} = \Pi_d(\mathcal{A} \cap \mathcal{K})$ and both $\Pi_{\mathcal{A}}$ and $\Pi_{\mathcal{K}}$ can be evaluated in closed form
\cite[Chapter 29]{bauschke_convex_2017}.

\paragraph{Second-order cones} are employed in portfolio optimization \citep{brodie2009sparse}, robust control \citep{chen2018approximating}, and support vector machines \citep{maldonado2014imbalanced}, among others. They involve constraints of the form:
\[
\{y \in \reals^d \st \norm{C y + c}_2 \leq f^\top y + e\}.
\]
Introducing the auxiliary variables $y_{\text{aux},1} = C y + c \in \reals^{n_\text{c}}$ and $ y_{\text{aux},2} = f^\top y + e \in \reals$,
of dimension $n-d = n_\text{c} + 1$, we obtain the representation:
\[
\mathcal{A} = \Bigg\{
\begin{bmatrix}
    y\\y_{\text{aux},1}\\y_{\text{aux},2}
\end{bmatrix}
\, \Bigg| \,
\underbrace{
\begin{bmatrix}
    C & -I & 0\\
    f^\top & 0 & -1
\end{bmatrix}
}_{=A}
\begin{bmatrix}
    y\\y_{\text{aux},1}\\y_{\text{aux},2}
\end{bmatrix}
=
\underbrace{
\begin{bmatrix}
    -c\\-e
\end{bmatrix}
}_{=b}
\Bigg\},
\quad
\mathcal{K} = \Bigg\{
\begin{bmatrix}
    y\\y_{\text{aux},1}\\y_{\text{aux},2}
\end{bmatrix}
\, \Bigg| \,
\norm{y_{\text{aux},1}}_2 \leq y_{\text{aux},2}
\Bigg\}
\]
for quantities of the appropriate dimensions. 
Again, $\Pi_{\mathcal{A}}$ and $\Pi_{\mathcal{K}}$ admit a closed form solution \citep{boyd2004convex}.%

\paragraph{Sparsity constraints} encourage solutions with fewer active variables, such as the number of open positions in portfolio optimization \citep{brodie2009sparse}, or the ``mass splitting'' in optimal transport \citep{dantzig2002linear}. Explicitly, they read
\[
\{
y \in \reals^d \st \norm{Cy + c}_1 \leq f^\top y + e
\}
\]

Analogously to second-order cones constraints, employing a slack variable they can be reduced to a projection onto the $l_1$ ball, which can be done efficiently \citep{condat2016fast}. 

\paragraph{Cartesian products of simplices} are the standard constraints in optimal transport \citep{dantzig2002linear} formulations, since they encode the set of admissible coupling measures.
While these constraints can also be expressed as generic polytopic sets, an alternative representation
using our proposed form
can potentially yield more efficient solutions. 
For instance, for some $v_1, \ldots, v_n, w_1, \ldots, w_m \geq 0$,
consider the following constraint set
\begin{align*}
\mathcal{C} = \biggl\{
y \in \reals^{d_1\times d_2} &\st y_{ij} \geq 0, \sum_{i = 1}^n y_{ij} = w_j, \sum_{j = 1}^m y_{ij} = v_i 
\biggr\}
\\&=
\left\{
y \in \reals^{d_1\times d_2} \st \sum_{i = 1}^n y_{ij} = w_j 
\right\} \cap \left\{
y \in \reals^{d_1\times d_2} \st y_{ij} \geq 0, \sum_{j = 1}^m y_{ij} = v_i 
\right\}.
\\&= 
\mathcal{A} \cap \mathcal{B}
\end{align*}
Notice, that $\mathcal{C} = \Pi_d(\mathcal{A} \cap \mathcal{K})$ as required by our constraint decomposition,
and $\Pi_d(\cdot)$ is redundant since $\mathcal{A}$ and $\mathcal{K}$ are of the same dimension as $\mathcal{C}$.
The projections $\Pi_\mathcal{A}$ and $\Pi_\mathcal{K}$ can be evaluated efficiently since
$\mathcal{A}$ is a hyperplane and $\mathcal{K}$ is a cartesian product of simplices \citep{condat2016fast}. 

\paragraph{The intersection of previous sets} can readily be expressed in our representation.
For instance, the intersection of a polytope and second-order cone
can be expressed as the following set:
\begin{align*}
    \{ y \in \reals^d \,|\, E y = q, ~ l \leq C y \leq u, ~ \norm{F y + c}_2 \leq f^{\top} y + e \}.
\end{align*}
We introduce the auxiliary variables 
$ y_\text{aux, 1} = C y \in \reals^{n_\textrm{ineq}}, ~ y_\text{aux, 2} = F y \in \reals^{n_c}, ~
y_\text{aux, 3} = f^{\top} y \in \reals
$
and obtain the representation:
\begin{align*}
    \mathcal{A} & = 
    \left\{
    \begin{bmatrix}
        y \\ 
        y_\text{aux, 1} \\
        y_\text{aux, 2} \\
        y_\text{aux, 3}
    \end{bmatrix} ~\left|~
    \begin{bmatrix}
        E           & 0     & 0     & 0 \\
        C           & -I    & 0     & 0 \\
        F           & 0     & -I    & 0 \\
        f^{\top}    & 0     & 0     & -1 
    \end{bmatrix}
    \begin{bmatrix}
        y \\ 
        y_\text{aux, 1} \\
        y_\text{aux, 2} \\
        y_\text{aux, 3}
    \end{bmatrix} 
    =
    \begin{bmatrix}
        q \\ 0 \\ -c \\ -e
    \end{bmatrix}
    \right.
    \right\}, \\
    \mathcal{K} & = 
    \reals^d \times 
    \{ y_\text{aux, 1} \in \reals^{n_\textrm{ineq}} \,|\,  l \leq y_\text{aux, 1} \leq u \} \times
    \left\{ 
    \begin{bmatrix}
        y_\text{aux, 2} \\ y_\text{aux, 3}
    \end{bmatrix} \in \reals^{n_c + 1} \,|\,
    \norm{y_\text{aux, 2}} \leq y_\text{aux, 3}
    \right\}.
\end{align*}
As before, the projections $\Pi_\mathcal{A}$ and $\Pi_\mathcal{K} $ admit closed-form expressions.
\clearpage
\section{On the differentiability of the projection layer and the applicability of the implicit function theorem}
\label{appendix:theory}
\reviewerThree{
In this section, we discuss the theoretical aspects of the differentiability of the proposed projection layer, as well as the applicability of the implicit function theorem.
}

\subsection{Almost everywhere differentiability}
\reviewerThree{
To start, we note that since the layers of the backbone network will be at most almost everywhere differentiable (e.g., for \texttt{ReLU} activations), all one may be interested in is to have almost everywhere differentiability of the projection layer. In fact, our experiments show that in practice this is sufficient for markedly surpassing the state of the art. On Euclidean spaces (more generally, on Hilbert spaces) the projection operator is globally 1-Lipschitz (see, e.g., \citep[Proposition 4.16]{bauschke_convex_2017}, and recall that firmly non-expansive implies 1-Lipschitz). Then, with Rademacher’s theorem \citep[Theorem 1.4]{simon1984lectures} we conclude that the projection layer is almost everywhere differentiable (in the Lebesgue measure sense).
}

\subsection{Applicability of the implicit function theorem}
\reviewerThree{
Providing a full theoretical analysis is beyond the scope of the paper, and we believe it would distract the reader from the main message of the paper. Nonetheless, we (i) believe this to be a very interesting mathematical quest for future work, and (ii) that the outline presented here can give a sense to the reader of why the proposed projection layer is theoretically sound.}

\reviewerThree{
The implicit function theorem is applicable under local differentiability conditions. The way that related works handle non-smooth points is to either (i) show that if the linear system \eqref{eq:linear-system} is non-singular, then multiple solutions exist and they belong to the subdifferential (see, e.g., \citep[Appendix C.1]{amos2017optnet}) or (ii) they resort to heuristics, e.g., by solving a least-squares approximation rather than the linear system itself (see, e.g., \citep[Section 4.3]{agrawal2019differentiable}). We similarly run finitely many \texttt{bicgstab} iterations, which is very computationally attractive as it avoids run-time checks and, in our numerical experience, is quite stable; see \cref{appendix:empirical-properties-backward}.
}

\reviewerThree{Perhaps surprisingly, one can define an implicit function theorem even on non-smooth points by employing \emph{conservative gradients}, a recent generalization of the subdifferential \citep{bolte2021conservative}. This formulation is valid for non-smooth locally Lipschitz mappings, to which our projection layer belongs. We refer to the seminal works of \citet{bolte2021conservative,bolte2021nonsmooth} for the precise statement of the non-smooth variant of the theorem. In this framework, the implicit function theorem formula and the chain rule used in \cref{sec:backward-pass} hold under very mild conditions. Roughly speaking, this would correspond to our operator $\Phi(x, y)$ being semialgebraic and contractive with respect to $y$ \citep{bolte2024differentiating}. With this, we could conclude that our derivation in \cref{sec:backward-pass} holds everywhere.}

\subsection{Effectiveness of the backward pass in approximating the gradient}
\label{appendix:empirical-properties-backward}
\reviewerThree{
In this section, we empirically assess how well the proposed backward pass approximates the true gradient induced by the projection operator. Concretely, we compare the \gls*{vjp} obtained from the implicit function theorem–based backward pass with finite-difference estimates on a representative problem instance.
}

\paragraph{Experimental setup}
\reviewerThree{
We fix two problem instances in the non-convex small benchmark (see \cref{sec:experiments:benchmarks}) by fixing the right-hand side of the equality constraints. We then generate a set of $100$ points-to-be-projected ${x_i}_{i=1}^{100}$ by sampling each coordinate from a normal distribution with standard deviation $10$. For each $x_i$, we also sample $100$ normally distributed vectors $v_i$ for comparing the \gls*{vjp}.
}
\reviewerThree{
As a reference ``ground truth'', we compute the \gls*{vjp} using central finite differences with step size $\varepsilon = 10^{-6}$, which yields an approximation with $\mathcal{O}(\varepsilon^2)$ truncation error. We then compute the \gls*{vjp} using our backward pass for varying numbers of backward iterations \texttt{n\_iter\_bwd}. For each configuration, we report (i) the $\ell_2$-norm of the difference and (ii) the cosine similarity, thereby quantifying both the absolute and directional accuracy of the proposed backward pass.
}
\reviewerThree{For both methods, we use $\texttt{n\_iter\_fwd} = 200$ forward iterations for computing the projection accurately.}

\begin{figure}
    \centering
    \includegraphics[width=\linewidth]{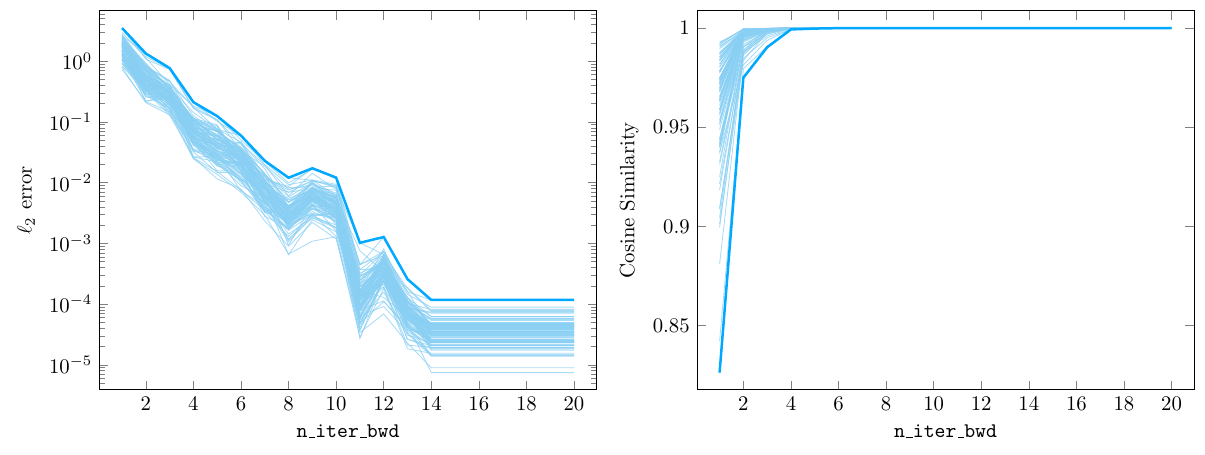}

    \includegraphics[width=\linewidth]{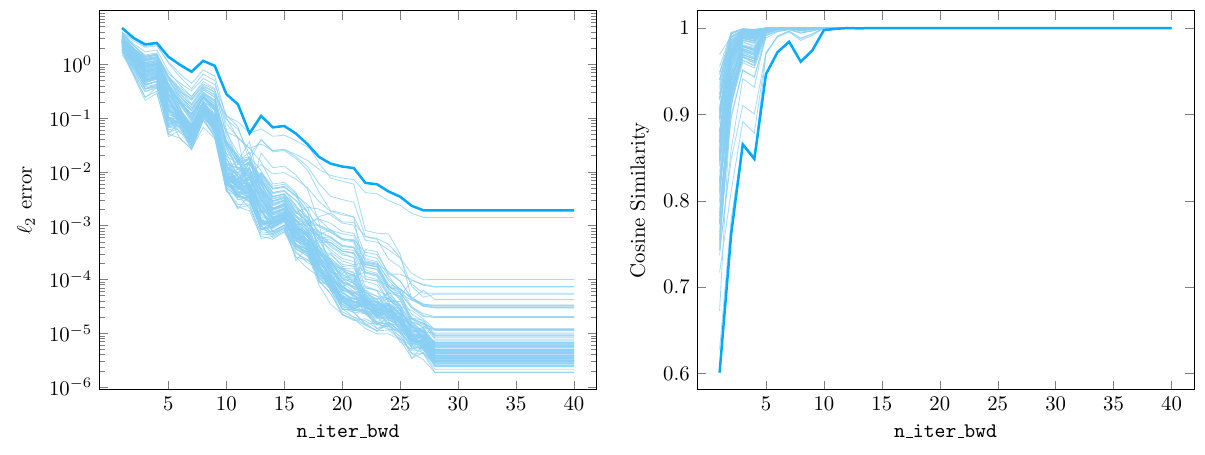}
    \caption{\reviewerThree{\gls*{vjp} estimation error (measured via the $\ell_2$-norm of the difference and the cosine similarity) for $100$ different vectors $v_i$ instances (light blue) for two different instances of the small non-convex benchmark as the number of iterations in the backward pass \texttt{n\_iter\_bwd} increases, as well as the maximum $\ell_2$-error and the minimum cosine similarity among all instances (dark blue).}}
    \label{fig:bwd-ablation}
\end{figure}

\paragraph{Results}
\reviewerThree{As shown in \cref{fig:bwd-ablation}, the discrepancy between the \gls*{vjp} computed with our backward pass and the finite-difference reference decreases rapidly as the number of backward iterations increases. These results indicate that the proposed backward pass yields a highly accurate approximation of the true gradient even with a relatively small number of iterations.}

\end{document}